\documentclass[]{style}

\usepackage[toc,page,header]{appendix}
\usepackage{enumitem}
\usepackage{xcolor}
\usepackage[dvipsnames]{xcolor}
\usepackage{listings}
\usepackage{caption}

\usepackage{natbib}

\usepackage{CJKutf8}

\usepackage{xargs}  

\usepackage{todonotes}  

\usepackage{multirow}

\usepackage{cleveref}

\usepackage{amsmath}
\usepackage{dsfont}

\usepackage{svg}

\usepackage{mathrsfs}
\usepackage{adjustbox}
\usepackage{multirow}
\usepackage{multicol}
\usepackage{tcolorbox}
\usepackage{changepage}
\usepackage{enumitem}
\usepackage{graphicx}
\usepackage{amssymb}
\usepackage{xcolor}
\usepackage{float}
\usepackage{multirow}
\usepackage{threeparttable}
\usepackage{graphicx}
\usepackage{subcaption}
\usepackage{algorithm}
\usepackage{algorithmic}
\usepackage{wrapfig}
\usepackage[table]{xcolor}  
\usepackage{colortbl}       
\usepackage{tabularx} 
\usepackage{makecell} 
\usepackage[dvipsnames]{xcolor}

\newcolumntype{g}{>{\columncolor{gray!10}}c} 
\newcolumntype{A}{>{\raggedright\arraybackslash}p{4.5cm}} 
\newcolumntype{Y}{>{\centering\arraybackslash}X} 

\definecolor{catgray}{gray}{0.9}
\definecolor{skyblue}{rgb}{0.53,0.81,0.92} 

\colorlet{skyblue!30}{skyblue!30!white} 

\definecolor{customblue}{RGB}{70,130,180}  

\newtcolorbox{evolbox}[2][]{%
  enhanced,
  colframe=customblue,
  colback=white,
  coltitle=white,
  rounded corners,
  boxrule=1pt,
  titlerule=0pt,
  toptitle=1mm,
  bottomtitle=1mm,
  fonttitle=\bfseries,
  width=#2\textwidth, 
  #1
}

\usepackage{minitoc}
\usepackage{url}

\PassOptionsToPackage{table,xcdraw}{xcolor}
\usepackage{titletoc}
\usepackage{placeins}
\usepackage{pifont}

\definecolor{RowBlue}{HTML}{E9F2FB}
\definecolor{RowRed}{HTML}{F9EAEA}
\definecolor{Top1}{HTML}{50DB4B} 
\definecolor{Top2}{HTML}{A5FFA2} 
\definecolor{Top3}{HTML}{D9FFD9} 
\definecolor{Sub1}{HTML}{EAB8B8}
\definecolor{Sub2}{HTML}{E4E4E4}

\definecolor{gearred}{HTML}{D85140}
\definecolor{reprablue}{HTML}{5384ED}
\definecolor{steorange}{HTML}{EF8444}
\definecolor{softgreen}{HTML}{658E40}

\renewcommand{\emph}[1]{\textit{#1}}

\definecolor{codepink}{RGB}{220,20,120}
\definecolor{codegreen}{RGB}{0,150,0}
\definecolor{codegray}{RGB}{140,140,140}
\definecolor{codeorange}{RGB}{230,120,60}

\lstdefinestyle{pytorchstyle}{
    language=Python,
    basicstyle=\ttfamily\small,
    keywordstyle=\color{codepink}\bfseries,
    commentstyle=\color{codegray}\itshape,
    stringstyle=\color{codeorange},
    numberstyle=\tiny\color{codegray},
    numbers=none,
    showstringspaces=false,
    breaklines=true,
    frame=none,
    columns=fullflexible,
    keepspaces=true,
    xleftmargin=1.5em
}

\title{Prism: Dynamic Sparse Attention for Native 2K Joint Video-Audio Generation Model Training}

\author{
  Shuyuan Tu\texorpdfstring{$^{1,2,*,\ddagger}$}{}
  \hspace{0.1cm}
  Qi Tian\texorpdfstring{$^{2,*}$}{}
  \hspace{0.1cm}
  Yinming Huang\texorpdfstring{$^{1}$}{}
  \hspace{0.1cm}
  Yue Wu\texorpdfstring{$^{2}$}{}
  \hspace{0.1cm}
  Xintong Han\texorpdfstring{$^{2}$}{}
  \hspace{0.1cm}
  Kaihang Pan\texorpdfstring{$^{3}$}{} \\
  \hspace{0.1cm}
  Weijie Kong\texorpdfstring{$^{2}$}{} \hspace{0.1cm}
  Jiangfeng Xiong{$^{2}$}{}
  \hspace{0.1cm}
  Jian-Wei Zhang{$^{2,\S}$}{}
  \hspace{0.1cm}
  Zuxuan Wu{$^{1,\dag}$}{}
  \hspace{0.1cm}
  Yu-Gang Jiang\texorpdfstring{$^{1,\dag}$}{}
  \hspace{0.1cm}
}

\makeatletter
\g@addto@macro\authorlist{\\[2mm]
  {\small
  \texorpdfstring{$^{*}$}{}Equal contribution \quad
  \texorpdfstring{$^{\ddagger}$}{}Work done during internship at Tencent Hunyuan \quad
  \texorpdfstring{$^{\S}$}{}Project leader \quad
  \texorpdfstring{$^{\dag}$}{}Corresponding author
  }\\[0.2mm]
}
\makeatother

\affiliation[1]{Fudan University}
\affiliation[2]{Tencent Hunyuan Foundation Model Team}
\affiliation[3]{Zhejiang University}

\newcommand{\answerTODO}[1][]{\textcolor{red}{\bfseries [TODO]}}
\newcommand{\justificationTODO}[1][]{\textcolor{red}{\bfseries [TODO]}}

\abstract{
Natively training joint video-audio generation models at higher resolutions empowers them to learn richer visual details and sharper motion dynamics.
However, full attention incurs quadratic cost and, as resolution increases, spreads attention over increasingly redundant tokens, diluting learning signals for informative content and disrupting pretrained priors.
Existing sparse attention methods either target training-free acceleration or overlook the unique structure of joint video-audio data, where cross-modal interactions are inherently concentrated around sound-producing regions. 
To address this, we propose Prism, a dynamic sparse attention framework for natively training joint video-audio generation models at 2K. 
In particular, Prism organizes the token sequence into spatiotemporal macro-zones, enabling the attention structure to adapt to local content. 
For each zone, it estimates local information structure via video feature variance along the channel and feature norms from the audio-to-video cross-attention, jointly capturing how visual content varies directionally and how strongly audio influences each visual region. 
Based on these signals, Prism dynamically assigns a tailored block shape to each zone, applying finer partitioning along axes of rapid visual content variation and strong audio-visual coupling.
This encourages tokens within each block to remain semantically coherent, allowing block-level features to capture both visual content and joint video-audio interaction patterns.
Prism further adopts a hybrid block selection strategy to dynamically determine per-query sparsity. 
Experiments show that Prism achieves 2.5$\times$ training speedup compared to full attention, while surpassing it in generation quality.
}

\checkdata[
\raisebox{-0.3em}{\includegraphics[width=0.025\linewidth]{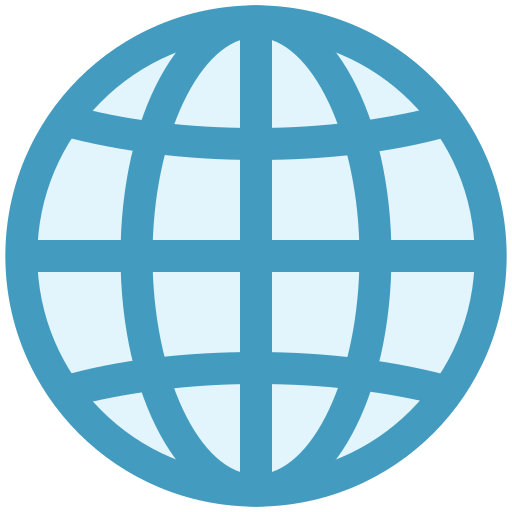}}~~Project Page]{\href{https://francis-rings.github.io/Prism}{\texttt{https://francis-rings.github.io/Prism}}
\\[-2.3ex]}

\checkdata[
\raisebox{-0.3em}{\includegraphics[width=0.025\linewidth]{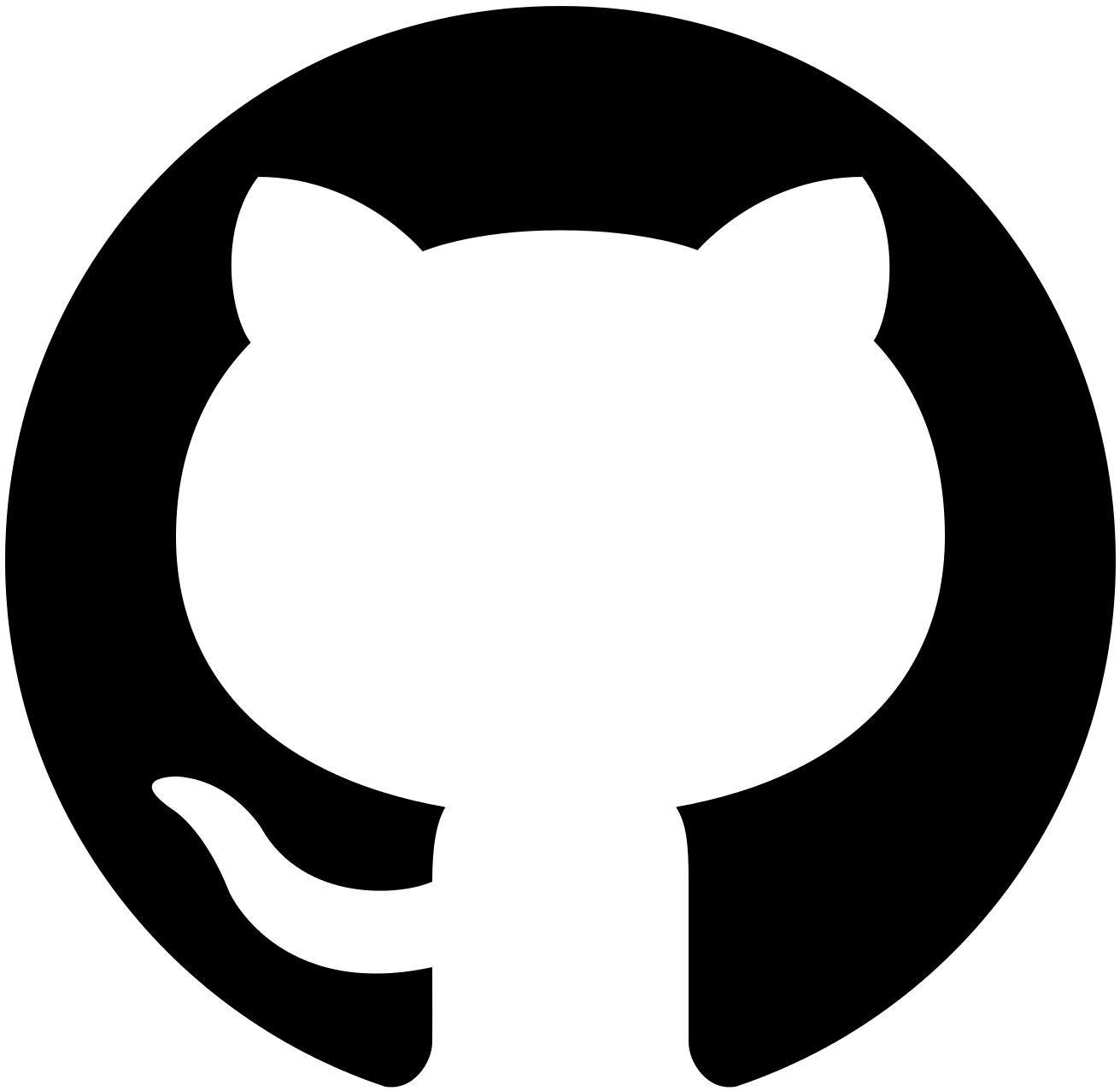}}~~GitHub Repo]{\href{https://github.com/Tencent-Hunyuan/Prism}{\texttt{https://github.com/Tencent-Hunyuan/Prism}}
\\[-2.3ex]}

\begin{document}
\maketitle

\section{Introduction}
\label{sec:intro}

\begin{figure*}[t!]
\begin{center}
\includegraphics[width=1.0\linewidth]{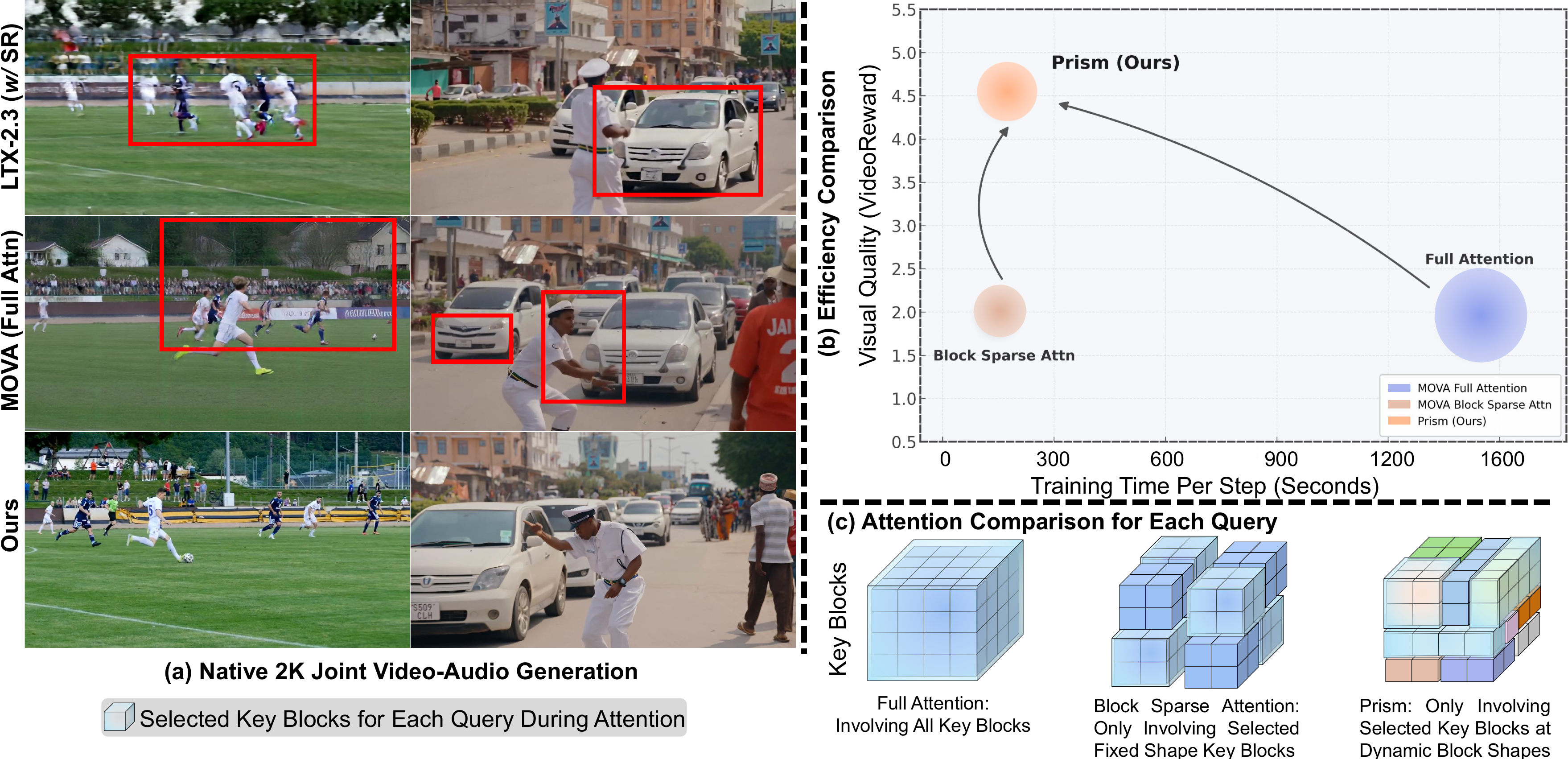}
\end{center}
\vspace{-0.2cm}
   \caption{Videos generated by Prism, showing its power to natively synthesize 2K video-audio. 
    LTX-2.3~\citep{hacohen2026ltx} first generates at 720p and then upsamples (SR) to 2K, and MOVA~\citep{team2026mova} performs native 2K generation by training with full attention on 2K videos.
   }
\label{fig:cover}
\vspace{-0.4cm}
\end{figure*}

Recent advances in scaling diffusion models~\citep{dhariwal2021diffusion,ho2020denoising} with larger datasets and model parameters significantly improve the quality of joint video-audio generation~\citep{liu2025javisdit, liu2025javisgpt, low2025ovi, hacohen2026ltx, ruan2023mm, wang2025universe, hu2025harmony, zhang2025uniavgen, team2026mova, tu2026baton, huang2026va, huang2026agentic}. As a result, joint video-audio generation emerges as a mainstream task in generative AI, synthesizing joint video-audio that is more natural and practical than previous silent video generation~\citep{hong2022cogvideo, wan2025}. 
Despite this progress, joint video-audio generation models have yet to scale natively to high-resolution training (2K).
While richer spatial detail and sharper motion dynamics directly improve video-audio fidelity, it is constrained by the quadratic cost of full attention. 
Under VAE compression~\citep{wan2025}, a 2K clip (12s, FPS=24) yields over one million tokens, making training costly.
Moreover, raising resolution adds far more tokens than genuinely informative content. Salient details such as faces and fast motion remain spatially localized, whereas repetitive backgrounds increasingly dominate the token sequence.
Full attention treats this ultra-long sequence indiscriminately. As redundant tokens vastly outnumber informative tokens, the attention weight on informative content is diluted (Appx. \ref{sec: attention_visualization_results}), degrading native high-resolution learning and pretrained priors.

To address this, researchers have explored various sparse attention methods~\citep{lu2026moba, yuan2025native, ma2025block, yuan2024ditfastattn, zhang2025fast, xi2025sparse, zhang2025spargeattn, xu2026deepseek, shao2026liveditor, wu2025hunyuanvideo1.5}. They basically reduce computational cost by restricting each query to interact with only a subset of key-value pairs. Most of them~\citep{xi2025sparse, zhang2025spargeattn, yang2026sparse_2, liu2025rectified, hu2026ultragen} are training-free methods. Only few trainable works~\citep{wu2025vmoba, zhang2026spargeattention2, team2025longcat, wu2025hunyuanvideo_1_5} focus on video generation. 
However, they accelerate pure video diffusion at 480P or 720P and have not scaled to native 2K joint video-audio training. 
Their sparsity patterns rely solely on visual redundancy and apply uniform criteria across tokens, ignoring the varying impact of audio across visual regions. As audio-visual synchronization is concentrated around sound-producing regions, such strategies fail to adapt sparsity to cross-modal structures, losing critical dependencies for joint generation. At 2K resolution, the abundance of repetitive background tokens further overwhelms sparse selection signals, making informative key-value pairs harder to identify.

In light of this, we propose Prism to perform native high-resolution training of joint video-audio generation models. 
Prism builds on block sparse attention~\citep{team2025longcat}, which partitions tokens into fixed-shape blocks, represents each block by mean pooling, scores query-key block relevance via representative dot products, and attends only to the highest-scoring key blocks (Fig. \ref{fig:cover}(c)).
However, recent methods~\citep{wu2025vmoba, zhang2026spargeattention2, team2025longcat, shao2026liveditor} use a fixed block shape for all regions. 
At high resolution, both visual details and redundant tokens increase drastically, yet they are distributed unevenly. 
Sound-producing regions exhibit rapid temporal variation and remain spatially compact, while static backgrounds contribute massive redundant tokens that are nearly uniform. 
Meanwhile, audio-visual coupling is concentrated around these active regions. 
Fixed-shape blocks fail to capture this structure, mixing semantically dissimilar tokens and producing unreliable block representatives, thereby degrading training.

To mitigate this, Prism organizes the token sequence into spatiotemporal macro-zones, enabling the attention structure to adapt to local content. 
For each zone, it estimates local information structure through two complementary guidance: video channel-wise variance guidance, which captures how rapidly visual content changes for each spatiotemporal axis, and audio-to-video cross-attention norm guidance, which measures how strongly audio influences each visual region. 
Based on them, Prism dynamically assigns a tailored 3D block shape to each zone (Fig. \ref{fig:cover}(c)), applying finer partitioning along axes of rapid content variation and strong audio-visual coupling. 
This encourages tokens within each block to remain semantically coherent, so their mean-pooling captures both visual content and cross-modal interaction patterns.
Furthermore, most block selection strategies apply a fixed sparsity ratio, selecting the same number of key blocks for every query.
In practice, queries in salient regions attend to a few nearby blocks while queries in uniform areas spread attention broadly, making fixed selection wasteful for the former and insufficient for the latter.
Prism addresses this with a hybrid Top-$k$ and Top-$p$ block selection strategy that dynamically determines per-query sparsity. 
Top-$k$ guarantees a minimum coverage, while Top-$p$ adaptively extends it for queries whose attention is broadly dispersed, concentrating computation on genuinely informative interactions.

As shown in Fig. \ref{fig:cover}(a), while LTX-2.3 (\textit{w/} SR)~\citep{hacohen2026ltx} and MOVA~\citep{team2026mova} suffer from body distortion and blurriness, Prism natively synthesizes stable 2K video-audio content, even involving complex human–object interactions. Fig. \ref{fig:cover}(b) shows that Prism enables efficient native high-resolution joint video-audio training compared with previous attention.

In conclusion, our contributions are: 
(1) We conduct an in-depth analysis of attention methods in native high-resolution joint video-audio training. We reveal that full attention over-models redundant tokens at 2K, while recent sparse methods miss audio-visual coupling. 
(2) We propose Prism, the first framework to explore native high-resolution training for joint video-audio generation.
To capture cross-modal coupling and visual variation, Prism performs sparse attention at dynamic block shapes guided by video channel-wise variance and audio-to-video cross-attention norms.
(3) Experiments show that Prism trains 2.5$\times$ faster than full attention while achieving better quality.

\section{Related Work}
\label{sec:related_work}

\noindent\textbf{Joint Video-Audio Generation.}
The advancement of industry-leading models~\citep{genimi_omni_2026, seedance2_2025} promotes the research interest in joint video-audio generation~\citep{haji2025av, ishii2025simple, liu2025javisdit, liu2025javisgpt, low2025ovi, hacohen2026ltx, ruan2023mm, wang2025universe, hu2025harmony, zhang2025uniavgen, tu2025stableavatar, team2026mova, tu2026baton}.
UniAVGen~\citep{zhang2025uniavgen} and the Javis series~\citep{liu2025javisdit, liu2025javisgpt} focus on joint video-speech generation.
UniVerse-1~\citep{wang2025universe}, Ovi~\citep{low2025ovi}, and Harmony~\citep{hu2025harmony} can synthesize diverse sounds. MOVA~\citep{team2026mova}, LTX-2.3~\citep{hacohen2026ltx}, JoyAI-Echo~\citep{echo2026joyai}, and daVinci-MagiHuman~\citep{davinci-magihuman-2026} scale the model parameters and training data. Baton~\citep{tu2026baton} uses blueprints for joint generation.
However, they cannot natively train at 2K or generate native 2K video-audio. Although LTX-2.3~\citep{hacohen2026ltx} generates at lower resolution and then upsamples to 2K, the visual details are limited during generation, and super-resolution only amplifies artifacts. 
By contrast, Prism enables native 2K training that directly learns rich visual details at full resolution.

\noindent\textbf{Efficient Attention.}
Linear attention methods~\citep{wang2025lingen, gu2023mamba, dao2024transformers, liu2024vmamba, peng2023rwkv} have linear complexity but require architectural modifications and cannot inherit pretrained priors.
Alternatively, sparse attention~\citep{lu2026moba, yuan2025native, ma2025block, xu2026deepseek, liu2026ab} restricts the set of keys attended by each query. 
In video generation, most of the methods~\citep{zhang2025fast, yuan2024ditfastattn, xi2025sparse, yang2026sparse_2, zhang2025spargeattn, liu2025rectified} are training-free.
To improve quality, some works explore trainable sparse attention.
VMoBA~\citep{wu2025vmoba} utilizes a layer-wise block partition scheme. 
SpargeAttention2~\citep{zhang2026spargeattention2} uses a distillation objective to train models. HunyuanVideo1.5~\citep{wu2025hunyuanvideo_1_5} and LongCat-Video~\citep{team2025longcat} scale the training data.
VSA~\citep{zhang2025vsa} uses coarse-to-fine attention without adapting to local structure.
LIVEditor~\citep{shao2026liveditor} applies 0th-order Taylor to redundant tokens.
However, these methods target inference acceleration at 480P/720P. Their fixed block shapes fail to capture the distinct structure of joint video-audio data. 
While VMoBA cycles through predefined 1/2/3D partitions, the partition is determined by layer index rather than content.

\noindent\textbf{Native High-Resolution Training.}
For image, FiT~\citep{lu2024fit} and NiT~\citep{wang2026native} use training-free extrapolation to higher unseen resolutions.
For video, UltraGen~\citep{hu2026ultragen} and T3~\citep{zhang2025transform} use window attention to infer/train at high resolution. 
PixelWizard~\citep{li2026pixelwizard} decouples spatial-temporal modeling. 
ViBe~\citep{wu2026vibe} adapts a Relay LoRA for a video DiT. 
Pyramidal Flow~\citep{jin2025pyramidal} and LUVE~\citep{zhao2026luve} adopt a cascaded pipeline at multiple resolutions.
However, none of them account for cross-modal coupling in joint video-audio. 
The closest UltraGen/T3 use fixed-shape windows that blur audio-visual coupling.

\section{Method}
\label{sec:method}

\begin{figure*}[t!]
\begin{center}
\includegraphics[width=1.0\linewidth]{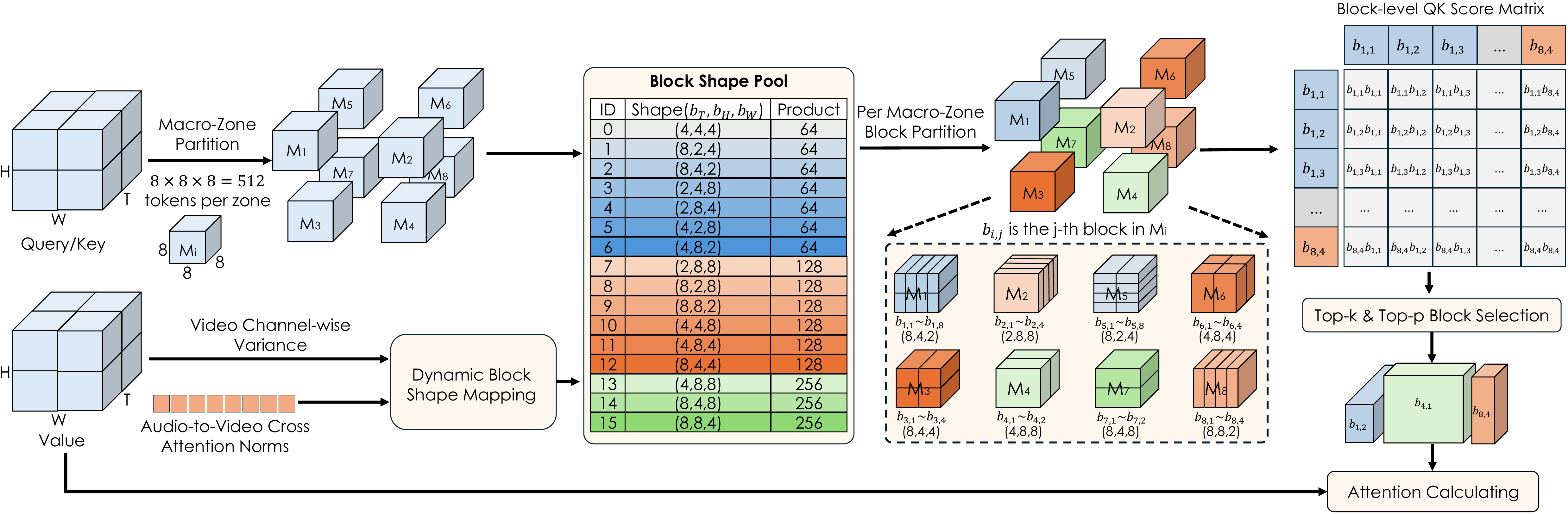}
\end{center}
\vspace{-0.2cm}
   \caption{Architecture of Prism. 
   Prism partitions the token sequence into spatiotemporal macro-zones, dynamically assigns each zone a tailored block shape, and applies hybrid Top-$k$/Top-$p$ block-sparse attention to focus computation on informative interactions.
   }
\label{fig:framework}
\vspace{-0.4cm}
\end{figure*}

As shown in Fig.~\ref{fig:framework}, Prism is a dynamic sparse attention framework that enables native 2K joint video-audio training. 
Prism inherits the MOVA~\citep{team2026mova} DiT and replaces video self-attention with its sparse attention (Fig.~\ref{fig:dit_framework}).
It organizes the token sequence into macro-zones, dynamically assigns a tailored block shape to each zone based on video channel-wise variance and audio-to-video cross-attention norms (Sec.~\ref{sec: dynamic_sparse_block_shape}), followed by a hybrid block selection (Sec.~\ref{sec: dynamic_sparsity}).

\subsection{Dynamic Block Shape}
\label{sec: dynamic_sparse_block_shape}

Native high-resolution training enables the model to learn richer visual details and sharper motion dynamics.
However, current models~\citep{team2026mova,hacohen2026ltx,davinci-magihuman-2026} cannot put this scaling law into practice due to their full attention, which incurs quadratic cost and spreads attention weight across the massive redundant tokens, diluting the attention weight on informative content and disrupting pretrained priors.
While recent trainable sparse attention methods~\citep{wu2025vmoba,zhang2026spargeattention2} reduce the cost, they rely on fixed-shape blocks, assuming semantic coherence within each block. 
This breaks down for 2K video-audio data, as the information structure varies across spatiotemporal axes. 
Intuitively, sound-producing regions are spatially compact yet temporally dynamic, while repetitive backgrounds make blocks highly redundant, overwhelming informative ones during sparse attention. 
Moreover, audio-visual interactions are highly localized around these active regions. 
To tackle this, we propose Prism, which dynamically assigns a tailored 3D block shape to each local region based on video channel-wise variance and audio-to-video cross-attention norms, capturing fine visual details and audio-visual coupling.

Concretely, to enable content-adaptive block partitioning, we first divide the video token grid ($T \times H \times W$) into non-overlapping macro-zones of shape $Z_T \times Z_H \times Z_W$, yielding $N_m = \frac{T}{Z_T} \cdot \frac{H}{Z_H} \cdot \frac{W}{Z_W}$ zones. 
For brevity, we omit padding details. 
Note that the Triton kernel uses a fixed 64-token tile size, so block sizes must be multiples of 64. 
Thus, we set $Z_{T/H/W}=8$, allowing each zone to be exactly tiled by any 3D block shape with per-axis edge lengths in $\{2,4,8\}$.
Prism applies sparse attention only to the video self-attention, since 
it dominates the cost at 2K, whereas the cross-attention and audio branch operate on a much shorter sequence, making the cost negligible.

In each zone $m$, Prism estimates a variation indicator $\mathbf{g}_m = (g_T, g_H, g_W)$ that jointly captures how visual content varies directionally and how strongly audio influences each visual region in the temporal/height/width axes. 
The computation is performed independently per attention head and per layer (indices omitted for clarity), allowing each to adapt its block shape to the evolving feature distribution across the network depth.
$\mathbf{g}_m$ combines the video channel-wise variance guidance and the audio-to-video cross-attention norm guidance.
The video channel-wise variance guidance measures how video features vary along each spatiotemporal axis at the channel level. 
In the video self-attention, we aggregate the value features $\mathbf{V}_m \in \mathbb{R}^{Z_T \times Z_H \times Z_W \times D}$ within zone $m$ along each axis, where $D$ is the head dimension. 
We use value features as they are directly aggregated into the attention output, making their directional variation the most faithful indicator of block boundaries. 
For $d \in \{T, H, W\}$, we collapse the other two axes by averaging to obtain the mean feature along $d$. Taking the temporal axis as example, $\bar{\mathbf{V}}_{m,t}^{(T)} = \frac{1}{Z_H Z_W}\sum_{h=1}^{Z_H}\sum_{w=1}^{Z_W} \mathbf{V}_{m,t,h,w} \in \mathbb{R}^D$ is the mean feature at temporal position $t$. We then measure how much this feature varies across $Z_{T/H/W}$:
\begin{equation}\small
\label{eq:video_variance}
\begin{aligned}
     \sigma^2_{v,d}(m) &= \frac{1}{D} \sum_{c=1}^{D} \frac{1}{Z_d}\sum_{i=1}^{Z_d}\!\left(\bar{V}_{m,i,c}^{(d)} - \mu_{m,c}^{(d)}\right)^2, \quad \mu_{m,c}^{(d)} = \frac{1}{Z_d}\sum_{i=1}^{Z_d} \bar{V}_{m,i,c}^{(d)}, \quad d \in \{T, H, W\},
\end{aligned}
\end{equation}
where $\bar{V}_{m,i,c}^{(d)}$ is the $c$-th channel of the axis-wise mean feature at the $i$-th position along axis $d$. 
As each channel encodes a distinct semantic aspect, averaging the per-channel variances retains sensitivity to all attributes.
We further normalize it into an anisotropy ratio $r_{v,d}(m) = \frac{\sigma^2_{v,d}}{\sigma^2_{v, T} + \sigma^2_{v, H} + \sigma^2_{v,W}}$ for $d \in \{T, H, W\}$.
A large $r_{v,d}$ signals that visual content varies rapidly along axis $d$, demanding finer block partitioning along that axis. Regarding the audio-to-video cross-attention norm guidance, it captures the directional structure of audio-visual coupling at each zone. 
In the DiT, each video hidden state $\mathbf{h}_{t,h,w}$ is modulated by an audio-to-video cross-attention $\hat{\mathbf{h}}_{t,h,w}=\mathrm{CrossAttn}(\mathbf{h}_{t,h,w},\, \mathbf{K}_a,\, \mathbf{V}_a)$ (Fig. \ref{fig:dit_framework}).
$\mathbf{K}_a$ and $\mathbf{V}_a$ are the key and value projected from the audio branch. 
$\hat{\mathbf{h}}_{t,h,w}$ encodes how much audio information is injected into each video token, and $\hat{\mathbf{h}}_{t,h,w}^{m}$ is in the macro-zone $m$. 
We derive the per-zone audio coupling strength $\bar{a}(m)$ as:
\begin{equation}\small
\label{eq:audio_video_coupling}
\begin{aligned}
     \bar{a}(m) = \frac{\frac{1}{Z_T Z_H Z_W}\sum_{t,h,w} a_{t,h,w}^{m}}{\max_{m' \in \{1,\dots,N_m\}} \frac{1}{Z_T Z_H Z_W}\sum_{t,h,w} a_{t,h,w}^{m'}}, \quad a_{t,h,w}^{m} = \|\hat{\mathbf{h}}_{t,h,w}^{m}\|_2,
\end{aligned}
\end{equation}
where $a_{t,h,w}^{m}$ is the audio L2-norm in $m$. It maps $\bar{a}(m) \in [0,1]$, removing the layer-dependent magnitude scale to ensure stable guidance.
To further capture the audio-visual coupling, we obtain:
\begin{equation}\small
\label{eq:audio_variance}
\begin{aligned}
     \bar{a}_{m,t}^{(T)} &= \frac{1}{Z_H Z_W}\sum_{h=1}^{Z_H}\sum_{w=1}^{Z_W} a^{m}_{t,h,w}, \quad \bar{a}_{m,h}^{(H)} = \frac{1}{Z_T Z_W}\sum_{t=1}^{Z_T}\sum_{w=1}^{Z_W} a^{m}_{t,h,w}, \quad
     \bar{a}_{m,w}^{(W)} = \frac{1}{Z_T Z_H}\sum_{t=1}^{Z_T}\sum_{h=1}^{Z_H} a^{m}_{t,h,w}, \\
     \sigma^2_{a,d}(m) &= \frac{1}{Z_d}\sum_{i=1}^{Z_d}\left(\bar{a}_{m,i}^{(d)} - \frac{1}{Z_d}\sum_{j=1}^{Z_d}\bar{a}_{m,j}^{(d)}\right)^2, \quad \hat{v}_{a,d}(m) = \frac{\sigma^2_{a,d}(m)}{\sigma^2_{a,T}(m) + \sigma^2_{a,H}(m) + \sigma^2_{a,W}(m)},
\end{aligned}
\end{equation}
where $d \in \{T, H, W\}$ and $\hat{v}_{a,d}(m)$ encodes the directional structure of audio influence (e.g., lip regions have high temporal audio variance as the mouth opens and closes). We then have $\mathbf{g}_m$:
\begin{equation}\small
\label{eq:effective_gradient}
\begin{aligned}
      g_d(m) = r_{v,d}(m) + \bar{a}(m) \cdot \hat{v}_{a,d}(m) + \epsilon, \quad d \in \{T, H, W\},
\end{aligned}
\end{equation}
where $\bar{a}$ acts as a multiplicative gate and $\epsilon{=}10^{-4}$. In sound-producing regions (high $\bar{a}$), it steers block shape assignment toward audio-visual coupling. In silent regions (low $\bar{a}$), the audio term vanishes.
Given $\mathbf{g}_m$, Prism selects a per-head block shape for each macro-zone from a candidate set $\mathcal{C}$ of 16 shapes, constructed by $(b_T, b_H, b_W)$ with $b_d \in \{2, 4, 8\}$ whose product $B$ falls in $\{64, 128, 256\}$:
\begin{equation}\small
\label{eq:candidate_set}
\begin{aligned}
      \mathcal{C}_{64} &= \{(4,4,4),(8,2,4),(8,4,2),(2,4,8),(2,8,4),(4,2,8),(4,8,2)\}, \\
    \mathcal{C}_{128} &= \{(2,8,8),(8,2,8),(8,8,2),(4,4,8),(4,8,4),(8,4,4)\}, 
    \mathcal{C}_{256} = \{(4,8,8),(8,4,8),(8,8,4)\},
\end{aligned}
\end{equation}
As a result, $B=64$ provides fine-grained blocks for regions rich in visual details and audio-visual coupling. 
$B{=}256$ is for the massive static backgrounds, whose tokens are highly repetitive.
$B{=}128$ serves intermediate regions. 
$B$ is determined by the zone's information density $\rho_{\mathrm{raw}}(m)$:
\begin{equation}\small
\label{eq:zone_information_density}
\begin{aligned}
      \rho_{\mathrm{raw}}(m) = \underbrace{\frac{1}{D}\sum_{c=1}^{D}\frac{1}{Z_TZ_HZ_W}\sum_{t,h,w}
    \!\left(V_{m,t,h,w,c}-\bar{\mu}_{m,t,c}\right)^2}_{\text{intra-zone spatial variance}}
    + \underbrace{\frac{1}{D}\sum_{c=1}^{D}\frac{1}{Z_T{-}1}\sum_{t=1}^{Z_T-1}
    \!\left(\bar{V}_{m,t+1,c}^{(T)}-\bar{V}_{m,t,c}^{(T)}\right)^2}_{\text{temporal frame-difference variance}},
\end{aligned}
\end{equation}
where $\bar{\mu}_{m,t,c} = \frac{1}{Z_H Z_W}\sum_{h,w} V_{m,t,h,w,c}$.
We further scale $\rho(m) = \frac{\rho_{\mathrm{raw}}(m)}{\max_{m' \in \{1,\dots,N_m\}} \rho_{\mathrm{raw}}(m')}\in [0,1]$.
High $\rho$ needs fine-grained blocks, while low $\rho$ is highly repetitive and benefits from larger blocks ($B(m)=64:\rho(m) \geq \tau_{128}$, $B(m)=128:\tau_{256} \leq \rho(m) < \tau_{128}$, $B(m)=256: \rho(m) < \tau_{256}$).
$\tau_{128}$ and $\tau_{256}$ are thresholds. 
We then select the block shape from $\mathcal{C}_{B(m)}$. We formulate shape assignment as minimizing intra-block information loss $\min_{b_T, b_H, b_W} \ g_T \cdot (b_T^2-1) + g_H \cdot (b_H^2-1) + g_W \cdot (b_W^2-1)$, where $b_T \cdot b_H \cdot b_W = B(m)$ and $\mathbf{g}_m=(g_T, g_H, g_W)$.
This penalizes long edges along axes with large variance, as grouping more tokens along rapidly varying axes causes greater loss in the mean-pooled representative.
We solve this objective via Lagrange multipliers and obtain the closed-form solution $b^*_d \propto g_d^{-1/2}$ for $d \in \{T, H, W\}$. 
The details are in Appx \ref{sec: proof_lagrange}. 
The axis with the largest variance receives the shortest edge (finest partitioning), keeping blocks internally coherent.

Since $b^*_d \propto g_d^{-1/2}$ is continuous while the candidate edge lengths are restricted to $\{2, 4, 8\}$, we quantize the continuous solution to the nearest candidate.
In each axis, we set $\ell_d = \log_2 b_d$. Thus, $b_Tb_Hb_W\rightarrow\ell_T + \ell_H + \ell_W = S$, where $S = \log_2 B(m)$. From $b^*_d \propto g_d^{-1/2}$, taking log$_2$ gives $\log_2 b^*_d = -\frac{1}{2}\log_2 g_d + C$, where $C$ is a constant shared across all three axes that controls the absolute scale.
Thus, we have $(-\frac{1}{2}\log_2 g_T + C) + (-\frac{1}{2}\log_2 g_H + C) + (-\frac{1}{2}\log_2 g_W + C) = S$, which gives $C = \frac{1}{3}(S + \frac{1}{2}\log_2 g_T + \frac{1}{2}\log_2 g_H + \frac{1}{2}\log_2 g_W)$. Substituting back:
\begin{equation}\small
\label{eq:quantize_solution}
\begin{aligned}
      \ell^*_d = -\tfrac{1}{2}\log_2 g_d + \tfrac{1}{3}\left(S + \tfrac{1}{2}\log_2 g_T + \tfrac{1}{2}\log_2 g_H + \tfrac{1}{2}\log_2 g_W\right), \quad d \in \{T, H, W\}, 
\end{aligned}
\end{equation}
Finally, we select the candidate shape $s^*$ from $\mathcal{C}_{B(m)}$, whose log$_2$ edge lengths are closest to the ideal:
\begin{equation}\small
\label{eq:final_decision}
\begin{aligned}
      s^* = \arg\min_{s \in \mathcal{C}_{B(m)}} \left[(\ell^{(s)}_T - \ell^*_T)^2 + (\ell^{(s)}_H - \ell^*_H)^2 + (\ell^{(s)}_W - \ell^*_W)^2\right], 
\end{aligned}
\end{equation}
where $(\ell^{(s)}_T, \ell^{(s)}_H, \ell^{(s)}_W)$ = $(\log_2 b^{(s)}_T, \log_2 b^{(s)}_H, \log_2 b^{(s)}_W)$.
After shape assignment, all tokens in a macro-zone share the same block shape and are rearranged into a block-contiguous layout for the downstream sparse kernel. 
We do not apply token-level dynamic shapes (different shapes coexisting within one zone), as the Triton kernel requires memory-contiguous tiles. Token-level shapes incur irregular clustering, creating outlier tokens that require padding and disrupt coalesced memory access. 
Moreover, at the 2K scale, each zone occupies a small region where content is already relatively homogeneous, making zone-level decisions sufficient. 
Note that zones determine only block shape, not which key blocks each query attends to. 
Cross-block/zone semantics are captured by the downstream block selection (Sec.~\ref{sec: dynamic_sparsity}), which operates globally across all blocks. During training, the model further co-adapts its block-level features, learning to capture cross-block/zone semantics and to select informative blocks across boundaries.

\subsection{Dynamic Per-Query Sparsity}
\label{sec: dynamic_sparsity}

After block shape assignment, Q, K, V share the same block partition (they are from the same latents in self-attention), and key blocks are mean-pooled into representatives. 
We compute the dot products between the query block $\mathbf{Q}_i$ and the key block representatives, followed by softmax, to obtain the block relevance $\mathbf{P}_i \in \mathbb{R}^{N_b}$.
$N_b$ is the total number of key blocks.
To determine per-query sparsity, Prism employs a hybrid Top-$k$ and Top-$p$ strategy that selects attended key/value blocks $\mathcal{S}_i$:
\begin{equation}\small
\label{eq:hybrid_block_selection}
\begin{aligned}
      \text{Top-}p(\mathbf{P}_i, p) &= \{j_1, \dots, j_L\}, \quad \text{s.t.} \quad \textstyle\sum_{l=1}^{L} P_{i,j_l} \geq p \;\;\text{and}\;\; \textstyle\sum_{l=1}^{L-1} P_{i,j_l} < p, \\
      \mathcal{S}_i &= \text{Top-}k(\mathbf{P}_i,\, k) \;\cup\; \text{Top-}p(\mathbf{P}_i,\, p) \rightarrow \mathbf{O}_i = \mathrm{Softmax}\!\left(\frac{\mathbf{Q}_i \mathbf{K}_{\mathcal{S}_i}^\top}{\sqrt{D}}\right) \mathbf{V}_{\mathcal{S}_i}
\end{aligned}
\end{equation}
where $j_1, \dots, j_{N_b}$ are indices sorted by $P_{i,j_l} \geq P_{i,j_{l+1}}$.
$\mathbf{K}_{\mathcal{S}_i}$ and $\mathbf{V}_{\mathcal{S}_i}$ are selected blocks in final attention $\mathbf{O}_i$.
$\text{Top-}k$ selects the $k$ blocks with highest weights, and $\text{Top-}p$ selects the smallest set whose cumulative weight exceeds $p$. 
Top-$p$ adaptively expands the attended set (Top-$k$) for queries with dispersed attention, allocating more computation where needed.
$|\mathcal{S}_i|$ varies dynamically. Sharp queries select $k$ blocks, while flat queries may select more via Top-$p$. 
The details are in Appx. \ref{sec: dynamic_block_sparse_attn_details}.

\subsection{Training}
Prism applies sparse attention only to video self-attention (Appx. \ref{sec: more_implementation_details}), which dominates the cost at 2K. 
We use the video/audio flow matching loss~\citep{tu2026baton} to train the video/audio DiT $\hat{\bm{v}}_\theta^{v/a}(\cdot)$:
 \begin{equation}\small
\label{eq:loss_fm}
\begin{aligned}
\mathcal{L} = \mathbb{E}_{\bm{z}_0^{v/a},\, \bm{z}_1^{v/a},\, t} \big[ \|\hat{\bm{v}}_\theta^v(\bm{z}_t^v, \bm{z}_t^a, t, \bm{c}) - (\bm{z}_1^v \!-\! \bm{z}_0^v)\|_2^2 + \|\hat{\bm{v}}_\theta^a(\bm{z}_t^v, \bm{z}_t^a, t, \bm{c}) - (\bm{z}_1^a \!-\! \bm{z}_0^a)\|_2^2 \big],
\end{aligned}
\end{equation}
where $\bm{z}_1^{v/a} \sim \mathcal{N}(0, I)$. $\bm{z}_0^{v/a}$ and $\bm{c}$ are the VAE-encoded clean latents and the conditioning signals.
\section{Experiment}
\label{sec:experiment}

\subsection{Implementation Details}
Our training dataset (100k 2K clips) is aggregated from UltraVideo~\citep{zhang2026ultravideo} and videos collected from the Internet (Appx. \ref{sec: dataset_details}). Following previous works~\citep{team2026mova}, we evaluate our model on MOVA-Bench~\citep{team2026mova} and VABench~\citep{hua2026vabench}.
We conduct additional experiments on 300 unseen 2K videos (10 seconds long) with more complex motion patterns and appearance details, referred to the 2K-Bench, selected from the internet to assess our model's capability for native 2K training. 
Prism is initialized by MOVA~\citep{team2026mova}.
We train our model for 5 epochs.
The learning rate is $1e-5$.
We set Top-$k$ (95\%), Top-$p$ (0.2), $\tau_{128}$=0.5, and $\tau_{256}$=0.25 
(Appx. \ref{sec: more_implementation_details}).

\subsection{Comparison with State-of-the-Art Methods}

\textbf{Quantitative Results.}
Following previous works~\citep{tu2026baton, team2026mova}, we utilize AQ~\citep{hu2025harmony}, TF~\citep{huang2024vbench}, DD~\citep{huang2024vbench}, and ID~\citep{hu2025harmony} to assess the video quality.
We further apply PQ~\citep{tjandra2025meta_audio_aesthetics}, CU~\citep{tjandra2025meta_audio_aesthetics}, cpCER~\citep{team2026mova}, Sync-C~\citep{chung2016out}, Sync-D~\citep{chung2016out}, and DeSync~\citep{iashin2024synchformer} to assess the audio quality and video-audio synchronization.
We use MUSIQ~\citep{ke2021musiq} and MANIQA~\citep{yang2022maniqa} to evaluate the high-resolution visual quality. MotionQ assesses the overall motion quality (Appx. \ref{sec: metrics}).
The quantitative comparison results on VABench, MOVA-Bench, and 2K-Bench are shown in Table \ref{table:quantitative_comparison_vabench}, Table \ref{table:quantitative_comparison_mova_bench}, and Table \ref{table:quantitative_comparison_2kbench}.
Notably, experiments on VABench and MOVA-Bench are conducted at the default 720p resolution, with Prism trained on our training set resized to 720p. For 2K-Bench, Prism and all competitors are trained on our native 2K training set. During evaluation, all models directly synthesize native 2K video-audio clips, except LTX-2.3, which follows its original pipeline of generating 720p outputs and upscaling them to 2K.
On 2K-Bench, MOVA, Ovi, and MagiHuman all employ full attention on the same native 2K data yet underperform Prism, confirming that full attention at 2K causes attention weight dilution (Appx. \ref{sec: attention_visualization_results}) where the massive redundant tokens claim most of the attention weight, leaving informative content under-attended and disrupting pretrained priors. LTX-2.3 generates at 720p and upsamples to 2K, so its visual details are bounded by the low-resolution stage, and super-resolution only amplifies existing artifacts. Prism overcomes both issues by dynamically concentrating attention on informative blocks, enabling the model to absorb the richer visual details and sharper motion dynamics that native 2K training provides.  
On 720p VABench and MOVA-Bench, Prism still outperforms all competitors, but by a smaller margin, as shorter sequences contain less redundancy, reducing the harm of full attention and the benefit of sparse attention.

\begin{figure*}[t!]
\begin{center}
\includegraphics[width=1.0\linewidth]{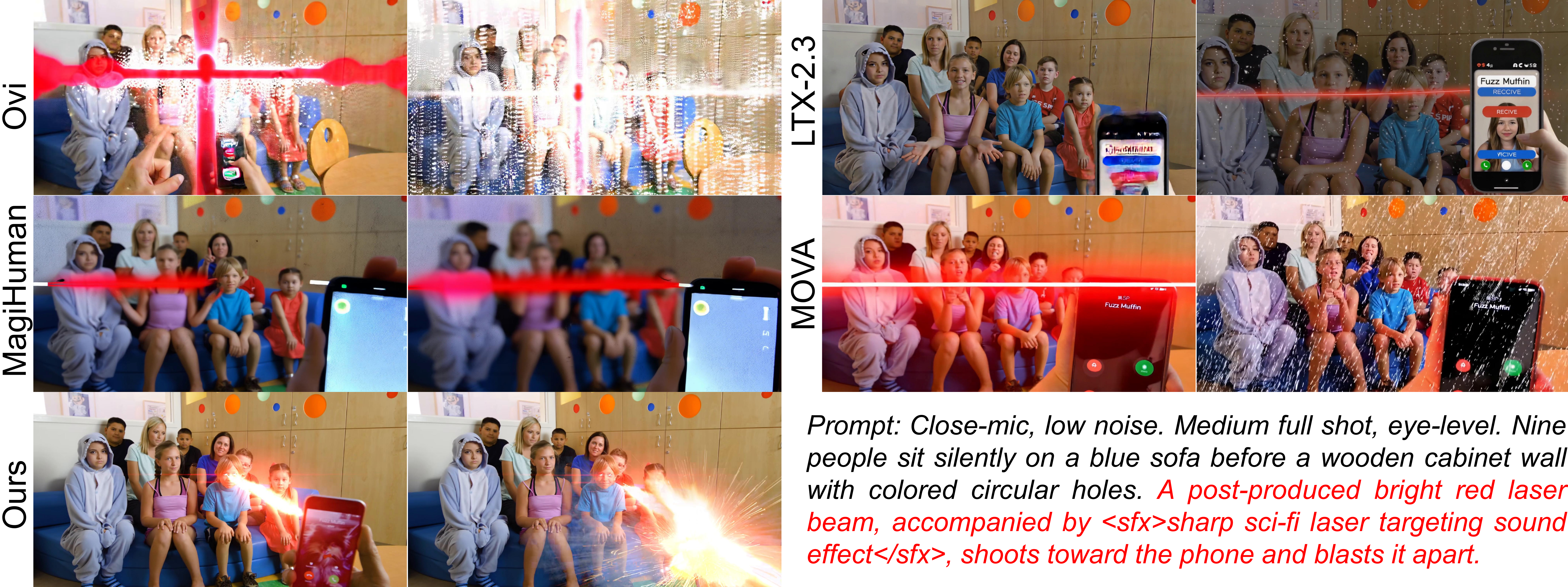}
\end{center}
\vspace{-0.2cm}
   \caption{Qualitative Comparisons with previous open-source methods. Please refer to the demo video for audio. More results are in the Appx. \ref{sec: more_comparison_results}.
   }
\label{fig:main_comparison}
\vspace{-0.2cm}
\end{figure*}

\begin{table}[t!]\small
\caption{Quantitative comparisons with previous open-source methods on 2K-Bench. 
}
\vspace{-0.05in}
\begin{center}
\renewcommand\arraystretch{1.1}
\scalebox{0.78}{
\begin{tabular}{lccccccccccccc}
\toprule
Model     & AQ$\uparrow$            & DD$\uparrow$            & TF$\uparrow$             & ID$\uparrow$            & PQ$\uparrow$            & CU$\uparrow$            & DeSync$\downarrow$        & Sync-D$\downarrow$        & Sync-C$\uparrow$        & cpCER$\downarrow$          & MUSIQ$\uparrow$          & MANIQA$\uparrow$         & MotionQ$\uparrow$       \\ \midrule
Ovi~\citep{low2025ovi}       & 0.38          & 0.35          & 0.891          & 0.85          & 6.68          & 5.92          & 1.12          & 8.28          & 5.21          & 0.468          & 51.43          & 0.326          & 0.42          \\
LTX-2.3~\citep{hacohen2026ltx}   & 0.48          & 0.41          & 0.943          & 0.91          & 7.05          & 6.83          & 0.95          & 7.62          & 5.92          & 0.382          & 55.60          & 0.403          & 0.66          \\
MagiHuman~\citep{davinci-magihuman-2026} & 0.40          & 0.37          & 0.912          & 0.87          & 6.72          & 6.14          & 1.08          & 8.05          & 5.45          & 0.420          & 52.68          & 0.337          & 0.58          \\
MOVA~\citep{team2026mova}      & 0.42          & 0.39          & 0.903          & 0.88          & 6.80          & 6.26          & 1.05          & 7.96          & 5.62          & 0.374          & 53.21          & 0.363          & 0.53          \\ \midrule
Ours      & \textbf{0.61} & \textbf{0.52} & \textbf{0.982} & \textbf{0.94} & \textbf{7.69} & \textbf{7.34} & \textbf{0.63} & \textbf{6.74} & \textbf{7.27} & \textbf{0.187} & \textbf{62.25} & \textbf{0.438} & \textbf{0.89} \\ \bottomrule
\end{tabular}
}
\end{center}
\label{table:quantitative_comparison_2kbench}
\vspace{-0.20in}
\end{table}

\noindent\textbf{Qualitative Results.}
The results are shown in Fig. \ref{fig:main_comparison}.
Notably, all models except LTX-2.3 are fine-tuned on our 2K training dataset for a fair comparison.
All competitors suffer from flickering, visual melting, and human distortions. In contrast, Prism remains structurally coherent by efficiently capturing the rich visual details and clear motion patterns that native 2K training provides.

\noindent\textbf{Comparison with Industry-Leading Models.}
We compare our model with commercial models, as shown in Appx. \ref{sec: commercial_model_comparison_results}.
Commercial models outperform our model in visual realism and audio aesthetics due to large model parameters and datasets. However, the gap is smaller in video-audio stability under large motions. Although our model is behind commercial models in these aspects, it significantly narrows the gap between open-source and industry-leading models.

\subsection{Ablation Study}

\textbf{Sparse Attention.}
We conduct an ablation study on sparse attention, as shown in Table \ref{table:ablation_sparse_attention} and Fig. \ref{fig:ablation_sparse_attn}. All ablated models are trained on the same 2K dataset under the same settings until loss convergence ($\geq$ 5 epochs).
Base Sparsity is the Top-$k$ budget and SSTA is Selective and Sliding Tile Attention.
We obtain the following observations: 
(1) Full Attn performs poorly at 2K as the massive redundant tokens dominate the attention weight, leaving informative content under-attended and destabilizing pretrained priors.
(2) BSA achieves the worst results, as fixed-shape blocks mix semantically dissimilar tokens in regions with heterogeneous spatiotemporal structure, corrupting block representatives.
(3) Other trainable sparse methods improve over BSA but share a common limitation of ignoring the distinct cross-modal coupling structure of joint video-audio data, making their uniform sparsity patterns insufficient at 2K.
(4) Training-free methods perform below Full Attn as they do not alter training and merely approximate an already suboptimal model output.
(5) Prism improves over SpargeAttn2 by 46\% in MotionQ and 30\% in DeSync, showing that content-adaptive block shapes concentrate on genuinely informative interactions that all competitors fail to capture.

\begin{table}[t!]\small
\caption{Ablation study on different sparse attention for native 2K joint video-audio training. Training-free methods are applied at inference to the Full Attn. Trainable methods replace original video self-attention during native 2K training. All competitors use their optimal sparsity settings.
}
\vspace{-0.05in}
\begin{center}
\renewcommand\arraystretch{1.1}
\scalebox{0.75}{
\begin{tabular}{l|l|ccccccccc}
\toprule
Category & Model     & AQ$\uparrow$ & TF$\uparrow$ & PQ$\uparrow$ & CU$\uparrow$ & DeSync$\downarrow$ & cpCER$\downarrow$ & MANIQA$\uparrow$ & MotionQ$\uparrow$ & Base Sparsity \\ \midrule
\multirow{2}{*}{Training-Free} & SVG-2~\citep{yang2026sparse_2}        & 0.35 & 0.861 & 6.23 & 5.71 & 1.28 & 0.462 & 0.318 & 0.38 & 71\% \\
& Sol-Attn~\citep{li2026sol}     & 0.40 & 0.892 & 6.64 & 6.08 & 1.11 & 0.398 & 0.349 & 0.47 & 85\% \\ \midrule
\multirow{7}{*}{Trainable} & Full Attn                              & 0.42 & 0.903 & 6.80 & 6.26 & 1.05 & 0.374 & 0.363 & 0.53 & 0\% \\
& Block Sparse Attn (BSA)~\citep{team2025longcat} & 0.33 & 0.842 & 6.12 & 5.58 & 1.36 & 0.487 & 0.306 & 0.34 & 90\% \\
& VSA~\citep{zhang2025vsa}              & 0.39 & 0.878 & 6.53 & 6.04 & 1.08 & 0.391 & 0.347 & 0.46 & 90\% \\
& SSTA~\citep{wu2025hunyuanvideo_1_5}   & 0.43 & 0.911 & 6.82 & 6.31 & 0.97 & 0.356 & 0.368 & 0.54 & 85\% \\
& VMoBA~\citep{wu2025vmoba}             & 0.46 & 0.924 & 7.04 & 6.52 & 0.91 & 0.328 & 0.381 & 0.59 & 90\% \\
& SpargeAttn2~\citep{zhang2026spargeattention2} & 0.47 & 0.928 & 7.10 & 6.55 & 0.90 & 0.319 & 0.384 & 0.61 & 95\% \\ \cmidrule{2-11}
& Ours      & \textbf{0.61} & \textbf{0.982} & \textbf{7.69} & \textbf{7.34} & \textbf{0.63} & \textbf{0.187} & \textbf{0.438} & \textbf{0.89} & 95\% \\ \bottomrule
\end{tabular}
}
\end{center}
\label{table:ablation_sparse_attention}
\vspace{-0.20in}
\end{table}

\noindent\textbf{Native 2K Training/Inference.}
We compare with other native 2K training/inference methods, as shown in Table \ref{table:native_2k_comparison} and Fig. \ref{fig:ablation_native_training}.
Training-free methods directly apply to backbones pretrained only at 720p and perform 2K inference.
PyramidFlow~\citep{jin2025pyramidal} uses multi-stage latent denoising to progressively upsample to 2K, with LoRA adaptation at each resolution. LUVE~\citep{zhao2026luve} uses dual-frequency LoRA experts with low-/high-pass filtering for attention and FFN. T3~\citep{zhang2025transform} replaces video self-attention with multi-scale window attention.
We have the following observations:
(1) Training-free methods perform poorly, as the 720p-pretrained backbone has never been exposed to the attention patterns that emerge only at 2K.
(2) Cascaded approaches (LUVE and PyramidFlow) never learn the joint distribution of all 2K tokens in a single denoising pass, introducing a low-resolution bias. Spatial structures and motion patterns are determined at the coarse grid with limited precision, and the refinement stage cannot restructure them after upsampling. Audio-visual synchronization is similarly bounded, as temporal alignment established at low resolution with only a few coarse tokens cannot be re-learned during refinement.
(3) T3~\citep{zhang2025transform} trains natively at 2K in a single pass but samples tokens at uniform strides across the spatiotemporal volume, which is content-agnostic. Prism overcomes these limitations via native 2K training with content-adaptive dynamic sparse attention in a single pass.

\noindent\textbf{Dynamic Block Shape.}
We first ablate block shape, dynamic guidance, and information density in our dynamic block shape mechanism in Table \ref{table:ablation_dynamic_block}, Fig. \ref{fig:ablation_block_shape}, and Fig. \ref{fig:ablation_guidance}, as detailed in Appx. \ref{sec: more_ablation_dynamic_block_shape}.
Fixed $8^3$ and Random Shape perform worst, confirming that content-agnostic assignment corrupts block representatives. 
Our three-tier allocation surpasses the single finest tier Only $\mathcal{C}_{64}$ in both quality and speed.
Removing $r_{v,d}$ causes the most severe visual drop, while \textit{w/o} Audio Guidance degrades cpCER and DeSync, revealing their complementary roles. 
Key features outperform Query features for variance estimation, as Key encodes content closer to Value. 
For $\rho$ design, results validate that both spatial and temporal variance are essential.
We further ablate shape mapping (Eq.~\ref{eq:quantize_solution}), heads/layers, and $\tau_{128}$/$\tau_{256}$ in Table \ref{table:ablation_shape_mapping} and Fig. \ref{fig:ablation_mapping}, as detailed in Appx. \ref{sec: more_ablation_dynamic_block_shape}.
Performance peaks at the Lagrange-derived $\alpha{=}1/2$ and degrades on both sides, as $\alpha{=}1/3$ forces blocks to remain nearly isotropic, while $\alpha{=}2$ is unstable under noisy variance.
Alternative mappings (Softmax, Rank, Entropy) impose functional forms that deviate from the optimal concave relationship, and Isotropic shows that anisotropic shape assignment is essential.
$\tau_{128}/\tau_{256}$ peaks at (0.5, 0.25), with tighter thresholds over-allocating fine blocks to backgrounds and looser ones assigning coarse blocks to informative zones.

\noindent\textbf{Resolution Scaling.}
We ablate training resolutions in Table \ref{table:ablation_resolution} and Fig. \ref{fig:ablation_different_resolution}, as detailed in Appx. \ref{sec: native_training_at_scaling_resolution}.
Unlike parameter scaling and data scaling, resolution scaling lifts the spatial ceiling itself, exposing genuinely new information such as finger articulations and fine lip movements that is not clear at lower-resolution grids. However, this scaling axis is fragile. Higher resolution also introduces a growing proportion of redundant tokens that can actively degrade training when the attention mechanism fails to distinguish informative content from repetitive backgrounds.
The results confirm this. 
Full Attn leads at 480P where redundancy is minimal, but drops at 2K, as the redundant tokens overwhelm the learning signal. Fixed-shape sparse attention methods filter some visual redundancy but apply uniform block shapes that cannot adapt to the increasingly uneven information distribution at higher resolution, missing the spatially compact sound-producing regions that demand finer partitioning. Cascaded PyramidFlow avoids the full 2K attention cost, but spatial structures and motion patterns are determined at the coarse grid and cannot be restructured after upsampling. 
All competitors degrade on cpCER and DeSync at 2K, as none of them incorporate audio-aware mechanisms and the shrinking spatial proportion of sound-producing regions at higher resolution makes cross-modal coupling difficult to capture. \textit{w/o} Video Guidance causes visual metrics to basically stagnate beyond 1080P, while \textit{w/o} Audio Guidance fails to improve synchronization steadily, confirming their complementary roles.
Prism is the only method that improves on all metrics from 480P to 2K due to its video channel-wise variance guidance and audio-to-video cross-attention norm guidance.

\noindent\textbf{Sparsity.}
We ablate sparsity in Table~\ref{table:ablation_sparsity}.
Only Top-$p$ lacks minimum coverage, losing critical context. Only Top-$k$ at 85\% achieves the best single-strategy quality, while 95\% aggressively drops informative blocks and 75\% admits more redundant tokens.
The hybrid Top-$k$(95\%)+Top-$p$(0.2) combines minimum coverage with adaptive expansion for dispersed queries, achieving the best quality-speed trade-off, as Top-$p$(0.3) yields negligible gains but incurs 27\% slower training time.

\noindent\textbf{Different Backbones.} We ablate the backbone of our DiT, as shown in Table \ref{table:ablation_backbone}. We replace the video self-attention with our dynamic sparse attention while keeping all other components unchanged.
The results validate the robustness of our model across different DiT backbones.

\noindent\textbf{Visualization.}
We visualize the training loss, as shown in Appx.~\ref{sec: native_2k_training_loss}. Prism achieves the lowest and most stable loss on both modalities, while Full Attn exhibits instability and other sparse methods show rising audio loss, confirming that content-agnostic attention fails to capture cross-modal coupling at 2K.
We further visualize the attention maps and block shape assignments in Appx.~\ref{sec: attention_visualization_results}.
The results show that $\mathcal{C}_{64}$ concentrates on dynamic regions while $\mathcal{C}_{256}$ covers static backgrounds. 
The same region adaptively switches temporal, spatial, and audio-visual partitioning granularity depending on local content, demonstrating principled rather than arbitrary shape assignment.

\subsection{Applications and User Study}

\noindent\textbf{Speed and GPU Resources.}
We compare the inference latency and GPU cost between Prism and previous models, as shown in Appx. \ref{sec: speed_gpu_resource_comparison}.
Compared to our Full Attn backbone MOVA, Prism reduces inference time by 3$\times$ and GPU memory by 47\%, while improving AQ by 45\%, MotionQ by 68\%, and DeSync by 40\%. Compared to the leading competitor LTX-2.3, Prism achieves comparable inference speed while substantially outperforming it across all quality metrics.

\noindent\textbf{Complex Scene.}
We evaluate our model on scenarios involving complex motion patterns and intricate human-object interactions, as shown in Appx. \ref{sec: complex_scene_results}.
Prism maintains high generation quality in long-horizon scenes requiring coherent multi-step actions and rich human-environment interactions.

\noindent\textbf{User Study.}
We conduct a user study on 30 selected videos at 2K. The participants are university faculty and students. In each case, participants are first presented with the reference image and text prompt. We then offer two videos, one of which is synthesized by Prism and the other is synthesized by a competitor. Participants are then asked to answer questions: 
V-A/A-A: "Which one has better video/audio alignment with text prompts?" V-Q/A-Q/M-Q/V-A-S: "Which one has better visual/acoustic/motion detail quality or video-audio synchronization?" Appx. \ref{sec: user_study_details} shows the superiority of our model in subjective evaluation.

\section{Conclusion}
\label{sec:conclusion}

In this paper, we proposed Prism, a dynamic sparse attention framework for natively training joint video-audio generation models at 2K resolution.
To address the quadratic cost and attention weight dilution caused by the massive redundant tokens at 2K, Prism organized the token sequence into spatiotemporal macro-zones and dynamically assigned a tailored 3D block shape to each zone based on video channel-wise variance and audio-to-video cross-attention norms.
This kept each block semantically coherent, allowing block-level features to capture visual content variation and audio-visual coupling.
Prism further adopted a hybrid strategy to dynamically determine per-query sparsity, concentrating computation on informative interactions.
Experiments demonstrated that Prism achieves 2.5$\times$ training speedup compared to full attention at 2K while surpassing it in generation quality.

\bibliographystyle{plainnat}
\setlength{\bibhang}{0pt}
\setlength\bibindent{0pt}
\bibliography{main}

\appendix
\section{Appendix}

\begin{figure*}[h]
\begin{center}
\includegraphics[width=0.75\linewidth]{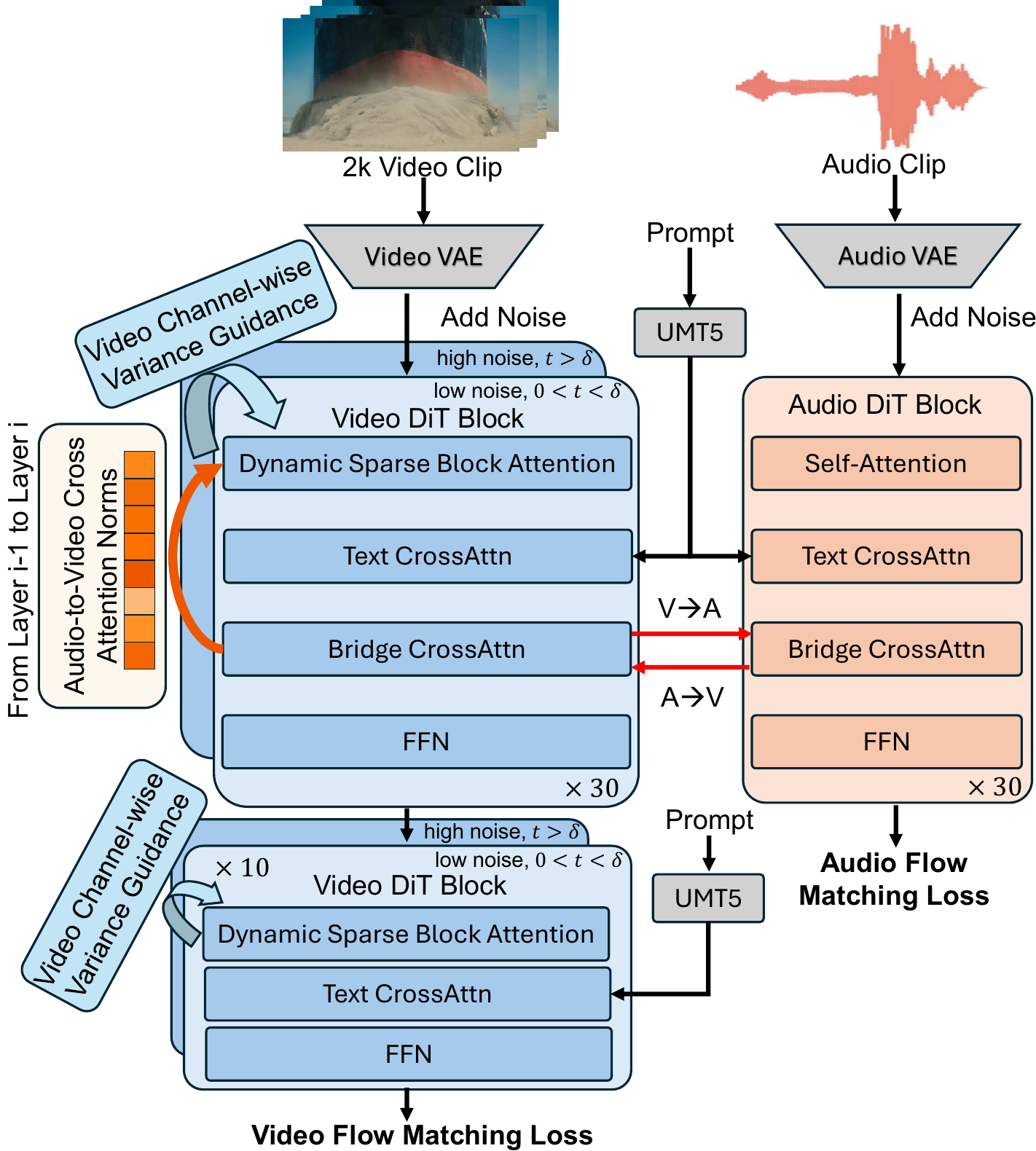}
\end{center}
\vspace{-0.2cm}
   \caption{Architecture of the dual-branch DiT of Prism.
   }
\label{fig:dit_framework}
\vspace{-0.2cm}
\end{figure*}

\begin{table}[h]\small
\caption{Quantitative comparisons with previous open-source methods on VABench.
We use their original evaluation metrics for fair comparison.
}
\vspace{-0.10in}
\begin{center}
\renewcommand\arraystretch{1.1}
\scalebox{0.72}{
\begin{tabular}{lcccccccccc}
\toprule
Model     & Audio-Aes$\uparrow$      & T-V Align$\uparrow$      & T-A Align$\uparrow$      & A-V Align$\uparrow$      & DeSync$\downarrow$        & Visual Realism$\uparrow$ & Audio Realism$\uparrow$  & Audio QA$\uparrow$       & Video QA$\uparrow$       \\ \midrule
Ovi~\citep{low2025ovi}       & 3.494          & 0.226          & 0.307          & 0.168          & 0.983          & 4.963          & 4.593          & 0.679          & 0.725          \\
LTX-2.3~\citep{hacohen2026ltx}   & 3.683          & 0.225          & 0.277          & 0.224          & 0.871          & 4.958          & 4.570          & 0.751          & 0.743          \\
MagiHuman~\citep{davinci-magihuman-2026} & 3.452          & 0.196          & 0.290          & 0.192          & 0.694          & 4.134          & 4.248          & 0.610          & 0.686          \\
MOVA~\citep{team2026mova}      & 3.318          & 0.219          & 0.386          & 0.243          & 0.942          & 4.969          & 4.537          & 0.783          & 0.716          \\ \midrule
Ours      & \textbf{3.691} & \textbf{0.228} & \textbf{0.394} & \textbf{0.262} & \textbf{0.691} & \textbf{4.973} & \textbf{4.608} & \textbf{0.786} & \textbf{0.772} \\ \bottomrule
\end{tabular}
}
\end{center}
\label{table:quantitative_comparison_vabench}
\vspace{-0.10in}
\end{table}

\begin{table}[h]\small
\caption{Quantitative comparisons with previous open-source methods on MOVA-Bench. We use their original evaluation metrics for a fair comparison.
}
\vspace{-0.02in}
\begin{center}
\renewcommand\arraystretch{1.1}
\scalebox{0.75}{
\begin{tabular}{lccccccc}
\toprule
Model     & IS$\uparrow$             & DNSMOS$\uparrow$         & DeSync$\downarrow$         & IB-Score$\uparrow$       & LSE-D$\downarrow$          & LSE-C$\uparrow$          & cpCER$\downarrow$          \\ \midrule
Ovi~\citep{low2025ovi}       & 3.680          & 3.516          & 0.515          & 0.190          & 7.468          & 6.378          & 0.436          \\
LTX-2.3~\citep{hacohen2026ltx}   & 3.326          & 3.708          & 0.342          & 0.312          & 8.063          & 6.447          & 0.197          \\
MagiHuman~\citep{davinci-magihuman-2026} & 3.075          & 3.420          & 0.561          & 0.238          & 11.637         & 2.566          & 0.487          \\
MOVA~\citep{team2026mova}      & 3.814          & 3.751          & 0.370          & 0.297          & 7.094          & 7.452          & 0.218          \\ \midrule
Ours      & \textbf{4.103} & \textbf{3.866} & \textbf{0.314} & \textbf{0.335} & \textbf{6.827} & \textbf{7.689} & \textbf{0.128} \\ \bottomrule
\end{tabular}
}
\end{center}
\label{table:quantitative_comparison_mova_bench}
\vspace{-0.20in}
\end{table}

\begin{table}[h]\small
      \caption{Comparison with native 2K training/inference methods. 
      Training Time is the per-step training time. GPU Mem is per-GPU memory during 4-GPU parallel inference, as the token count at 2K (10s, FPS=24) reaches approximately 870K, making single-GPU inference prohibitively expensive.
      Each competitor is trained until loss convergence. 
      }
      \vspace{-0.10in}
      \begin{center}
      \renewcommand\arraystretch{1.1}
      \scalebox{0.59}{
      \begin{tabular}{l|l|cccccccccc}
      \toprule
      Category & Model     & AQ$\uparrow$ & TF$\uparrow$ & PQ$\uparrow$ & CU$\uparrow$ & DeSync$\downarrow$ & cpCER$\downarrow$ & MANIQA$\uparrow$ & MotionQ$\uparrow$ & Training Time$\downarrow$ & Infer GPU Mem$\downarrow$ \\ \midrule
      \multirow{2}{*}{Training-Free} & MOVA-720p$\to$2K~\citep{team2026mova}       & 0.36 & 0.871 & 6.37 & 5.79 & 1.21 & 0.443 & 0.291 & 0.37 & - & 72.8G \\
      & UltraGen~\citep{hu2026ultragen}     & 0.43 & 0.908 & 6.71 & 6.15 & 1.06 & 0.386 & 0.347 & 0.46 & - & 48.0G \\ \midrule
      \multirow{6}{*}{Natively Training} & Full Attn                              & 0.42 & 0.903 & 6.80 & 6.26 & 1.05 & 0.374 & 0.363 & 0.53 & 26.5min & 72.8G \\
      & SpargeAttn2~\citep{zhang2026spargeattention2} & 0.47 & 0.928 & 7.10 & 6.55 & 0.90 & 0.319 & 0.384 & 0.61 & \textbf{8.4min} & \textbf{36.7G} \\
      & LUVE~\citep{zhao2026luve}              & 0.50 & 0.926 & 7.19 & 6.68 & 0.85 & 0.312 & 0.402 & 0.62 & 18.7min & 62.0G \\
      & PyramidFlow~\citep{jin2025pyramidal}       & 0.51 & 0.944 & 7.08 & 6.59 & 0.84 & 0.295 & 0.391 & 0.70 & 18.3min & 65.8G \\
      & T3~\citep{zhang2025transform}                & 0.53 & 0.939 & 7.28 & 6.84 & 0.79 & 0.268 & 0.396 & 0.71 & 16.9min & 52.0G \\ \cmidrule{2-12}
      & Ours      & \textbf{0.61} & \textbf{0.982} & \textbf{7.69} & \textbf{7.34} & \textbf{0.63} & \textbf{0.187} & \textbf{0.438} & \textbf{0.89} & 10.6min & 38.0G \\ \bottomrule
      \end{tabular}
      }
      \end{center}
      \label{table:native_2k_comparison}
      \vspace{-0.10in}
      \end{table}

\begin{table}[h]\small
\caption{Ablation on different DiT backbones.
}
\vspace{-0.02in}
\begin{center}
\renewcommand\arraystretch{1.1}
\scalebox{0.78}{
\begin{tabular}{lcccccccc}
\toprule
Model     & AQ$\uparrow$ & TF$\uparrow$ & PQ$\uparrow$ & CU$\uparrow$ & DeSync$\downarrow$ & cpCER$\downarrow$ & MANIQA$\uparrow$ & MotionQ$\uparrow$ \\ \midrule
LTX-2.3~\citep{hacohen2026ltx}    & 0.48 & 0.943 & 7.05 & 6.83 & 0.95 & 0.382 & 0.403 & 0.66 \\
LTX-2.3 + Prism & 0.63 & 0.986 & 7.74 & 7.41 & 0.66 & 0.208 & 0.452 & 0.86 \\ \midrule
Ovi~\citep{low2025ovi}        & 0.38 & 0.891 & 6.68 & 5.92 & 1.12 & 0.468 & 0.326 & 0.42 \\
Ovi + Prism     & 0.53 & 0.952 & 7.38 & 6.68 & 0.79 & 0.251 & 0.387 & 0.71 \\ \bottomrule
\end{tabular}
}
\end{center}
\label{table:ablation_backbone}
\end{table}

\begin{table}[h]\small
\caption{Ablation study on sparsity strategies. Top-$k$($x$\%) applies $x$ sparsity per query. Top-$p$($y$) selects the smallest set of key blocks whose cumulative attention weight exceeds $y$.
}
\vspace{-0.05in}
\begin{center}
\renewcommand\arraystretch{1.1}
\scalebox{0.75}{
\begin{tabular}{lccccccccc}
\toprule
Setting     & AQ$\uparrow$ & TF$\uparrow$ & PQ$\uparrow$ & CU$\uparrow$ & DeSync$\downarrow$ & cpCER$\downarrow$ & MANIQA$\uparrow$ & MotionQ$\uparrow$ & Training Time$\downarrow$ \\ \midrule
Only Top-$k$ (75\%)   & 0.50 & 0.937 & 7.02 & 6.58 & 0.87 & 0.294 & 0.386 & 0.64 & 14.3min \\
Only Top-$k$ (85\%)   & 0.56 & 0.961 & 7.41 & 6.97 & 0.74 & 0.243 & 0.417 & 0.76 & 11.8min \\
Only Top-$k$ (95\%)   & 0.48 & 0.926 & 6.89 & 6.43 & 0.92 & 0.312 & 0.379 & 0.61 & 9.8min \\ \midrule
Only Top-$p$ (0.1)    & 0.29 & 0.831 & 5.87 & 5.32 & 1.48 & 0.513 & 0.291 & 0.26 & 9.2min \\
Only Top-$p$ (0.2)    & 0.36 & 0.862 & 6.28 & 5.74 & 1.24 & 0.441 & 0.321 & 0.37 & 9.9min \\
Only Top-$p$ (0.3)    & 0.43 & 0.894 & 6.71 & 6.18 & 1.02 & 0.378 & 0.356 & 0.49 & 11.1min \\ \midrule
Top-$k$(95\%)+Top-$p$(0.1)  & 0.54 & 0.952 & 7.32 & 6.84 & 0.76 & 0.258 & 0.409 & 0.74 & 10.2min \\
Top-$k$(95\%)+Top-$p$(0.2)(Ours)  & 0.61 & 0.982 & 7.69 & 7.34 & 0.63 & 0.187 & 0.438 & 0.89 & 10.6min \\
Top-$k$(95\%)+Top-$p$(0.3)  & \textbf{0.63} & \textbf{0.984} & \textbf{7.72} & \textbf{7.39} & \textbf{0.62} & \textbf{0.183} & \textbf{0.442} & \textbf{0.90} & 13.5min \\ \bottomrule
\end{tabular}
}
\end{center}
\label{table:ablation_sparsity}
\vspace{-0.20in}
\end{table}

\subsection{Preliminaries}
\label{sec: preliminaries}

We adopt the Rectified Flow~\citep{lipman2022flow} formulation for diffusion-based generation~\citep{dhariwal2021diffusion, ho2020denoising, zheng2024videogen, nichol2021improved, song2020score, song2020denoising, zheng2026manifold, pan2026omniweaving, qiu2026spatialfusion, wang2025selftok, jiang2026empirical,long2025follow,pang2025dreamdance}. Given a clean latent $\bm{z}_{0}\sim\bm{p}_{\text{data}}$ and a noise sample $\bm{z}_{1}\sim\mathcal{N}(0,I)$, the forward corruption constructs a straight-line interpolation parameterized~\citep{yang2026arcflow, tu2025stableanimator, tu2025stableanimator++, tu2024motioneditor, tu2025motionfollower, tu2026flashportrait} by a continuous timestep $t \in [0,1]$:
\begin{equation}\small
\label{eq:forward_diffusion}
\begin{aligned}
    \bm{z}_{t}=(1-t)\bm{z}_{0}+t\bm{z}_{1},
\end{aligned}
\end{equation}
A velocity prediction network $\hat{\bm{v}}_{\theta}(\bm{z}_{t},t)$ learns to estimate the direction $\bm{z}_{1}-\bm{z}_{0}$ at each noise level, which is then used to iteratively transport pure noise back to the data manifold during inference. The model is optimized with the mean squared error between the predicted and ground-truth velocity:
\begin{equation}\small
\label{eq:mse_loss}
\begin{aligned}
     \mathcal{L} = \mathbb{E}_{\bm{z}_{0},\bm{z}_{1},t}(\left \| (\bm{z}_{1}-\bm{z}_{0}) - \hat{\bm{v}}_{\theta}(\bm{z}_{t}, t)  \right \|^{2}).
\end{aligned}
\end{equation}

\subsection{Evaluation Metrics}
\label{sec: metrics}

To comprehensively evaluate model performance, we assess 2K-Bench using multiple metrics covering video quality, audio quality, video-audio synchronization, high-resolution visual details, and motion quality.

\noindent\textbf{Video Quality.} We leverage Temporal Flickering (TF) and dynamic degree (DD) from VBench~\citep{huang2024vbench} to validate the synthesized video quality. 
We use the pretrained aesthetic-predictor-v2-5~\citep{aesthetic_predictor_v2_5} to predict the aesthetic quality (AQ)~\citep{hu2025harmony}.
We further evaluate the identity consistency (ID) by comparing the mean DINOv3~\citep{simeoni2025dinov3} representation between the first synthesized frame and the rest of the synthesized frames.

\noindent\textbf{Audio Quality.} We utilize the pretrained AudioBox~\citep{tjandra2025meta_audio_aesthetics} to evaluate the perceptual audio quality across production quality (PQ) and content usefulness (CU).
To assess the accuracy of multi-speaker speech, we use cpCER~\citep{team2026mova} to assess whether the generated outputs correctly preserve speaker identities and dialogue content. 
We compute cpCER using MOSS Transcribe Diarize~\citep{team2026mova}, which first performs speaker diarization with explicit speaker tags (e.g., [S01] and [S02]) and then applies automatic speech recognition to transcribe the corresponding spoken content for each identified speaker. 
The resulting transcripts are then compared with the ground-truth audio transcripts to compute the character error rate~\citep{radford2023robust}.

\noindent\textbf{Video-Audio Synchronization.}
We utilize Sync-C~\citep{chung2016out} and Sync-D~\citep{chung2016out} to measure the synchronization of lips with audio. 
We further use the pretrained Synchformer~\citep{iashin2024synchformer} to quantify the temporal misalignment between video and audio streams (DeSync).

\noindent\textbf{High-Resolution Visual Quality.}
To evaluate the visual detail quality after native high-resolution training, we leverage MUSIQ~\citep{ke2021musiq} and MANIQA~\citep{yang2022maniqa} to assess the perceptual quality and detail fidelity of individual frames.
The final score is obtained by averaging the metric values over all sampled frames.

\begin{figure*}[t!]
\begin{center}
\includegraphics[width=1.0\linewidth]{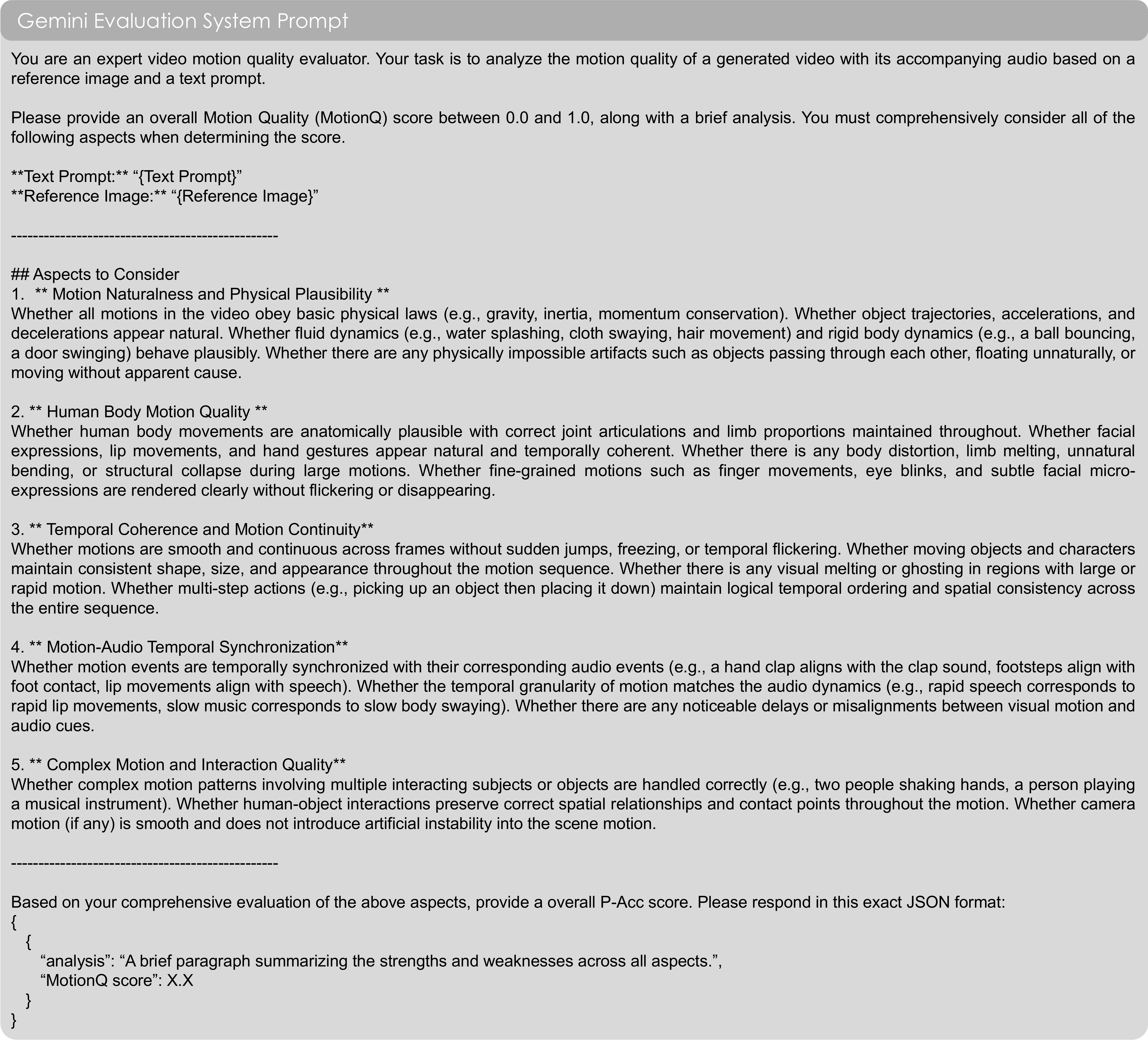}
\end{center}
\vspace{-0.3cm}
   \caption{The system prompt used for Gemini-based motion quality evaluation.
   }
\label{fig:motionq_format}
\vspace{-0.4cm}
\end{figure*}

\noindent\textbf{Motion Quality.}
Since conventional automatic metrics struggle to accurately assess video motion quality~\citep{li2024fine, pan2025generative, pan2026janus, zheng2025cml, tu2023implicit, tu2022multiple}, we leverage the powerful multimodal understanding capability of Gemini-3.1~\citep{gemini_3} to evaluate the overall motion quality of joint video-audio generation models (MotionQ).
MotionQ aims to demonstrate that native high-resolution training enhances the model's ability to learn complex motion patterns, leading to more effective motion modeling. 
The evaluation prompts are shown in Fig. \ref{fig:motionq_format}.

\subsection{Dataset Details}
\label{sec: dataset_details}

Regarding the training dataset, our training dataset (100k 2K video-audio clips) is aggregated from UltraVideo~\citep{zhang2026ultravideo} and videos ($8\sim12s$, FPS=24) collected from the internet (YouTube and Bilibili). 
Since our work targets native 2K training, we apply a multi-stage filtering pipeline with particular emphasis on high-resolution quality.
First, we retain only clips whose shorter side is at least 1440 pixels to ensure sufficient native resolution for 2K training, and discard clips that are upscaled from lower resolution by detecting low-frequency energy dominance in the Fourier spectrum.
We then use aesthetic-predictor-v2-5~\citep{aesthetic_predictor_v2_5} to predict the visual aesthetic quality and retain only clips with a score above 0.4.
To further ensure that the retained clips contain genuine high-resolution visual details rather than merely large-resolution but visually flat content, we compute MANIQA~\citep{yang2022maniqa} on randomly sampled frames and discard clips with a score below 0.3.
We filter out static or near-static videos by requiring a Dynamic Degree~\citep{huang2024vbench} above 0.2. 
For audio quality, we use AudioBox~\citep{tjandra2025meta_audio_aesthetics} to evaluate audio aesthetics and keep only clips with PQ above 6.0. 
For videos containing speech, we apply SyncNet~\citep{chung2016out} to assess lip-sync accuracy and retain only clips with a confidence score above 0.9.
Furthermore, we utilize Qwen3-VL-235B-A22B-Instruct~\citep{Qwen3-VL} and Qwen3-Omni~\citep{xu2025qwen3_omni} to caption the training videos, generating the text prompts.

The training dataset covers a diverse range of audiovisual scenarios, with a deliberate emphasis on categories involving substantial motion, as native high-resolution training is particularly beneficial for learning complex motion patterns that are poorly captured at lower-resolution grids.
Specifically, human-object and human-environment interaction account for approximately 20\% of the dataset, covering cooking, sports with fast and large body movements (basketball, tennis, swimming, dancing), crafting, and tool manipulation, all of which involve complex multi-step motions and frequent spatial displacement.
Multi-speaker conversation and interaction account for approximately 18\%, including face-to-face dialogues with expressive gestures, group discussions with overlapping body movements, and debate scenes with rapid turn-taking.
Singing and musical performance account for approximately 15\%, spanning solo vocal performances with expressive body swaying, instrument playing (piano, guitar, drums, violin) with intricate hand and arm coordination, and concert recordings with stage movement.
Single-speaker dialogue and monologue account for approximately 25\% of the dataset, covering interview-style narration, vlog-style talking heads, and lecture presentations with hand gestures and body language.
Nature and environmental scenes with dynamic elements account for approximately 10\%, including landscapes with wind-driven vegetation, rain, thunderstorms, ocean waves, and animal locomotion.
Urban and street scenes account for approximately 7\%, covering traffic with fast-moving vehicles, crowd activity with pedestrian flow, and city ambience.
The remaining approximately 5\% consists of sci-fi and virtual environment content, animation, and miscellaneous audiovisual scenarios.
Notably, over 75\% of the dataset contains significant motion dynamics (human-object interaction, multi-speaker interaction, musical performance, and expressive single-speaker content), providing the model with abundant motion-rich supervision to exploit the finer spatial grid that 2K resolution offers for motion pattern modeling.

\begin{figure*}[t!]
\begin{center}
\includegraphics[width=1.0\linewidth]{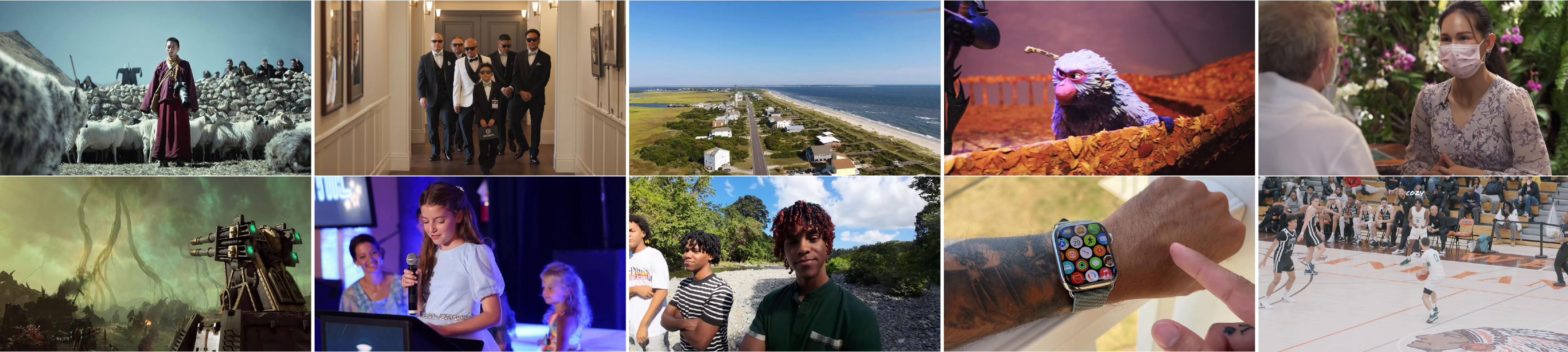}
\end{center}
\vspace{-0.2cm}
   \caption{Examples of 2K-Bench.
   }
\label{fig:2k-bench}
\vspace{-0.2cm}
\end{figure*}

In terms of the testing dataset, we select 300 unseen 2K videos (10 seconds long, FPS=24) from the internet to construct the testing dataset 2K-Bench. Some ground truth examples are shown in Fig. \ref{fig:2k-bench}.
The sources of videos mainly come from social media platforms, including YouTube and Bilibili.
These videos feature individuals across diverse ethnicities, genders, and age groups, portrayed in full-body, half-body, and close-up shots against varied indoor and outdoor settings.
\textbf{In contrast to existing open-source testing datasets such as VABench~\citep{hua2026vabench} and MOVA-Bench~\citep{team2026mova}, 2K-Bench differs in two fundamental aspects.}
\textbf{First, all reference images in 2K-Bench are natively captured at 2K resolution, preserving the full pixel-level detail of the original content.} Existing benchmarks provide reference images at 480p or 720p, and even when upsampled to 2K via super-resolution, these images inevitably suffer from pixel-level information loss, blurred high-frequency textures, and hallucinated details that do not faithfully represent the true 2K visual content.
\textbf{Second, 2K-Bench deliberately emphasizes complex visual details and challenging motion patterns in both reference images and text prompts, targeting the capabilities that native high-resolution training is specifically designed to improve.}
Existing benchmarks typically feature visually simple reference images (e.g., a person standing against a plain background with minimal texture) and motion-light prompts (e.g., ``a woman is talking'' or ``a man walks forward''), which can be adequately handled even by models trained at lower resolution. 
In contrast, 2K-Bench reference images contain rich fine-grained details such as intricate jewelry patterns, individual hair strands, visible skin texture, and complex fabric weaves that are only resolvable at 2K. The text prompts demand complex multi-stage motion sequences (e.g., ``a chef chops vegetables rapidly, tosses them into a sizzling wok, and stirs while steam rises''), compositional human-object interactions with precise spatial contact (e.g., ``a violinist's fingers shift between positions on the fingerboard while the bow arm executes rapid string crossings''), and intricate audio-visual temporal correspondence (e.g., ``a drummer performs a fill pattern where each stick hit on the snare, tom, and cymbal aligns precisely with the corresponding percussive sound'').
We utilize Qwen3-VL-235B-A22B-Instruct~\citep{Qwen3-VL} and Qwen3-Omni~\citep{xu2025qwen3_omni} to generate the text prompts, followed by manual verification and refinement to ensure accuracy.

2K-Bench is designed with a roughly uniform distribution across categories to provide balanced evaluation coverage. Specifically, single-speaker dialogue with expressive gestures accounts for 17\%, multi-speaker conversation and interaction accounts for 17\%, singing and musical instrument performance accounts for 17\%, human-object and human-environment interaction (cooking, sports, crafting) accounts for 16\%, nature and environmental scenes with dynamic elements (wind-driven vegetation, animal locomotion, water dynamics) accounts for 17\%, and urban and complex scenes (traffic, crowd activity, sci-fi content) accounts for 16\%.
Following CelebV-HQ~\citep{zhu2022celebvhq}, 2K-Bench is released under a non-commercial research license. The dataset is available for non-commercial research purposes only, and users agree not to reproduce, sell, or commercially exploit any portion of the videos or derived data.

To verify that 2K-Bench is out-of-distribution with respect to our training set and contains no data leakage, we perform three complementary checks. First, we conduct exact URL deduplication to ensure no testing video shares the same source URL as any training video. Second, we compute perceptual hashing on uniformly sampled frames from both sets and verify that no testing frame has a Hamming distance below 8 to any training frame, ruling out near-duplicate content even under different encodings or minor cropping. 
Third, we extract I3D features from all ground truth videos in both the training set and 2K-Bench, and compute the FVD between them. The FVD between the 2K-Bench ground truth videos and the training set videos is over 5.5$\times$ larger than the FVD computed between two random equal halves of the training set, confirming that the 2K-Bench ground truth videos come from a different distribution than our training data in the I3D feature space.

\begin{algorithm}[t!]
      \caption{Dynamic Block Sparse Attention}
      \label{alg:prism_alignmth}
      \small
      \begin{algorithmic}[1]
      \STATE \textbf{Input:} Video self-attention $\mathbf{Q}, \mathbf{K}, \mathbf{V} \in \mathbb{R}^{N \times D}$ ($T{\times}H{\times}W=N$), audio-to-video cross-attention norms $\{a_{t,h,w}\}$ (cached from the preceding layer; absent for the first layer, where $g_d$ reduces to $r_{v,d}{+}\epsilon$), zone edge length $Z_T{=}Z_H{=}Z_W{=}8$, candidate set $\mathcal{C}$, thresholds $\tau_{128}, \tau_{256}$, Top-$k$ threshold $k$, Top-$p$ threshold $p$.
      \STATE \textbf{Stage 1: Dynamic Block Shape Assignment} (per head, per layer)
      \STATE Divide $T{\times}H{\times}W$ grid into $N_m$ macro-zones; reshape $\mathbf{V}$ into $\{\mathbf{V}_m \in \mathbb{R}^{Z_T \times Z_H \times Z_W \times D}\}_{m=1}^{N_m}$.
      \FOR{each zone $m$}
          \STATE Compute axis-wise mean: $\bar{V}_{m,i,c}^{(T)} = \frac{1}{Z_H Z_W}\sum_{h,w} V_{m,i,h,w,c}$ (analogously $\bar{V}_{m,i,c}^{(H)}$, $\bar{V}_{m,i,c}^{(W)}$).
          \STATE Compute position mean: $\mu_{m,c}^{(d)} = \frac{1}{Z_d}\sum_{i=1}^{Z_d} \bar{V}_{m,i,c}^{(d)}$.
          \STATE Compute channel-wise variance: $\sigma^2_{v,d}(m) = \frac{1}{D}\sum_{c=1}^{D} \frac{1}{Z_d}\sum_{i=1}^{Z_d} (\bar{V}_{m,i,c}^{(d)} - \mu_{m,c}^{(d)})^2$.
          \STATE Compute anisotropy ratio: $r_{v,d}(m) = \sigma^2_{v,d} / (\sigma^2_{v,T} + \sigma^2_{v,H} + \sigma^2_{v,W})$.
          \STATE Compute audio axis-wise mean: $\bar{a}_{m,i}^{(T)} = \frac{1}{Z_H Z_W}\sum_{h,w} a^{m}_{i,h,w}$ (analogously for $H$, $W$).
          \STATE Compute audio coupling: $\bar{a}(m) = \frac{\frac{1}{Z_T Z_H Z_W}\sum_{t,h,w} a_{t,h,w}^{m}}{\max_{m' \in \{1,\dots,N_m\}} \frac{1}{Z_T Z_H Z_W}\sum_{t,h,w} a_{t,h,w}^{m'}}$.
          \STATE Compute audio directional variance: $\sigma^2_{a,d}(m) = \frac{1}{Z_d}\sum_{i=1}^{Z_d}(\bar{a}_{m,i}^{(d)} - \frac{1}{Z_d}\sum_j \bar{a}_{m,j}^{(d)})^2$; \\ 
          $\hat{v}_{a,d}(m) = \sigma^2_{a,d}(m) / (\sigma^2_{a,T}(m) + \sigma^2_{a,H}(m) + \sigma^2_{a,W}(m))$.
          \STATE Compute variation indicator: $g_d(m) = r_{v,d}(m) + \bar{a}(m) \cdot \hat{v}_{a,d}(m) + \epsilon$, \quad $d \in \{T, H, W\}$.
          \STATE Compute information density $\rho(m)$ via Eq.~\ref{eq:zone_information_density} (max-normalized to $[0,1]$).
          \STATE Set product level: $B(m) = \begin{cases} 64, & \rho(m) \geq \tau_{128} \\ 128, & \tau_{256} \leq \rho(m) < \tau_{128} \\ 256, & \rho(m) < \tau_{256} \end{cases}$, $S = \log_2 B(m)$.
          \STATE Compute ideal shape: $\ell^*_d = -\frac{1}{2}\log_2 g_d + \frac{1}{3}(S + \frac{1}{2}\log_2 g_T + \frac{1}{2}\log_2 g_H + \frac{1}{2}\log_2 g_W)$.
          \STATE Select shape: $s^* = \arg\min_{s \in \mathcal{C}_{B(m)}} [(\ell^{(s)}_T - \ell^*_T)^2 + (\ell^{(s)}_H - \ell^*_H)^2 + (\ell^{(s)}_W - \ell^*_W)^2]$.
      \ENDFOR
      \STATE \textbf{Stage 2: Token Partition}
      \STATE For each zone $m$ with selected shape $s^*=(b_T, b_H, b_W)$, partition its $Z_T{\times}Z_H{\times}Z_W$ tokens into $(Z_T/b_T){\cdot}(Z_H/b_H){\cdot}(Z_W/b_W)$ blocks of size $b_T{\times}b_H{\times}b_W$. Total $N_b$ blocks across all zones.
      \STATE \textbf{Stage 3: Block-Level Scoring \& Selection}
      \STATE Let $\{\mathcal{B}_1, \dots, \mathcal{B}_{N_b}\}$ denote the $N_b$ blocks from Stage 2. 
      \STATE Compute representatives: $\bar{\mathbf{Q}}_i = \frac{1}{|\mathcal{B}_i|}\sum_{n \in \mathcal{B}_i} \mathbf{Q}_n$, $\bar{\mathbf{K}}_j = \frac{1}{|\mathcal{B}_j|}\sum_{n \in \mathcal{B}_j} \mathbf{K}_n$. Let $\bar{\mathbf{K}} = [\bar{\mathbf{K}}_1;\dots;\bar{\mathbf{K}}_{N_b}] \in \mathbb{R}^{N_b \times D}$.
      \STATE Compute block scores: $\mathbf{P}_i = \mathrm{Softmax}(\bar{\mathbf{Q}}_i \bar{\mathbf{K}}^\top / \sqrt{D}) \in \mathbb{R}^{N_b}$.
      \STATE Hybrid selection: $\mathcal{S}_i = \text{Top-}k(\mathbf{P}_i, k) \cup \text{Top-}p(\mathbf{P}_i, p)$.
      \STATE $\mathcal{S}_i \subseteq \{1,\dots,N_b\}$ is selected key block indices for query block $i$.
      \STATE \textbf{Stage 4: Sparse Attention Kernel} (online softmax)
      \FOR{each query block $i = 1, \dots, N_b$}
          \STATE Load $\mathbf{Q}_i$ into SRAM (resident); load $\mathcal{S}_i$ from Stage 3.
          \STATE Initialize: $\boldsymbol{\xi}_i = -\infty \cdot \mathbf{1}_{|\mathcal{B}_i|}$, $\boldsymbol{\zeta}_i = \mathbf{0}_{|\mathcal{B}_i|}$, $\mathbf{O}_i = \mathbf{0}_{|\mathcal{B}_i| \times D}$. 
          \FOR{$j \in \mathcal{S}_i$}
              \STATE Load $\mathbf{K}_j, \mathbf{V}_j$ into SRAM; compute $\mathbf{A}_{ij} = \mathbf{Q}_i \mathbf{K}_j^\top / \sqrt{D} \in \mathbb{R}^{|\mathcal{B}_i| \times |\mathcal{B}_j|}$.
              \STATE $\boldsymbol{\xi}_{\text{new}} = \max(\boldsymbol{\xi}_i, \mathrm{rowmax}(\mathbf{A}_{ij}))$; \quad $\tilde{\mathbf{P}}_{ij} = \exp(\mathbf{A}_{ij} - \boldsymbol{\xi}_{\text{new}} \mathbf{1}^\top)$.
              \STATE $\boldsymbol{\zeta}_i = e^{\boldsymbol{\xi}_i - \boldsymbol{\xi}_{\text{new}}} \odot \boldsymbol{\zeta}_i + \mathrm{rowsum}(\tilde{\mathbf{P}}_{ij})$; \quad $\mathbf{O}_i = \mathrm{diag}(e^{\boldsymbol{\xi}_i - \boldsymbol{\xi}_{\text{new}}}) \mathbf{O}_i + \tilde{\mathbf{P}}_{ij} \mathbf{V}_j$.
              \STATE $\boldsymbol{\xi}_i = \boldsymbol{\xi}_{\text{new}}$.
          \ENDFOR
          \STATE $\mathbf{O}_i = \mathrm{diag}(\boldsymbol{\zeta}_i)^{-1} \mathbf{O}_i$; store $\mathbf{O}_i$ to HBM.
      \ENDFOR
      \STATE \textbf{Stage 5: Output}
      \STATE Scatter $\mathbf{O}=[\mathbf{O}_1; \dots; \mathbf{O}_{N_b}]$ back to $T{\times}H{\times}W$ token order (inverse of Stage 2 partition).
      \RETURN $\mathbf{O}$
      \end{algorithmic}
      \end{algorithm}

\subsection{Intra-Block Information Loss and Proof of Lagrange Multipliers}
\label{sec: proof_lagrange}

\noindent\textbf{Derivation of Intra-Block Information Loss.}
In block sparse attention, each block of shape $b_T \times b_H \times b_W$ is represented by the mean-pooled feature of its tokens. Since the variation indicator $g_d$ is computed from features already averaged over the other two axes (Eq.~\ref{eq:video_variance}), it captures the marginal variation along axis $d$ alone. 
We define intra-block information loss as the sum of three independent per-axis terms, each measuring the information loss from collapsing $b_d$ positions along axis $d$ into a single representative. 
We derive each per-axis term by combining a positional deviation factor that captures the geometric spread of tokens around the block center (depending solely on the block edge length $b_d$) with a content variation factor that measures how rapidly features change along that axis (driven by $g_d$).
We first derive the positional deviation factor. Mean-pooling collapses $b_d$ tokens along axis $d$ into a single representative located at the block center $\bar{i} = (b_d+1)/2$. Each token at position $i$ is displaced from this center by $(i - \bar{i})$, and the squared displacement $(i - \bar{i})^2$ quantifies how much information that token loses after being replaced by the block average. Intuitively, a longer edge (larger $b_d$) spreads more tokens farther from the center, amplifying the total information loss. The total squared positional deviation is:
\begin{equation}\small
\label{eq:ssd_derivation}
\begin{aligned}
    \sum_{i=1}^{b_d} \left(i - \frac{b_d+1}{2}\right)^2 = \frac{b_d(b_d^2 - 1)}{12},
\end{aligned}
\end{equation}
where the closed-form follows from $\sum_{i=1}^{b_d} i = \frac{b_d(b_d+1)}{2}$ and $\sum_{i=1}^{b_d} i^2 = \frac{b_d(b_d+1)(2b_d+1)}{6}$.
Here, $d \in \{T, H, W\}$ denotes the axis. 
In a 3D block, each position $i$ along axis $d$ corresponds not to a single token but to a slice of $\frac{b_Tb_Hb_W}{b_d} = \frac{B(m)}{b_d}$ tokens spanned by the other two axes (e.g., $b_H b_W$ tokens for $d{=}T$). All share the same displacement $(i - \bar{i})$. Thus, the total squared positional deviation over all tokens of the block is $\frac{B(m)}{b_d}\cdot\frac{b_d(b_d^2-1)}{12} = \frac{B(m)(b_d^2-1)}{12}$.
Regarding the content variation factor, not all axes contribute equally to the information loss. 
The variation indicator $g_d$ captures how rapidly content varies along axis $d$, combining both video channel-wise variance and audio-visual coupling. Axes with larger $g_d$ incur greater feature deviation. 
We weight the positional deviation by $g_d$, since $g_d$ is built from variances and thus already reflects the squared content variation along each axis.
Combining both factors and summing over three axes independently, the total intra-block information loss is (dropping $\frac{B(m)}{12}$, which is fixed for a given zone):
\begin{equation}\small
\label{eq:total_information_loss}
\begin{aligned}
    \mathcal{J}(b_T, b_H, b_W) = g_T \cdot (b_T^2-1) + g_H \cdot (b_H^2-1) + g_W \cdot (b_W^2-1),
\end{aligned}
\end{equation}
We note that this derivation assumes approximately linear feature variation within each block, which may not hold at sharp object boundaries where features change abruptly. In practice, this approximation remains effective for two reasons. First, the macro-zone size $8^3$ is small enough that most zones contain locally smooth content, and the few zones straddling object boundaries still benefit from anisotropic partitioning that aligns the shortest edge with the boundary direction, limiting the number of tokens affected by the discontinuity. Second, the discrete candidate set $\mathcal{C}$ contains only 16 shapes, so moderate deviations from the continuous optimum map to the same quantized shape, making the final selection inherently robust to approximation errors in the variation indicator estimates. This formulation also provides a simple and direct implementation.

\noindent\textbf{Optimal Shape via Lagrange Multipliers.}
Given the intra-block information loss $\mathcal{J}$, we solve the constrained optimization $\min_{b_T, b_H, b_W} \mathcal{J}(b_T, b_H, b_W)$ subject to $b_T \cdot b_H \cdot b_W = B(m)$.
We introduce a Lagrange multiplier $\eta$ and form the Lagrangian:
\begin{equation}\small
\label{eq:lagrange_multipliers_formulation}
\begin{aligned}
    \mathfrak{L}(b_T, b_H, b_W, \eta) = g_T (b_T^2-1) + g_H (b_H^2-1) + g_W (b_W^2-1) - \eta\,(b_T b_H b_W - B(m)),
\end{aligned}
\end{equation}
Taking partial derivatives and setting them to zero:
\begin{equation}\small
\label{eq:lagrange_multipliers_formulation_derivatives}
\begin{aligned}
    \frac{\partial \mathfrak{L}}{\partial b_T} &= 2g_T b_T - \eta\, b_H b_W = 0, \\
    \frac{\partial \mathfrak{L}}{\partial b_H} &= 2g_H b_H - \eta\, b_T b_W = 0, \\
    \frac{\partial \mathfrak{L}}{\partial b_W} &= 2g_W b_W - \eta\, b_T b_H = 0,
\end{aligned}
\end{equation}
Dividing the first equation by the second eliminates $\eta$:
\begin{equation}\small
\begin{aligned}
    \frac{g_T b_T}{g_H b_H} = \frac{b_H b_W}{b_T b_W} = \frac{b_H}{b_T} \quad \rightarrow \quad g_T b_T^2 = g_H b_H^2,
\end{aligned}
\end{equation}
Similarly, $g_T b_T^2 = g_W b_W^2$. By the same pairwise argument, $g_T b_T^2 = g_H b_H^2 = g_W b_W^2$, so $g_d b_d^2$ is identical for all three axes. Rearranging gives $b_d^{*2} \propto g_d^{-1}$, and thus $b_d^* \propto g_d^{-1/2}$ for $d \in \{T, H, W\}$. The axis with the largest variation indicator receives the shortest edge (finest partitioning), keeping blocks internally coherent.

\noindent\textbf{Discrete Decisions and Gradient Flow.}
Both the shape assignment (Eq.~\ref{eq:final_decision}) and the hybrid block selection (Eq.~\ref{eq:hybrid_block_selection}) are discrete and non-differentiable. We treat every statistic that drives them ($\rho$, $\mathbf{g}_m$, and the block scores $\mathbf{P}_i$) as a detached constant, so gradients propagate only through the sparse attention over the selected blocks. This is exactly the regime in which prior trainable block sparse attention operates~\citep{lu2026moba, yuan2025native, wu2025vmoba, team2025longcat}, where the hard Top-$k$ over blocks is equally non-differentiable, and those models train stably at scale. 

\subsection{Dynamic Block Sparse Attention Details}
\label{sec: dynamic_block_sparse_attn_details}

Algorithm \ref{alg:prism_alignmth} depicts the dynamic block sparse attention details.
The computation proceeds in five stages:
(1) Stage 1 computes the dynamic block shape for each macro-zone independently per attention head and per layer. The video channel-wise variance guidance $r_{v,d}$ and the audio-to-video cross-attention norm guidance $\bar{a} \cdot \hat{v}_{a,d}$ are combined into the variation indicator $\mathbf{g}_m$, which drives the information density estimation and the Lagrange-derived shape quantization (Eq.~\ref{eq:quantize_solution} and Eq.~\ref{eq:final_decision}). 
(2) Stage 2 rearranges tokens within each zone into a block-contiguous memory layout according to the selected shape. Since different zones may have different block shapes with block sizes in $\{64, 128, 256\}$, the Triton kernel uses a fixed tile size of 64 tokens as the atomic compute unit. A $B{=}64$ block maps to exactly one tile. A $B{=}128$ block is stored as 2 contiguous 64-token tiles, and a $B{=}256$ block as 4 contiguous tiles. The inner loop of the attention kernel iterates over these tiles sequentially within each selected block. This design avoids variable-size kernel launches and ensures that all blocks, regardless of their 3D shape, map to coalesced memory access patterns.
(3) Stage 3 computes block-level representatives via mean pooling and performs the hybrid Top-$k$/Top-$p$ selection. The mean-pooled query and key representatives $\bar{\mathbf{Q}}_i, \bar{\mathbf{K}}_j$ are used to compute block relevance scores, which are then passed through softmax for Top-$p$ cumulative thresholding. The selected block indices $\mathcal{S}_i$ are stored as a sparse index tensor for the downstream kernel.
(4) Stage 4 executes the actual sparse attention using an online softmax. For each query block $i$, the full query tokens $\mathbf{Q}_i$ (not the mean-pooled representative) are loaded into SRAM as the resident tile. The kernel then iterates only over the selected key blocks $j \in \mathcal{S}_i$, loading each $\mathbf{K}_j, \mathbf{V}_j$ tile from HBM. The per-row running maximum $\boldsymbol{\xi}_i$, running sum $\boldsymbol{\zeta}_i$, and running output $\mathbf{O}_i$ are maintained in registers and updated via the rescaling trick $e^{\boldsymbol{\xi}_{\text{old}} - \boldsymbol{\xi}_{\text{new}}}$ to ensure numerical stability without requiring a separate pass for the softmax denominator. After iterating over all selected blocks, the output is normalized by $\boldsymbol{\zeta}_i$ and written back to HBM. Since the kernel skips all non-selected blocks entirely, the wall-clock time scales with the average number of selected blocks $|\mathcal{S}_i|$ rather than the total $N_b$, yielding the sparsity-proportional speedup.
(5) Stage 5 scatters the block-contiguous output back to the original $T{\times}H{\times}W$ token order via an inverse permutation of the Stage 2 rearrangement.

\subsection{More Implementation Details}
\label{sec: more_implementation_details}

We use the AdamW optimizer with parameters $\beta_{1}=0.9$, $\beta_{2}=0.999$. The model is trained at bf16 precision, equipped with FSDP for distributed data-parallel training.
The learning rate is $1e-5$.
We set Top-$k$ (95\%), Top-$p$ (0.2), $\tau_{128}$=0.5, and $\tau_{256}$=0.25.
The backbone (dual-branch DiT) is initialized by MOVA-720p~\citep{team2026mova}, and our dynamic block sparse attention can directly inherit the full attention weights of MOVA-720p.
We only replace the original video self-attention with our dynamic block sparse attention while keeping all other components of the DiT entirely unchanged. The audio self-attention remains full attention, as the audio token sequence is inherently short (typically under 1K tokens after VAE compression) and introduces negligible computational overhead, making sparse attention unnecessary. The cross-modal attention modules (text-to-video, text-to-audio, audio-to-video cross-attention, and video-to-audio cross-attention) also remain unmodified for two reasons. First, previous works~\citep{zhang2025vsa, wu2025vmoba, zhang2026spargeattention2, yang2026sparse_2} have empirically observed that applying sparse attention to cross-modal attention causes significant quality degradation. Second, the key or query tokens in these cross-modal attention modules originate from text or audio modalities whose token sequences are much shorter than the video sequence, so their computational cost is already minimal.
During training, only the attention modules are trainable while all other parameters remain frozen.
We train our model for 5 epochs on 160 NVIDIA H800 GPUs, with one batch per node and Sequence Parallelism (SP=8) across the eight GPUs within each node.
All competitors and ablated variants are instead trained until their own loss converges.
Regarding the audio-to-video cross-attention norm guidance, every DiT layer applies its video self-attention before its audio-to-video cross-attention, so the $\hat{\mathbf{h}}$ of the current layer does not yet exist when the block shape has to be decided. Thus, we take the $\hat{\mathbf{h}}$ cached from the preceding layer, and the first layer uses no audio-to-video cross-attention norms at all, where $g_d$ reduces to $r_{v,d}$.

For the attention implementation, the audio self-attention and all cross-modal attention modules use FlashAttention-3 with bf16 precision. 
Our dynamic block sparse attention for the video self-attention is implemented as a custom Triton~\citep{tillet2019triton} kernel. We use Triton 3.3.1 with CUDA 12.8. 
The kernel is compiled with \texttt{num\_warps=8} and \texttt{num\_stages=3} for pipelined global memory loads. The BLOCK\_M (query tile) and BLOCK\_N (key tile) are both set to 64 tokens, matching the minimum block size in our candidate set $\mathcal{C}_{64}$. The head dimension BLOCK\_D is set to 128, matching the per-head dimension of the MOVA backbone. Each Triton program instance processes one query tile and iterates over the selected key tiles indexed by $\mathcal{S}_i$ from Stage 3. 
The Stage 1 shape assignment and Stage 2 token rearrangement are implemented as separate lightweight Triton kernels that execute prior to the main attention kernel. 
The Stage 2 rearrangement uses a precomputed permutation index tensor to scatter tokens from the original $T{\times}H{\times}W$ layout into the block-contiguous layout, and Stage 5 applies the inverse permutation. 

\subsection{More Comparison Results}
\label{sec: more_comparison_results}

\begin{figure*}[t!]
\begin{center}
\includegraphics[width=1.0\linewidth]{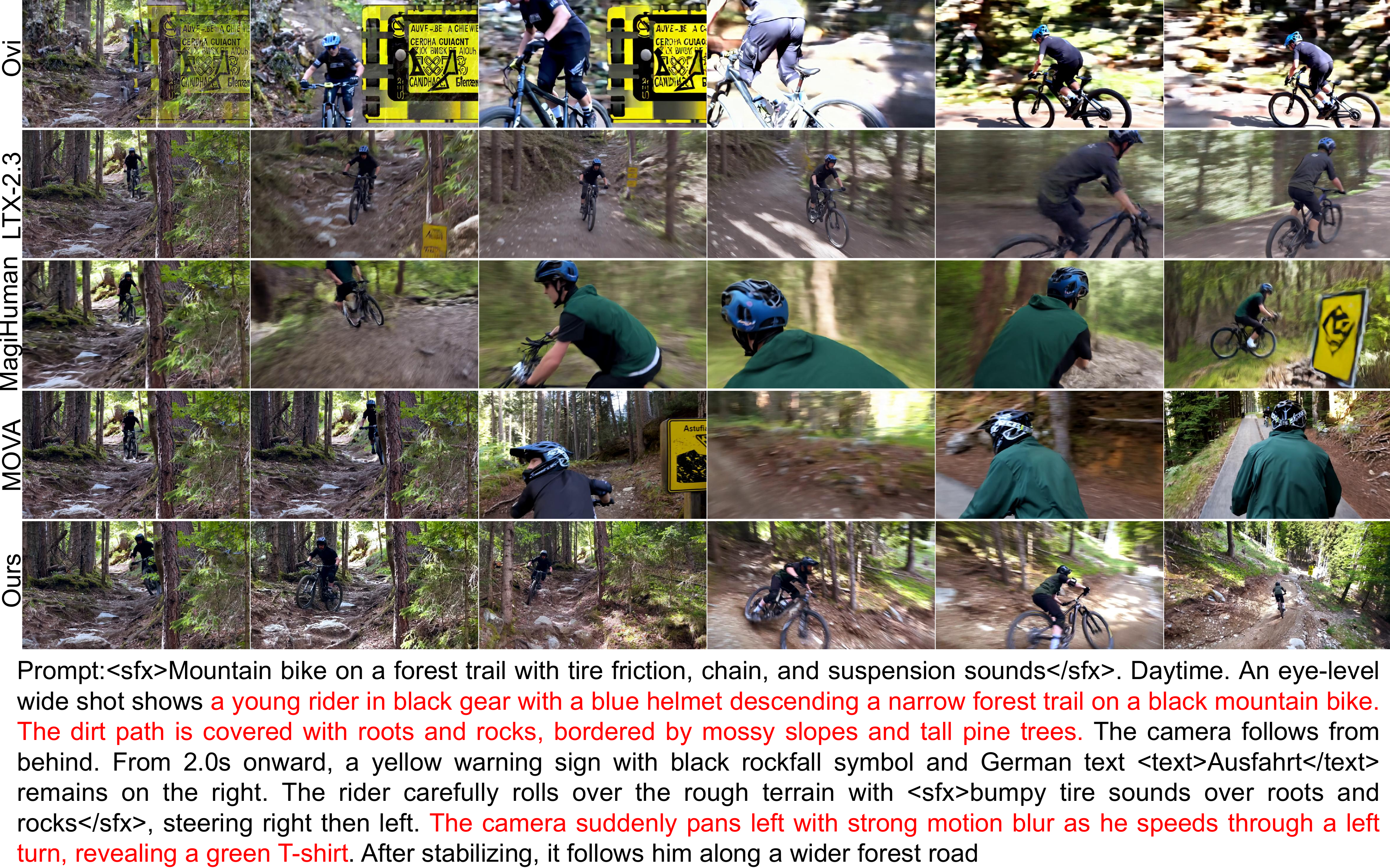}
\end{center}
\vspace{-0.3cm}
   \caption{More comparison results (1/4). Please refer to the demo video for audio.
   }
\label{fig:supp_comparison_1}
\vspace{-0.4cm}
\end{figure*}

\begin{figure*}[t!]
\begin{center}
\includegraphics[width=1.0\linewidth]{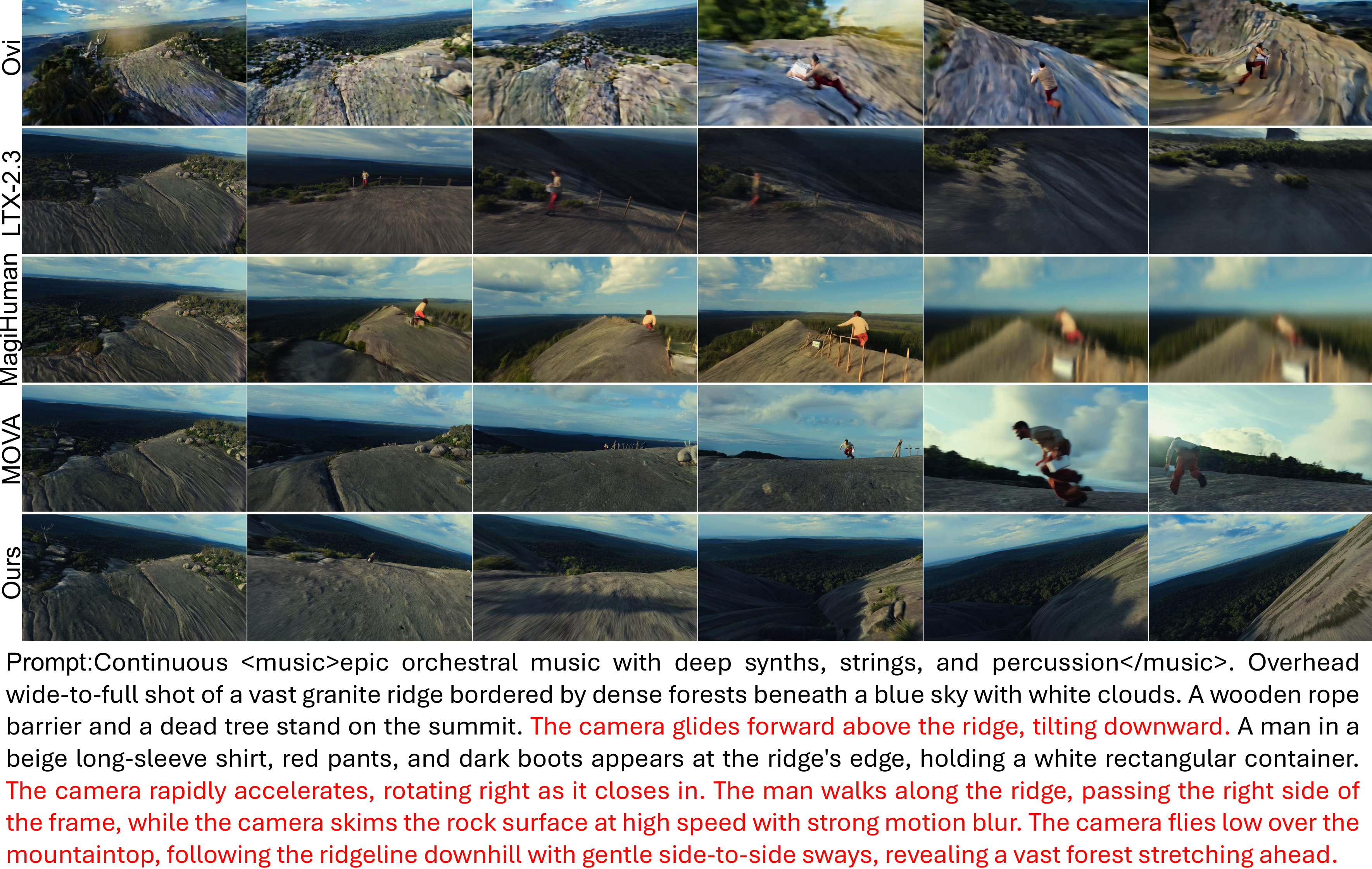}
\end{center}
\vspace{-0.2cm}
   \caption{More comparison results (2/4). Please refer to the demo video for audio.
   }
\label{fig:supp_comparison_2}
\vspace{-0.4cm}
\end{figure*}

\begin{figure*}[t!]
\begin{center}
\includegraphics[width=0.83\linewidth]{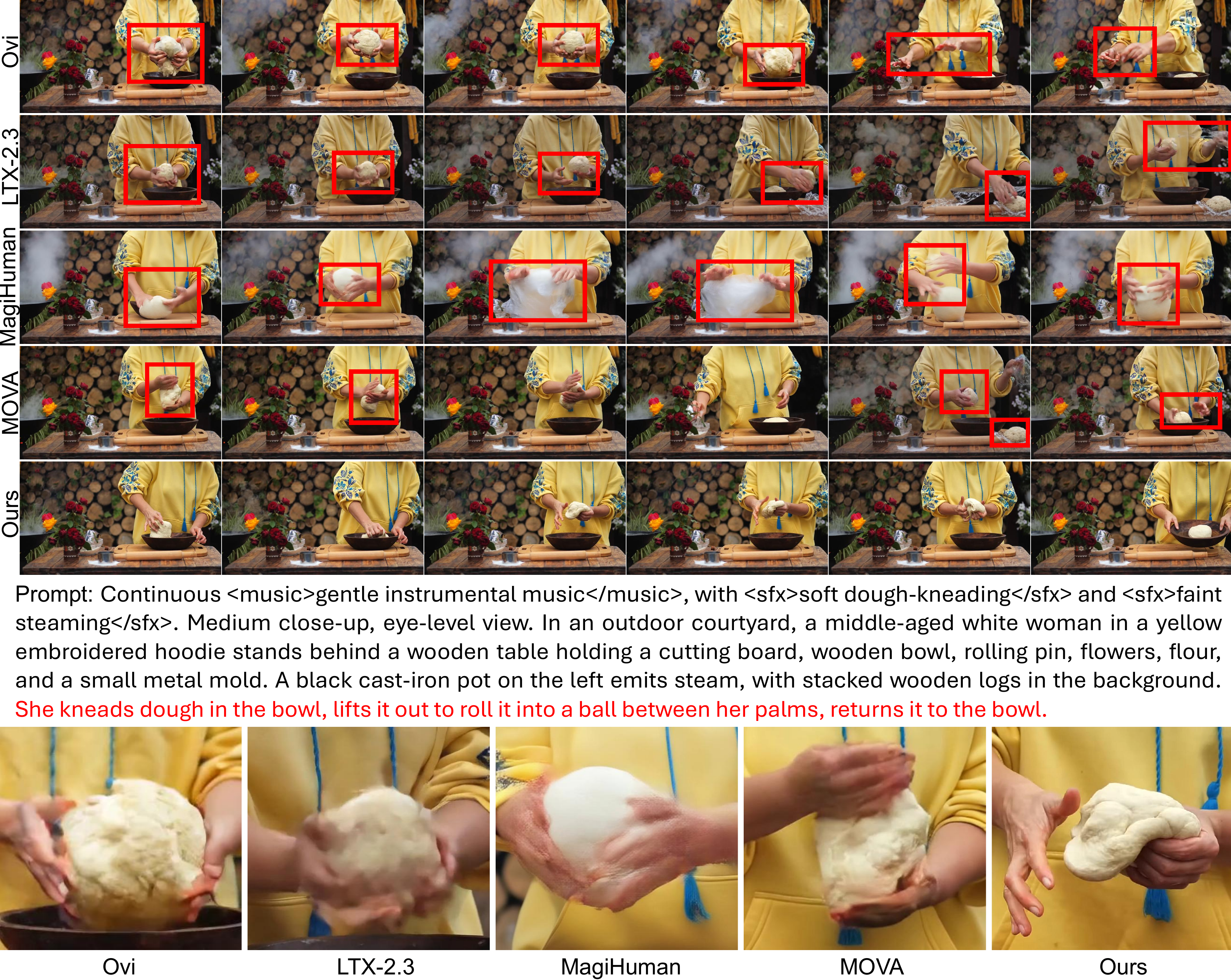}
\end{center}
\vspace{-0.3cm}
   \caption{More comparison results (3/4). Please refer to the demo video for audio.
   }
\label{fig:supp_comparison_3}
\vspace{-0.4cm}
\end{figure*}

\begin{figure*}[t!]
\begin{center}
\includegraphics[width=0.83\linewidth]{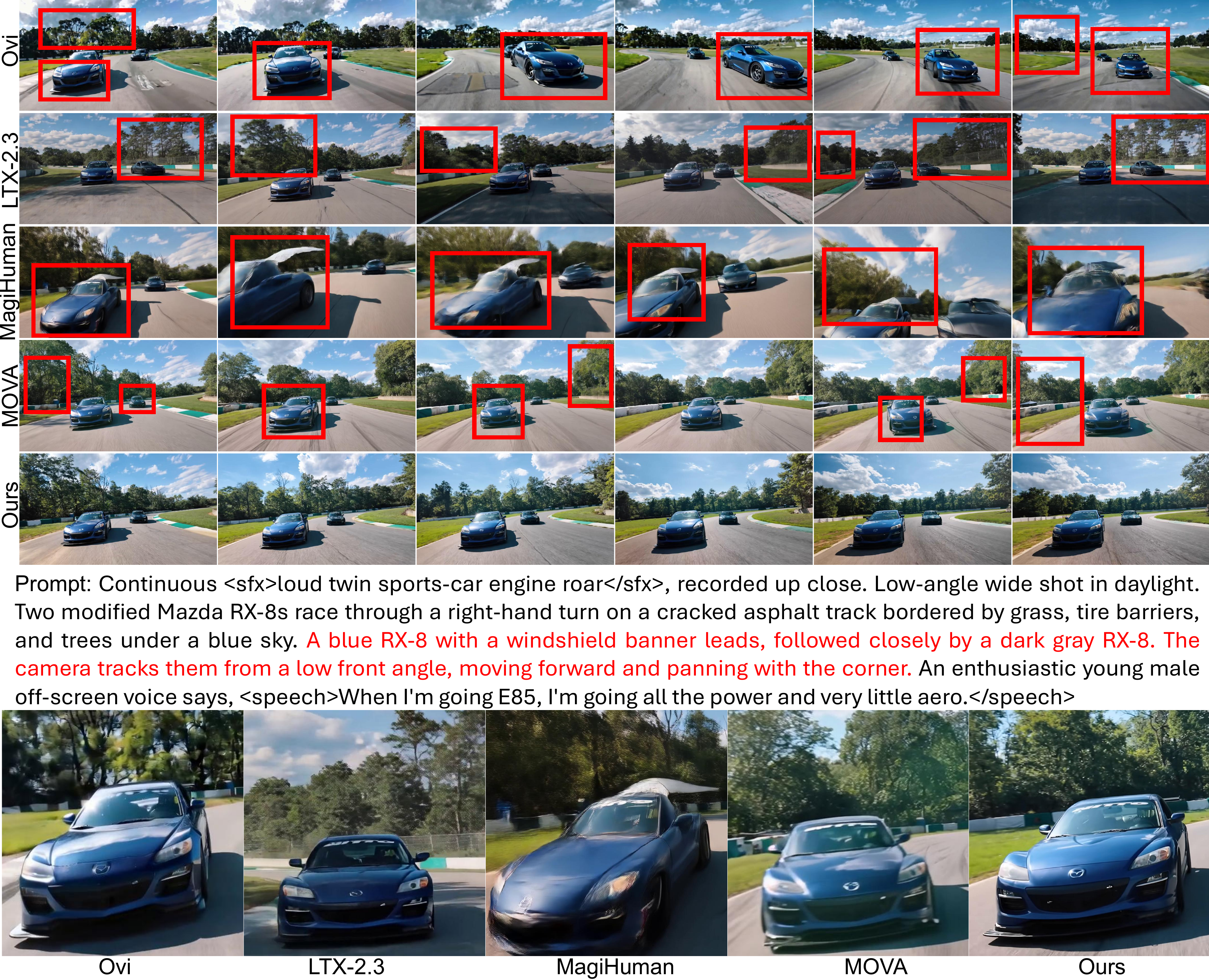}
\end{center}
\vspace{-0.3cm}
   \caption{More comparison results (4/4). Please refer to the demo video for audio.
   }
\label{fig:supp_comparison_4}
\vspace{-0.4cm}
\end{figure*}

Fig. \ref{fig:supp_comparison_1}, Fig. \ref{fig:supp_comparison_2}, Fig. \ref{fig:supp_comparison_3}, Fig. \ref{fig:supp_comparison_4} show additional comparison results between our model and state-of-the-art open-source models. We can see that our Prism has the best performance.

\subsection{Commercial Model Comparison Results}
\label{sec: commercial_model_comparison_results}
\begin{figure*}[t!]
\begin{center}
\includegraphics[width=0.91\linewidth]{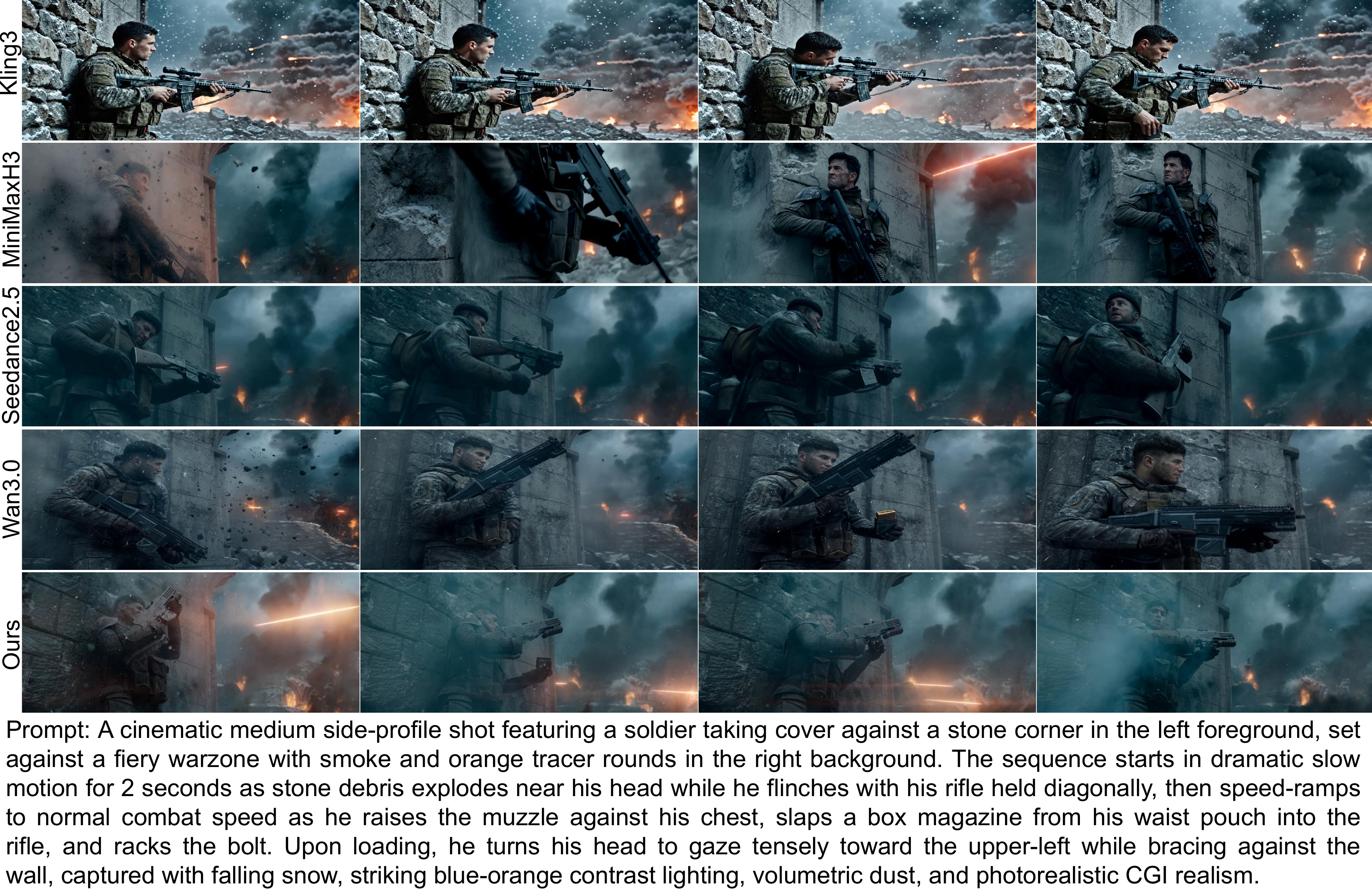}
\end{center}
\vspace{-0.4cm}
   \caption{Qualitative Comparisons with commercial models (1/2).
   }
\label{fig:commercial_comparison_1}
\vspace{-0.4cm}
\end{figure*}

\begin{figure*}[t!]
\begin{center}
\includegraphics[width=0.91\linewidth]{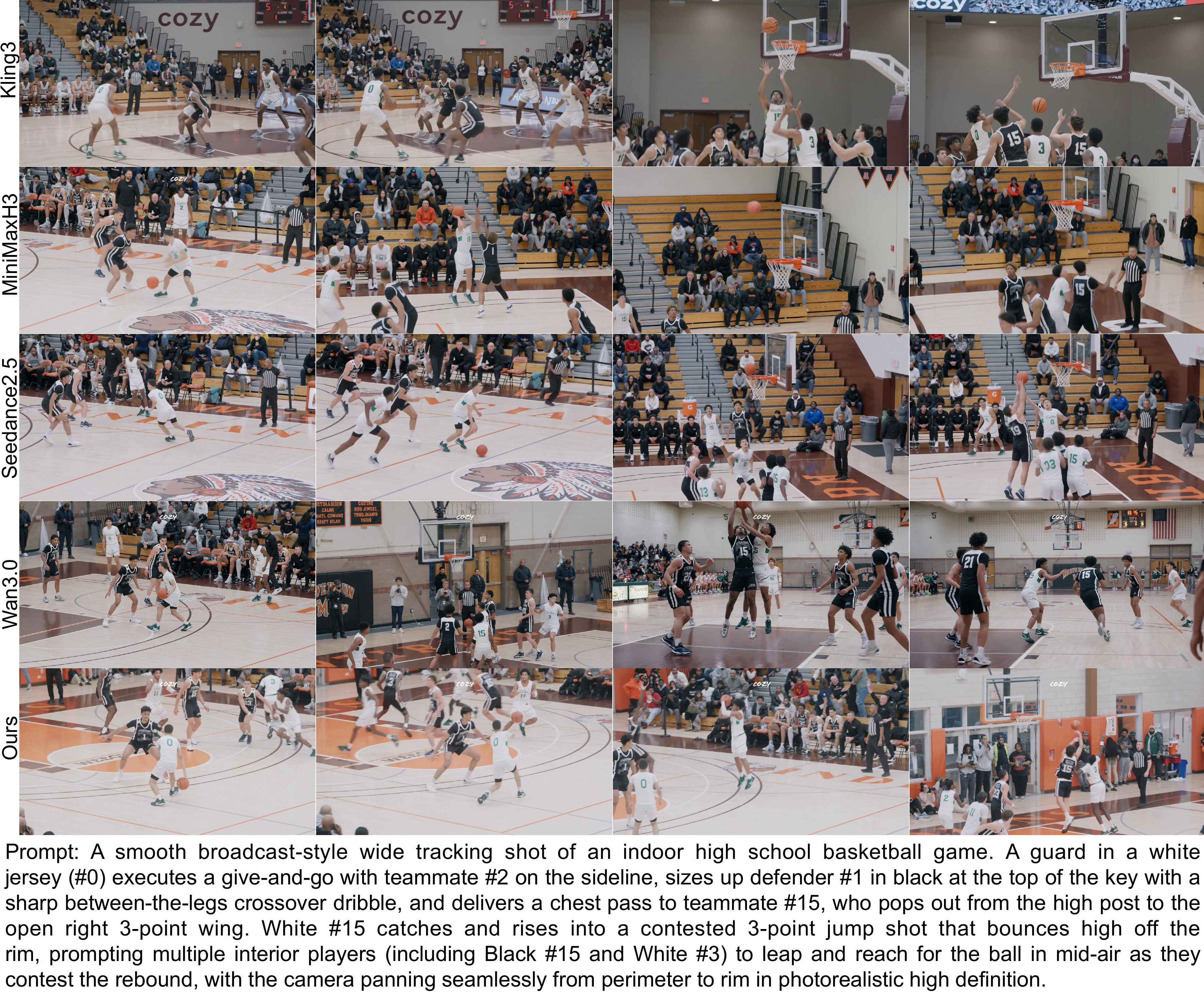}
\end{center}
\vspace{-0.4cm}
   \caption{Qualitative Comparisons with commercial models (2/2)
   }
\label{fig:commercial_comparison_2}
\vspace{-0.4cm}
\end{figure*}

Fig. \ref{fig:commercial_comparison_1}, Fig. \ref{fig:commercial_comparison_2}, and Table \ref{table:commercial_comparison} show the commercial model comparison results.
Kling3.0~\citep{kling3}, Seedance2.5~\citep{seedance25}, and Wan3.0~\citep{wan3} are fully closed-source.
MiniMax-H3~\citep{minimaxh3} is partially open-source (30B dense model) and generates 2K videos through a latent-space regeneration pipeline that first synthesizes at 720p and then upscales, similar to LTX-2.3. Our model (16B) is the only open-source method that performs native 2K training with dynamic sparse attention.
MiniMax-H3 surpasses Prism despite its upsampling pipeline, but with a 30B backbone and far larger data. At matched backbone scale, Prism still outperforms LTX-2.3.
All commercial models substantially outperform Prism across all metrics, benefiting from much larger model scales, larger-scale training data, and extensive RLHF post-training.
Despite the significant gap, Prism narrows the distance between open-source and commercial models in visual detail fidelity and motion quality. 
The gap is larger in audio quality, where commercial models benefit from dedicated audio post-processing pipelines and larger paired training data. 
These results suggest that native high-resolution training with dynamic sparse attention is a promising direction for closing the gap with commercial models, as the quality advantages of Prism over open-source competitors (Table~\ref{table:quantitative_comparison_2kbench}) demonstrate that the sparse attention mechanism itself is not the bottleneck. 
Scaling model parameters, training data, and incorporating RLHF are complementary directions that could further improve Prism.

\begin{table}[t!]\small
\caption{Comparison with industry-leading models on 2K-Bench. All commercial models are evaluated using their native 2K/1080p generation APIs.
}
\vspace{-0.05in}
\begin{center}
\renewcommand\arraystretch{1.1}
\scalebox{0.84}{
\begin{tabular}{lcccccccc}
\toprule
Model     & AQ$\uparrow$ & TF$\uparrow$ & PQ$\uparrow$ & CU$\uparrow$ & DeSync$\downarrow$ & cpCER$\downarrow$ & MANIQA$\uparrow$ & MotionQ$\uparrow$ \\ \midrule
Kling3.0~\citep{kling3}       & 0.65 & 0.986 & 7.93 & 7.61 & 0.52 & 0.143 & 0.456 & 0.91 \\
MiniMax-H3~\citep{minimaxh3}  & 0.66 & 0.983 & 8.01 & 7.72 & 0.54 & 0.148 & 0.449 & 0.90 \\
Seedance2.5~\citep{seedance25} & 0.71 & 0.989 & 8.27 & 7.98 & 0.44 & 0.112 & 0.471 & 0.96 \\
Wan3.0~\citep{wan3}           & 0.74 & 0.991 & 8.38 & 8.11 & 0.40 & 0.103 & 0.482 & 0.96 \\ \midrule
Ours (16B)                    & 0.61 & 0.982 & 7.69 & 7.34 & 0.63 & 0.187 & 0.438 & 0.89 \\ \bottomrule
\end{tabular}
}
\end{center}
\label{table:commercial_comparison}
\vspace{-0.10in}
\end{table}

\begin{table}[t!]\small
\caption{Ablation study on dynamic block shape. $r_{v,d}$ is the video channel-wise variance guidance. $\bar{a}$ is the audio coupling gate. $\hat{v}_{a,d}$ is the audio directional variance. Global $\sigma^2_{v,d}$ (\textit{w/o} ch.) replaces the per-channel variance in Eq.~\ref{eq:video_variance} with global feature variance computed across all channels jointly. $\sigma^2_{q,d}$/$\sigma^2_{k,d}$ replace value features with query/key features for variance computation.
}
\vspace{-0.05in}
\begin{center}
\renewcommand\arraystretch{1.05}
\scalebox{0.65}{
\begin{tabular}{l|l|cccccccccc}
\toprule
Category & Setting     & AQ$\uparrow$ & TF$\uparrow$ & PQ$\uparrow$ & CU$\uparrow$ & DeSync$\downarrow$ & cpCER$\downarrow$ & MANIQA$\uparrow$ & MotionQ$\uparrow$ & Training Time$\downarrow$ & Infer GPU Mem$\downarrow$ \\ \midrule
\multirow{9}{*}{Block Shape} 
& Random Shape          & 0.40 & 0.894 & 6.73 & 6.18 & 1.03 & 0.372 & 0.339 & 0.46 & 8.7min & 36.2G \\
& Fixed $4^3$            & 0.55 & 0.961 & 7.43 & 6.99 & 0.72 & 0.234 & 0.413 & 0.77 & 12.2min & 42.3G \\
& Fixed $8^3$            & 0.43 & 0.906 & 6.86 & 6.32 & 1.02 & 0.367 & 0.352 & 0.51 & 6.8min & 33.4G \\
& Only $\mathcal{C}_{64}$       & 0.57 & 0.965 & 7.49 & 7.08 & 0.70 & 0.221 & 0.419 & 0.80 & 13.5min & 44.1G \\
& Only $\mathcal{C}_{128}$      & 0.54 & 0.953 & 7.36 & 6.91 & 0.76 & 0.251 & 0.406 & 0.74 & 10.9min & 39.3G \\
& Only $\mathcal{C}_{256}$      & 0.49 & 0.934 & 7.17 & 6.64 & 0.86 & 0.298 & 0.388 & 0.64 & 8.4min & 35.2G \\
& $\mathcal{C}_{64}$+$\mathcal{C}_{256}$  & 0.55 & 0.967 & 7.38 & 6.94 & 0.67 & 0.226 & 0.410 & 0.78 & 11.2min & 39.1G \\
& $\mathcal{C}_{64}$+$\mathcal{C}_{128}$  & 0.58 & 0.958 & 7.56 & 7.19 & 0.72 & 0.213 & 0.428 & 0.84 & 12.1min & 42.4G \\
& Macro-Zone $16^3$           & 0.52 & 0.941 & 7.26 & 6.78 & 0.81 & 0.264 & 0.396 & 0.68 & 11.9min & 41.2G \\ \midrule
\multirow{7}{*}{Guidance} 
& \textit{w/o} Audio Guidance ($\bar{a}\!\cdot\!\hat{v}_{a,d}$) & 0.57 & 0.969 & 7.46 & 6.97 & 0.79 & 0.263 & 0.420 & 0.78 & 9.2min & 36.5G \\
& \textit{w/o} $\bar{a}$      & 0.58 & 0.972 & 7.52 & 7.10 & 0.71 & 0.231 & 0.423 & 0.81 & 10.3min & 37.8G \\
& \textit{w/o} $\hat{v}_{a,d}$     & 0.59 & 0.974 & 7.57 & 7.15 & 0.69 & 0.222 & 0.427 & 0.83 & 10.0min & 37.3G \\
& \textit{w/o} $r_{v,d}$           & 0.53 & 0.943 & 7.31 & 6.82 & 0.73 & 0.219 & 0.399 & 0.70 & 8.4min & 36.4G \\
& Global $\sigma^2_{v,d}$ (\textit{w/o} ch.) & 0.55 & 0.956 & 7.38 & 6.93 & 0.74 & 0.242 & 0.407 & 0.74 & 9.3min & 37.2G \\
& $\sigma^2_{q,d}$ (use Q) & 0.56 & 0.962 & 7.44 & 7.01 & 0.72 & 0.236 & 0.411 & 0.76 & 10.6min & 38.0G \\
& $\sigma^2_{k,d}$ (use K) & 0.58 & 0.968 & 7.51 & 7.09 & 0.69 & 0.224 & 0.418 & 0.80 & 10.6min & 38.0G \\ \midrule
\multirow{5}{*}{$\rho$ Design} 
& Spatial Only     & 0.58 & 0.949 & 7.48 & 7.05 & 0.73 & 0.231 & 0.421 & 0.74 & 10.3min & 38.1G \\
& Temporal Only    & 0.55 & 0.967 & 7.40 & 6.92 & 0.71 & 0.241 & 0.407 & 0.81 & 10.1min & 37.9G \\
& L2 Norm $\rho$          & 0.56 & 0.952 & 7.37 & 6.88 & 0.76 & 0.249 & 0.405 & 0.72 & 10.2min & 38.1G \\
& $\bar{a}(m)$ as $\rho$  & 0.53 & 0.946 & 7.33 & 6.82 & 0.69 & 0.214 & 0.396 & 0.70 & 9.0min & 37.9G \\
& Random $\rho$          & 0.51 & 0.938 & 7.22 & 6.71 & 0.83 & 0.276 & 0.391 & 0.65 & 9.9min & 38.3G \\ \midrule
& Ours      & \textbf{0.61} & \textbf{0.982} & \textbf{7.69} & \textbf{7.34} & \textbf{0.63} & \textbf{0.187} & \textbf{0.438} & \textbf{0.89} & 10.6min & 38.0G \\ \bottomrule
\end{tabular}
}
\end{center}
\label{table:ablation_dynamic_block}
\vspace{-0.10in}
\end{table}

\begin{table}[t!]\small
      \caption{Ablation study on dynamic block shape. Shape Mapping ablates the quantization in Eq.~\ref{eq:quantize_solution}, where $\alpha$ controls $b^*_d \propto g_d^{-\alpha}$. Heads/Layers ablates whether block shape is computed independently per attention head and per layer. $\tau_{128}/\tau_{256}$ ablates the thresholds for block shape assignment.
      }
      \vspace{-0.05in}
      \begin{center}
      \renewcommand\arraystretch{1.05}
      \scalebox{0.62}{
      \begin{tabular}{l|l|cccccccccc}
      \toprule
      Category & Setting     & AQ$\uparrow$ & TF$\uparrow$ & PQ$\uparrow$ & CU$\uparrow$ & DeSync$\downarrow$ & cpCER$\downarrow$ & MANIQA$\uparrow$ & MotionQ$\uparrow$ & Training Time$\downarrow$ & Infer GPU Mem$\downarrow$ \\ \midrule
      \multirow{7}{*}{Shape Mapping} 
      & $\alpha{=}1/3$         & 0.56 & 0.958 & 7.41 & 6.97 & 0.72 & 0.236 & 0.411 & 0.76 & 10.6min & 38.0G \\
      & $\alpha{=}1$           & 0.58 & 0.965 & 7.53 & 7.14 & 0.68 & 0.214 & 0.421 & 0.82 & 10.6min & 38.0G \\
      & $\alpha{=}2$           & 0.54 & 0.949 & 7.31 & 6.86 & 0.76 & 0.248 & 0.402 & 0.71 & 10.6min & 38.0G \\
      & Softmax          & 0.57 & 0.964 & 7.46 & 7.03 & 0.69 & 0.227 & 0.414 & 0.77 & 10.6min & 38.0G \\
      & Rank             & 0.55 & 0.963 & 7.43 & 6.98 & 0.73 & 0.239 & 0.417 & 0.75 & 10.6min & 38.0G \\
      & Entropy          & 0.58 & 0.971 & 7.49 & 7.08 & 0.71 & 0.223 & 0.423 & 0.79 & 10.6min & 38.0G \\
      & Isotropic        & 0.53 & 0.946 & 7.28 & 6.83 & 0.78 & 0.254 & 0.401 & 0.70 & 10.6min & 38.0G \\ \midrule
      \multirow{3}{*}{Heads/Layers} 
      & Per-Layer, Shared Heads  & 0.57 & 0.966 & 7.48 & 7.06 & 0.71 & 0.228 & 0.415 & 0.78 & 10.6min & 38.0G \\
      & Per-Head, Shared Layers  & 0.55 & 0.953 & 7.38 & 6.91 & 0.76 & 0.246 & 0.404 & 0.72 & 10.6min & 38.0G \\
      & Global Fixed           & 0.52 & 0.942 & 7.24 & 6.79 & 0.79 & 0.261 & 0.397 & 0.68 & 10.6min & 38.0G \\ \midrule
      \multirow{4}{*}{$\tau_{128}/\tau_{256}$} 
      & 0.3\,/\,0.1            & 0.57 & 0.968 & 7.49 & 7.09 & 0.69 & 0.215 & 0.423 & 0.81 & 13.5min & 39.4G \\
      & 0.4\,/\,0.15           & 0.59 & 0.976 & 7.61 & 7.24 & 0.65 & 0.198 & 0.432 & 0.86 & 11.2min & 39.2G \\
      & 0.6\,/\,0.35           & 0.58 & 0.967 & 7.51 & 7.11 & 0.70 & 0.221 & 0.421 & 0.82 & 10.0min & 36.8G \\
      & 0.7\,/\,0.5            & 0.54 & 0.948 & 7.32 & 6.86 & 0.78 & 0.256 & 0.401 & 0.71 & 8.7min & 36.5G \\ \midrule
      & Ours ($\alpha{=}1/2$, 0.5/0.25)                   & \textbf{0.61} & \textbf{0.982} & \textbf{7.69} & \textbf{7.34} & \textbf{0.63} & \textbf{0.187} & \textbf{0.438} & \textbf{0.89} & 10.6min & 38.0G \\ \bottomrule
      \end{tabular}
      }
      \end{center}
      \label{table:ablation_shape_mapping}
      \vspace{-0.10in}
      \end{table}

\subsection{More Ablation on Dynamic Block Shape}
\label{sec: more_ablation_dynamic_block_shape}

\begin{figure*}[t!]
\begin{center}
\includegraphics[width=0.9\linewidth]{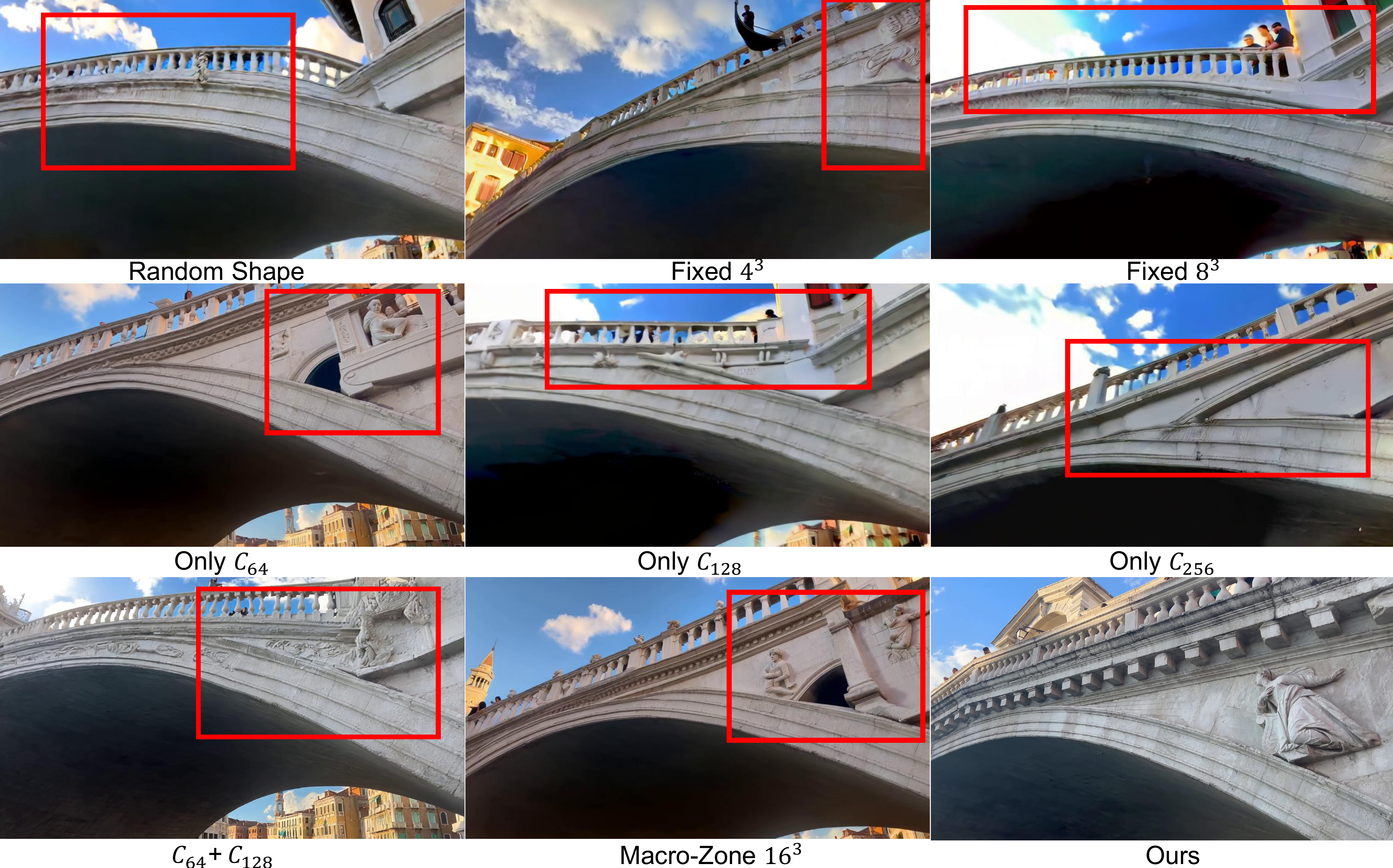}
\end{center}
\vspace{-0.4cm}
   \caption{Ablation study on block shape.
   }
\label{fig:ablation_block_shape}
\vspace{-0.35cm}
\end{figure*}

\begin{figure*}[t!]
\begin{center}
\includegraphics[width=0.88\linewidth]{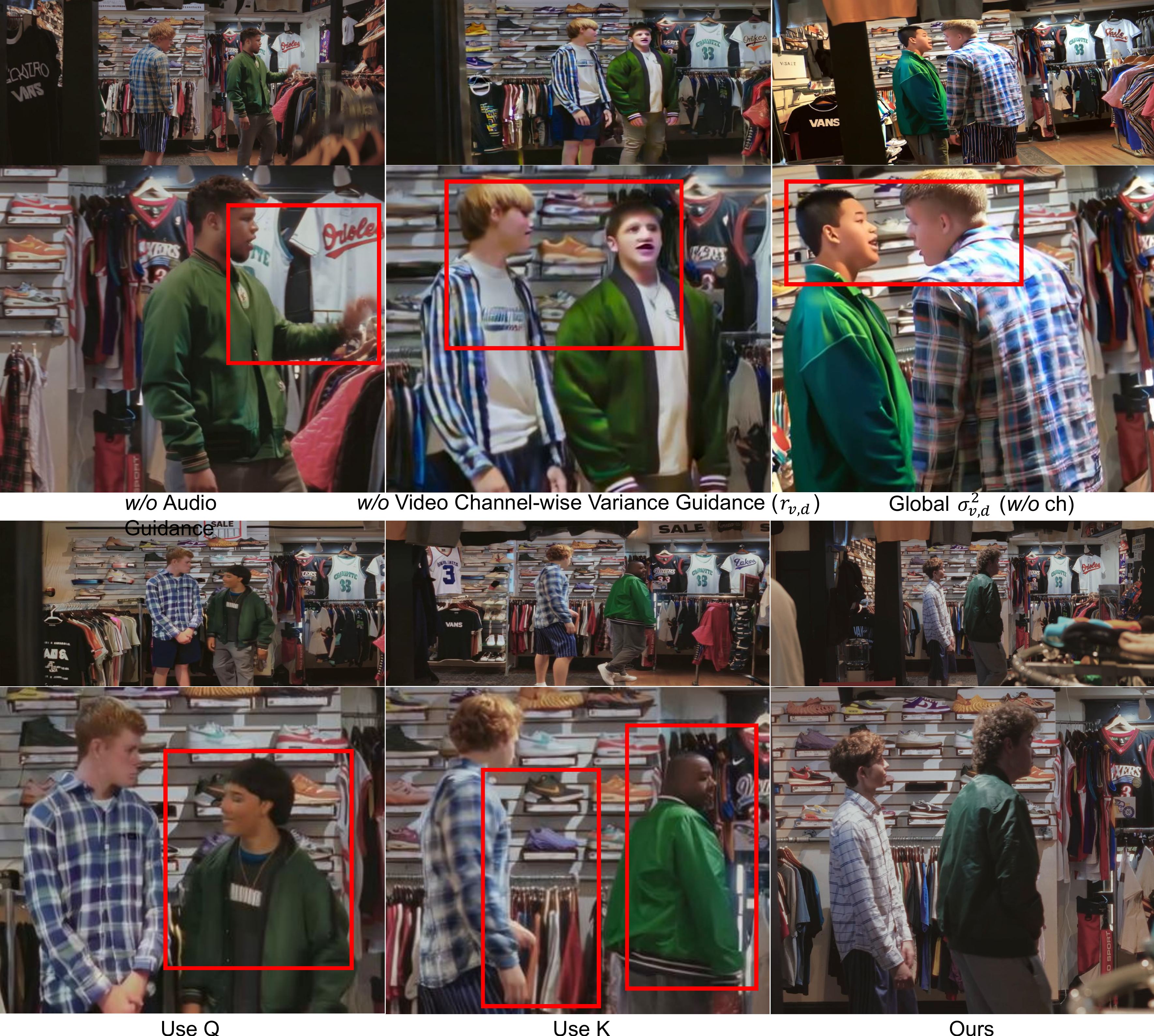}
\end{center}
\vspace{-0.4cm}
   \caption{Ablation study on guidance.
   }
\label{fig:ablation_guidance}
\vspace{-0.4cm}
\end{figure*}

We first ablate block shape, dynamic guidance, and information density in our dynamic block shape mechanism, as shown in Table \ref{table:ablation_dynamic_block}, Fig. \ref{fig:ablation_block_shape}, and Fig. \ref{fig:ablation_guidance}.
Spatial Only uses only the spatial channel variance term in Eq.~\ref{eq:zone_information_density}. Temporal Only uses only the temporal frame-difference variance. 
L2 Norm $\rho$ replaces $\rho_{\mathrm{raw}}$ with the mean L2 norm of value features within the zone. 
$\bar{a}(m)$ as $\rho$ directly uses the normalized audio coupling strength as the information density signal. Random $\rho$ randomly assigns $B$.
We have the following observations:
(1) For Block Shape, Fixed $8^3$ and Random Shape perform worst, confirming that content-agnostic block assignment corrupts block representatives. Only $\mathcal{C}_{64}$ achieves the best single-set quality but incurs the highest overhead, while our three-tier allocation surpasses it in quality at lower cost, reducing total block count without sacrificing quality on informative regions. 
$\mathcal{C}_{64}$+$\mathcal{C}_{128}$ achieves stronger MANIQA and MotionQ than $\mathcal{C}_{64}$+$\mathcal{C}_{256}$, since $\mathcal{C}_{128}$ blocks partition backgrounds more finely and preserve more spatial detail. However, $\mathcal{C}_{64}$+$\mathcal{C}_{256}$ achieves better DeSync, since $\mathcal{C}_{256}$ blocks aggressively compress redundant backgrounds into highly uniform representatives, making it easier for the block selection to distinguish informative regions from backgrounds and concentrate on audio-visual coupling.
(2) For Guidance, removing $r_{v,d}$ causes the most severe drop across visual metrics, confirming that video channel-wise variance is the primary driver of block shape adaptation. Notably, \textit{w/o} $r_{v,d}$ retains competitive cpCER since the audio-to-video cross-attention norm guidance alone suffices to steer block shapes toward sound-producing regions. Conversely, \textit{w/o} Audio Guidance preserves visual quality but degrades cpCER and DeSync significantly, revealing that video channel-wise variance guidance and audio-to-video cross-attention norm guidance play complementary roles. Using K features outperforms Q features, as K encodes token content semantically closer to V, while Q encodes query intent that is less informative for estimating directional content variation.
(3) For $\rho$ Design, Random $\rho$ performs worst, confirming that content-aware density estimation is indispensable. L2 Norm $\rho$ improves over Random $\rho$ but remains inferior to our $\rho_{\mathrm{raw}}$, as feature norms reflect magnitude rather than the directional variation that determines block boundaries. 
Spatial Only achieves high MANIQA but low MotionQ and TF, as temporally dynamic zones are left with coarse blocks. Temporal Only shows the opposite pattern. Using $\bar{a}(m)$ as $\rho$ achieves the best cpCER among $\rho$ ablations yet sacrifices MANIQA and MotionQ, as visually rich zones uncoupled from audio receive coarse blocks. These complementary failures confirm that both spatial and temporal variance are essential for capturing the full information density at 2K.

\begin{figure*}[t!]
\begin{center}
\includegraphics[width=0.84\linewidth]{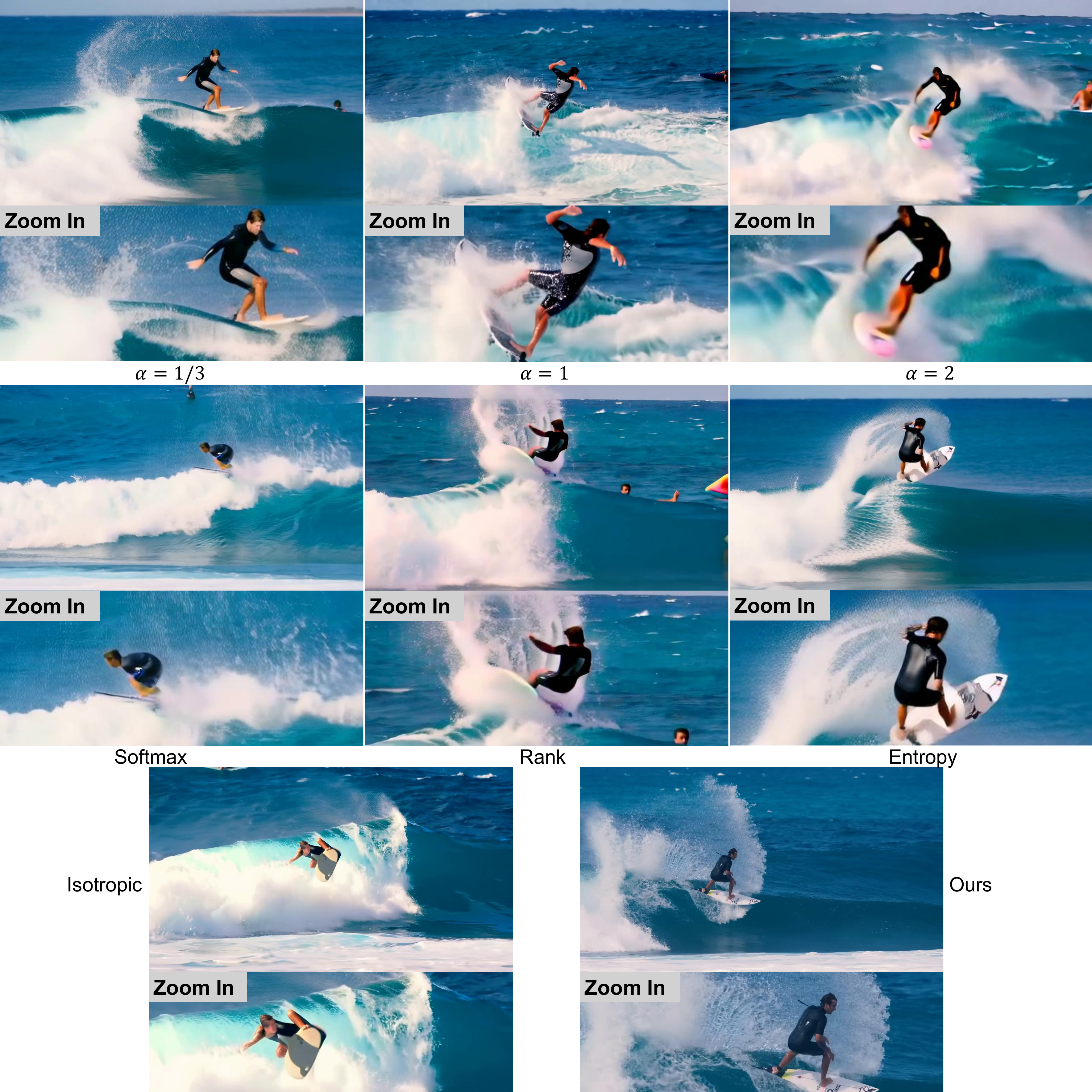}
\end{center}
\vspace{-0.3cm}
   \caption{Ablation study on different dynamic block shape mapping functions.
   }
\label{fig:ablation_mapping}
\vspace{-0.3cm}
\end{figure*}

We also ablate Eq.\ref{eq:quantize_solution}, attention heads/diffusion layers, and $\tau_{128}$/$\tau_{256}$ in our dynamic block shape mechanism, as shown in Table \ref{table:ablation_shape_mapping} and Fig. \ref{fig:ablation_mapping}.
$d \in \{T, H, W\}$.
Softmax computes $p_d {=} e^{-g_d}/(e^{-g_T}{+}e^{-g_H}{+}e^{-g_W})$ and sets $b_d \propto p_d$. Larger $g_d$ yields smaller $p_d$ and thus shorter $b_d$, giving a smooth nonlinear mapping from variance to edge lengths.
Rank sorts $g_T, g_H, g_W$ in descending order, maps the largest variance to the shortest edge, the smallest to the longest, and the middle to the middle, then selects the closest matching candidate from $\mathcal{C}_B$ (e.g., for $B{=}128$ with $g_T {>} g_H {>} g_W$, it selects $(2,8,8)$). This is a pure ordinal mapping that discards all magnitude information.
Entropy computes $p_d {=} g_d / (g_T{+}g_H{+}g_W)$ and sets $b_d \propto (1{-}p_d) \cdot B^{1/3}$, then quantizes to the nearest candidate in $\mathcal{C}_B$.
Isotropic always selects the candidate from $\mathcal{C}_B$ with the smallest edge variance (e.g., $(4,4,4)$ for $B{=}64$, $(4,4,8)$ for $B{=}128$), ignoring $\mathbf{g}_m$.
We can see that $\alpha$ governs how aggressively edge lengths respond to variance differences. $\alpha{=}1/2$ is the exact optimum derived from the Lagrange condition under the per-token mean squared deviation (Eq.~\ref{eq:total_information_loss}). Shortening one edge to capture fast-varying content along that axis inevitably lengthens the other edges under the fixed volume constraint $\prod_d b_d {=} B$, grouping more tokens along the other axes. 
$\alpha{=}1/2$ balances this reallocation against noise in the variance estimates. 
$\alpha{=}1/3$ responds too weakly, so even when one axis varies much faster than others, the block remains nearly isotropic and mixes semantically distinct tokens together. 
$\alpha{=}2$ responds too aggressively, making shape assignment unstable under the noisy variation indicator estimates during training. Among alternative mappings, Softmax saturates for large variance values, losing the ability to distinguish axes with strong variance. Entropy uses a linear mapping that cannot capture the $g_d^{-1/2}$ relationship. Rank discards magnitude entirely, and Isotropic ignores $\mathbf{g}_m$ entirely, indicating that anisotropic shape assignment is essential. 

\begin{table}[t!]\small
      \caption{Block shape proportion across DiT layers, denoised steps, and attention heads for the same 2K video-audio clip (a two-speaker dialogue with a sounding hand action). 
      Our MOVA~\citep{team2026mova} backbone follows the Wan2.2-style MoE design~\citep{wan2025}, with a high-noise (HN) and a low-noise (LN) expert of 40 layers and 40 heads each.
      Shapes are denoted $b_Tb_Hb_W$ (e.g., 824 means $b_T{=}8, b_H{=}2, b_W{=}4$)
      }
      \vspace{-0.05in}
      \begin{center}
      \renewcommand\arraystretch{1.02}
      \setlength{\tabcolsep}{2.2pt}
      \scalebox{0.78}{
      \begin{tabular}{l|c|c|ccccccc|c|cccccc|c|ccc}
      \toprule
      & & \multicolumn{8}{c|}{$\mathcal{C}_{64}$ (\%)} & \multicolumn{7}{c|}{$\mathcal{C}_{128}$ (\%)} & \multicolumn{4}{c}{$\mathcal{C}_{256}$ (\%)} \\ \cmidrule(lr){3-10}\cmidrule(lr){11-17}\cmidrule(lr){18-21}
      & Index & All & 444 & 824 & 842 & 248 & 284 & 428 & 482 & All & 288 & 828 & 882 & 448 & 484 & 844 & All & 488 & 848 & 884 \\ \midrule
      \multirow{5}{*}{Layer (HN)} 
      & 10  &  40.9 & 36.8 &  2.3 &  0.6 &  0.2 &  0.4 &  0.4 &  0.2 &  59.1 &  0.4 &  0.2 &  0.2 & 16.9 & 15.0 & 26.4 &   0.0 &   -- &   -- &   -- \\
      & 20  &  27.0 & 20.9 &  3.3 &  1.5 &  0.3 &  0.4 &  0.2 &  0.4 &  65.2 &  0.3 &  0.6 &  0.3 & 18.6 & 14.1 & 31.3 &   7.8 &  0.5 &  4.3 &  3.0 \\
      & 24  &  25.2 & 18.1 &  4.2 &  1.5 &  0.4 &  0.4 &  0.4 &  0.2 &  28.8 &  0.2 &  1.4 &  0.4 &  7.1 &  3.7 & 16.0 &  46.0 & 11.3 & 21.6 & 13.1 \\
      & 32  &  21.4 & 12.6 &  4.6 &  1.0 &  0.2 &  0.2 &  2.5 &  0.3 &  22.1 &  0.4 &  6.3 &  0.7 &  2.4 &  1.8 & 10.5 &  56.5 &  8.1 & 36.6 & 11.8 \\
      & 40  &  16.6 &  8.3 &  3.6 &  0.5 &  0.3 &  0.3 &  3.2 &  0.4 &  20.5 &  0.4 & 10.3 &  0.4 &  1.5 &  0.4 &  7.5 &  62.9 &  5.7 & 49.7 &  7.5 \\ \midrule
      \multirow{5}{*}{Layer (LN)} 
      & 10   &  22.3 & 17.0 &  2.1 &  1.8 &  0.2 &  0.5 &  0.4 &  0.3 &  71.6 &  0.2 &  0.4 &  0.4 & 14.1 & 14.8 & 41.7 &   6.1 &  0.2 &  2.0 &  3.9 \\
      & 20   &  22.0 & 15.5 &  3.0 &  2.3 &  0.3 &  0.2 &  0.4 &  0.3 &  46.0 &  0.4 &  2.1 &  0.6 & 13.1 &  3.5 & 26.3 &  32.0 &  0.3 & 11.2 & 20.5 \\
      & 24  &  23.0 & 14.1 &  4.7 &  2.4 &  0.2 &  0.4 &  1.0 &  0.2 &  32.2 &  0.3 &  7.0 &  0.7 &  7.4 &  0.5 & 16.3 &  44.8 &  0.2 & 23.4 & 21.2 \\
      & 32  &  26.4 & 12.3 &  7.2 &  2.4 &  0.7 &  0.3 &  3.3 &  0.2 &  24.3 &  0.3 &  5.9 &  0.5 &  4.4 &  0.4 & 12.8 &  49.3 &  0.2 & 40.6 &  8.5 \\
      & 40  &  20.8 &  6.0 &  6.6 &  1.3 &  0.9 &  0.4 &  5.4 &  0.2 &  19.4 &  0.3 &  6.9 &  0.5 &  1.0 &  2.3 &  8.4 &  59.8 &  0.2 & 56.6 &  3.0 \\ \midrule
      \multirow{5}{*}{Step} 
      & 12  &  25.9 & 19.9 &  3.5 &  1.4 &  0.4 &  0.1 &  0.2 &  0.4 &  59.4 &  0.3 &  0.8 &  0.3 & 21.0 & 10.1 & 26.9 &  14.7 &  2.7 &  8.0 &  4.0 \\
      & 15  &  23.7 & 15.7 &  4.2 &  1.5 &  0.3 &  0.2 &  1.4 &  0.4 &  25.7 &  0.2 &  2.7 &  0.8 &  5.3 &  3.2 & 13.5 &  50.6 & 12.4 & 24.7 & 13.5 \\
      & 18  &  22.2 & 13.2 &  4.9 &  1.2 &  0.5 &  0.3 &  1.8 &  0.3 &  24.2 &  0.3 &  5.2 &  1.0 &  3.2 &  2.9 & 11.6 &  53.6 &  7.2 & 28.0 & 18.4 \\
      & 21  &  27.2 & 13.9 &  7.1 &  1.5 &  0.4 &  0.3 &  3.7 &  0.3 &  21.6 &  0.3 &  5.7 &  0.6 &  5.1 &  0.4 &  9.5 &  51.2 &  0.5 & 45.8 &  4.9 \\
      & 33  &  20.5 &  6.1 &  6.9 &  2.4 &  0.8 &  0.5 &  3.2 &  0.6 &  22.6 &  0.3 &  8.1 &  0.8 &  1.8 &  1.6 & 10.0 &  56.9 &  0.2 & 52.9 &  3.8 \\ \midrule
      \multirow{5}{*}{Head} 
      & 3   &  18.3 &  2.4 &  0.3 &  8.2 &  0.4 &  0.2 &  0.3 &  6.5 &   8.9 &  0.3 &  0.3 &  0.4 &  0.4 &  0.2 &  7.3 &  72.8 &  0.3 & 12.4 & 60.1 \\
      & 5   &  18.7 &  8.1 &  1.8 &  3.7 &  0.4 &  0.6 &  0.3 &  3.8 &  29.0 &  0.2 &  5.3 &  1.3 &  3.7 &  4.8 & 13.7 &  52.3 &  0.9 & 14.6 & 36.8 \\
      & 8   &  20.9 &  5.7 &  3.4 &  6.8 &  0.4 &  0.3 &  0.2 &  4.1 &  17.1 &  0.2 &  5.8 &  0.6 &  0.4 &  0.7 &  9.4 &  62.0 &  0.2 & 19.1 & 42.7 \\
      & 12  &  25.4 & 10.8 &  7.3 &  3.5 &  0.4 &  0.3 &  1.7 &  1.4 &  30.5 &  0.4 &  6.6 &  0.3 &  4.1 &  0.5 & 18.6 &  44.1 &  0.2 & 22.1 & 21.8 \\
      & 37  &  28.1 & 12.4 &  8.0 &  1.8 &  0.8 &  0.2 &  4.6 &  0.3 &  32.6 &  0.2 &  4.9 &  0.8 & 12.1 &  0.6 & 14.0 &  39.3 &  0.3 & 34.7 &  4.3 \\ \bottomrule
      \end{tabular}
      }
      \end{center}
      \label{table:shape_distribution}
      \vspace{-0.1in}
      \end{table}

\begin{figure*}[t!]
\begin{center}
\includegraphics[width=0.85\linewidth]{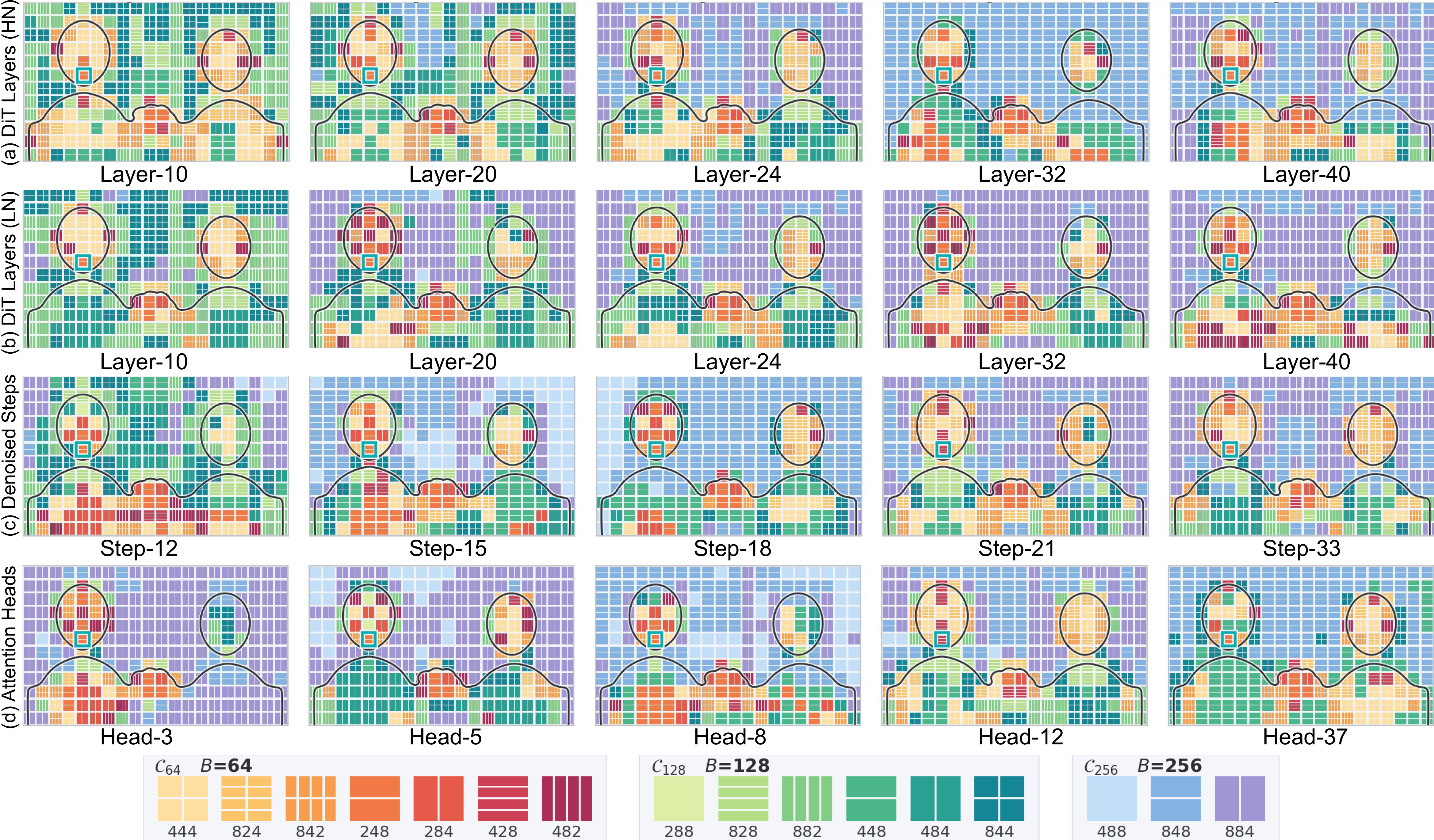}
\end{center}
\vspace{-0.3cm}
   \caption{Ablation study on different DiT layers, denoised steps, and attention heads.
   }
\label{fig:ablation_layers_heads_steps}
\vspace{-0.4cm}
\end{figure*}

We further visualize the block shape layout across different Video DiT layers, denoised steps, and video self-attention heads in Fig. \ref{fig:ablation_layers_heads_steps} and report the proportion of each block shape in Table \ref{table:shape_distribution}.
We have the following observations: 
(1) Across layers, shallow layers use mostly isotropic $\mathcal{C}_{64}$ shapes (dominated by 444) and $\mathcal{C}_{128}$, as early features have not yet developed strong directional variation. Deeper layers progressively replace isotropic shapes with anisotropic ones (824, 428, 842) as the features become more semantically specialized and exhibit clearer directional structure along specific spatiotemporal axes. The $\mathcal{C}_{256}$ share grows with depth as deeper layers aggregate information over broader receptive fields, making many zones sufficiently represented by coarser blocks. However, the anisotropic fraction within the remaining $\mathcal{C}_{64}$ increases, showing that Prism concentrates its fine blocks on the most directionally informative regions rather than distributing them uniformly.
(2) Across denoised steps, as the latent transitions from noisy to clean, the isotropic 444 share within $\mathcal{C}_{64}$ decreases steadily while shapes (824, 428) grow, indicating that high-frequency spatial detail sharpens faster than temporal variation as the latent becomes clean.
The $\mathcal{C}_{256}$ composition also shifts from a balanced directional mixture at early steps to a concentration on temporally elongated coarse blocks at later steps, as background temporal redundancy becomes increasingly apparent in the cleaner latent.
(3) Across attention heads, some heads assign the vast majority of zones to $\mathcal{C}_{256}$ while others retain a large $\mathcal{C}_{64}$ share, revealing that different heads have fundamentally different levels of spatial differentiation in their value features. Within the same block size tier, heads also select orthogonal directional shapes, with some preferring height-oriented shapes (842, 482) and others preferring temporal shapes (824, 428). This confirms that per-head shape assignment captures not only different information density levels but also different directional sensitivities, and forcing heads to share a single shape would destroy this specialization.

\begin{table}[t!]\small
\caption{Ablation study on different training resolutions. GPU Mem is per-GPU memory during 4-GPU parallel inference. Each model generates at its natively training resolution (480p/720p/1080p rows), with all outputs bicubically resized to 1080p for comparable resolution-sensitive AQ/MANIQA across rows. The 2K row is additionally evaluated at its native 2K resolution.
}
\vspace{-0.05in}
\begin{center}
\renewcommand\arraystretch{1.02}
\scalebox{0.64}{
\begin{tabular}{l|l|cccccccccc}
\toprule
Res. & Method & AQ$\uparrow$ & TF$\uparrow$ & PQ$\uparrow$ & CU$\uparrow$ & DeSync$\downarrow$ & cpCER$\downarrow$ & MANIQA$\uparrow$ & MotionQ$\uparrow$ & Training Time$\downarrow$ & Infer GPU Mem$\downarrow$ \\ \midrule
\multirow{7}{*}{480P}
& Full Attn      & \underline{0.56} & \underline{0.968} & \underline{7.42} & \underline{6.95} & \underline{0.71} & \underline{0.234} & \underline{0.348} & \underline{0.59} & 0.6min & 10.4G \\
& VMoBA          & 0.52 & 0.961 & 7.26 & 6.78 & 0.77 & 0.258 & 0.336 & 0.54 & 0.5min & 7.6G \\
& SpargeAttn2    & 0.49 & 0.951 & 7.16 & 6.68 & 0.84 & 0.276 & 0.326 & 0.49 & 0.3min & 5.1G \\
& PyramidFlow    & 0.50 & 0.964 & 7.11 & 6.63 & 0.75 & 0.249 & 0.323 & 0.47 & 0.7min & 11.4G \\
& \textit{w/o} Video Guidance  & 0.48 & 0.953 & 7.19 & 6.71 & 0.81 & 0.244 & 0.328 & 0.50 & 0.3min & 4.7G \\
& \textit{w/o} Audio Guidance      & 0.53 & 0.963 & 7.32 & 6.85 & 0.76 & 0.268 & 0.339 & 0.55 & 0.5min & 5.3G \\
& Ours           & 0.54 & 0.966 & 7.36 & 6.88 & 0.74 & 0.241 & 0.342 & 0.57 & 0.5min & 5.8G \\ \midrule
\multirow{7}{*}{720P}
& Full Attn      & 0.55 & 0.956 & 7.34 & 6.87 & 0.78 & 0.271 & 0.367 & 0.62 & 2.1min & 21.6G \\
& VMoBA          & 0.54 & 0.963 & 7.36 & 6.88 & 0.74 & 0.258 & 0.368 & 0.61 & 1.9min & 14.2G \\
& SpargeAttn2    & 0.57 & 0.961 & 7.44 & 6.97 & 0.73 & 0.243 & 0.374 & 0.64 & 1.2min & 9.8G \\
& PyramidFlow    & 0.52 & 0.969 & 7.28 & 6.79 & 0.76 & 0.254 & 0.362 & 0.60 & 2.2min & 23.8G \\
& \textit{w/o} Video Guidance  & 0.55 & 0.960 & 7.38 & 6.89 & 0.78 & 0.238 & 0.371 & 0.63 & 1.1min & 8.6G \\
& \textit{w/o} Audio Guidance      & 0.57 & 0.971 & 7.46 & 6.98 & 0.77 & 0.276 & 0.381 & 0.68 & 1.4min & 10.1G \\
& Ours           & \underline{0.58} & \underline{0.974} & \underline{7.52} & \underline{7.06} & \underline{0.70} & \underline{0.221} & \underline{0.382} & \underline{0.71} & 1.4min & 11.3G \\ \midrule
\multirow{7}{*}{1080P}
& Full Attn      & 0.48 & 0.931 & 7.08 & 6.56 & 0.89 & 0.328 & 0.376 & 0.57 & 8.7min & 41.2G \\
& VMoBA          & 0.51 & 0.948 & 7.29 & 6.80 & 0.82 & 0.289 & 0.378 & 0.56 & 7.6min & 27.8G \\
& SpargeAttn2    & 0.54 & 0.951 & 7.38 & 6.89 & 0.78 & 0.271 & 0.381 & 0.58 & 3.8min & 16.5G \\
& PyramidFlow    & 0.53 & 0.961 & 7.24 & 6.74 & 0.80 & 0.276 & 0.383 & 0.68 & 7.8min & 44.3G \\
& \textit{w/o} Video Guidance  & 0.55 & 0.953 & 7.36 & 6.87 & 0.76 & 0.228 & 0.393 & 0.69 & 4.0min & 14.8G \\
& \textit{w/o} Audio Guidance      & 0.58 & 0.971 & 7.49 & 7.02 & 0.76 & 0.281 & 0.412 & 0.76 & 4.3min & 16.9G \\
& Ours           & \underline{0.59} & \underline{0.979} & \underline{7.62} & \underline{7.26} & \underline{0.65} & \underline{0.196} & \underline{0.416} & \underline{0.82} & 4.6min & 18.4G \\ \midrule
\multirow{7}{*}{2K}
& Full Attn      & 0.42 & 0.903 & 6.80 & 6.26 & 1.05 & 0.374 & 0.363 & 0.53 & 26.5min & 72.8G \\
& VMoBA          & 0.46 & 0.924 & 7.04 & 6.52 & 0.91 & 0.328 & 0.381 & 0.59 & 15.4min & 43.6G \\
& SpargeAttn2    & 0.47 & 0.928 & 7.10 & 6.55 & 0.90 & 0.319 & 0.384 & 0.61 & 8.4min & 36.7G \\
& PyramidFlow    & 0.51 & 0.944 & 7.08 & 6.59 & 0.84 & 0.295 & 0.391 & 0.70 & 18.3min & 65.8G \\
& \textit{w/o} Video Guidance  & 0.53 & 0.943 & 7.31 & 6.82 & 0.73 & 0.219 & 0.399 & 0.70 & 8.4min & 36.4G \\
& \textit{w/o} Audio Guidance      & 0.57 & 0.969 & 7.46 & 6.97 & 0.79 & 0.263 & 0.420 & 0.78 & 9.2min & 36.5G \\
& Ours           & \underline{\textbf{0.61}} & \underline{\textbf{0.982}} & \underline{\textbf{7.69}} & \underline{\textbf{7.34}} & \underline{\textbf{0.63}} & \underline{\textbf{0.187}} & \underline{\textbf{0.438}} & \underline{\textbf{0.89}} & 10.6min & 38.0G \\ \bottomrule
\end{tabular}
}
\end{center}
\label{table:ablation_resolution}
\vspace{-0.20in}
\end{table}

\subsection{Natively Training at Scaling Resolution}
\label{sec: native_training_at_scaling_resolution}

\begin{figure*}[t!]
\begin{center}
\includegraphics[width=0.65\linewidth]{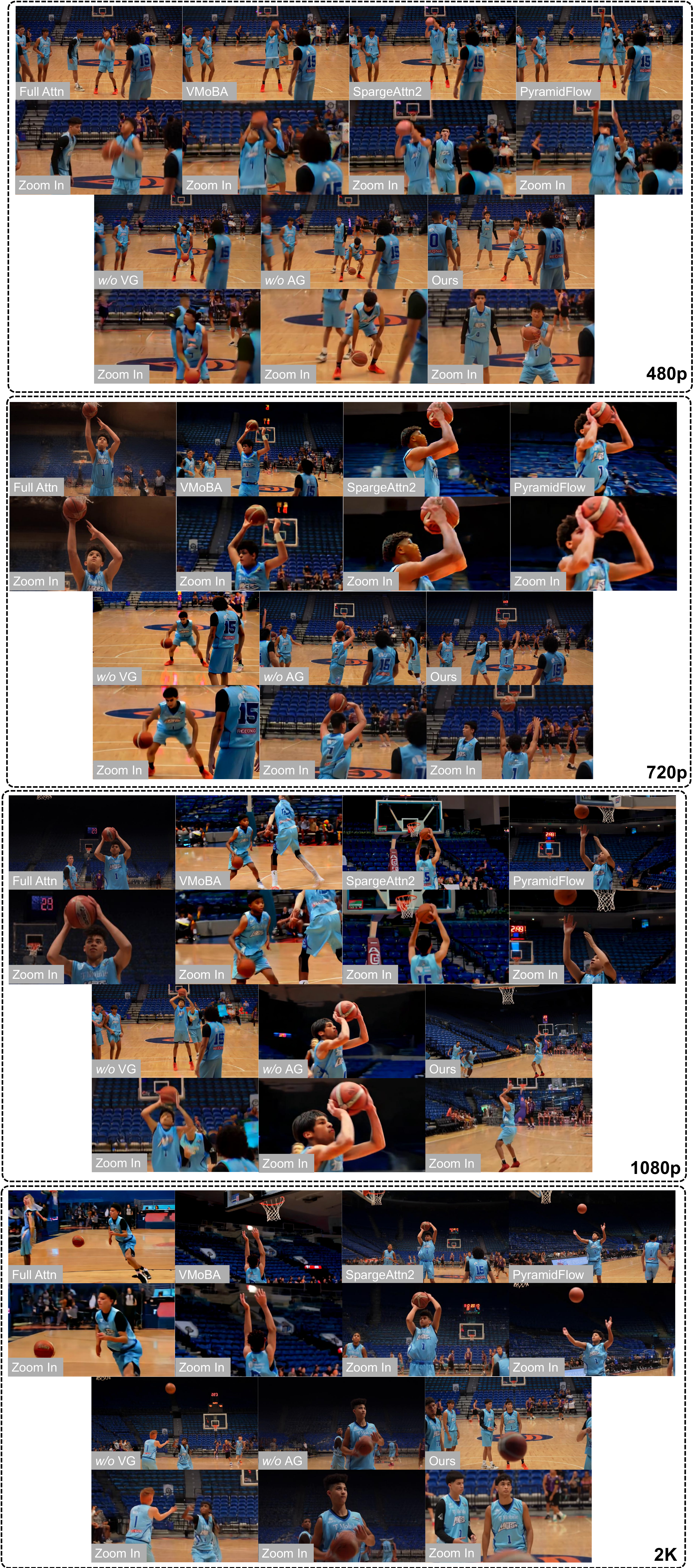}
\end{center}
\vspace{-0.4cm}
   \caption{Ablation study on different training resolutions. Please refer to the demo video for a clear comparison. VG and AG refer to video channel-wise variance guidance and audio-to-video cross-attention norm guidance.
   }
\label{fig:ablation_different_resolution}
\vspace{-0.4cm}
\end{figure*}

We conduct an ablation study on different resolutions, as shown in Table \ref{table:ablation_resolution} and Fig. \ref{fig:ablation_different_resolution}.
We perform native joint video-audio training at 480P ($854 \times 480$), 720P ($1280 \times 720$), 1080P ($1920 \times 1080$), and 2K ($2560 \times 1440$). Notably, all training sets are derived from the same videos by resizing to different resolutions. \textit{w/o} Video Guidance and \textit{w/o} Audio Guidance remove the video channel-wise variance guidance and audio-to-video cross-attention norm guidance during the dynamic block shape mapping.
Four key findings emerge.
(1) Native high-resolution training introduces a fundamental tension between the richer visual details and the growing proportion of redundant tokens. At 480P, the token sequence is short and contains minimal redundancy, so Full Attn models all interactions effectively and leads all methods. As resolution scales to 720P and beyond, the number of redundant tokens grows quadratically while informative content remains spatially localized. 
For methods that do not explicitly address this redundancy, quality degrades at 1080P and sharply declines at 2K. Full Attn even drops below its own 1080P performance at 2K.
Cascaded methods (PyramidFlow) avoid the full 2K attention cost but inherit low-resolution structural bias. 
Sparse methods (VMoBA, SpargeAttn2) mitigate some redundancy but their fixed block shapes fail to adapt to the increasingly uneven information distribution and visual-audio coupling at high resolution.
\textit{w/o} Video Guidance loses the ability to adapt block shapes to visual content variation, causing AQ and TF to stagnate beyond 1080P.
\textit{w/o} Audio Guidance preserves visual quality scaling, but cpCER and DeSync basically degrade with resolution, as block shapes can no longer concentrate on sound-producing regions whose spatial proportion shrinks at higher resolution.
(2) Despite these challenges, MANIQA and MotionQ improve with resolution across most methods, confirming that native high-resolution training exposes genuinely finer visual details and motion patterns. In contrast, among all compared methods, only Prism and its \textit{w/o} Video Guidance ablation (which retains audio guidance) consistently improve DeSync and cpCER with resolution.
The audio-only metrics PQ and CU in Prism also improve with video resolution, which may appear counter-intuitive since the audio branch and its token length are identical at all four resolutions. The reason is that the audio branch is conditioned on video tokens through the video-to-audio cross-attention, so the visual stream acts as the conditioning signal for audio denoising. Sharper motion dynamics and cleaner object boundaries at higher resolution yield a less ambiguous conditioning signal about what is producing sound and when, which in turn improves the perceptual quality of the generated audio. 
From 480P to 2K, Prism improves MANIQA by 28\% and MotionQ by 56\%, demonstrating that higher resolution provides genuinely new spatial information (skin textures, fabric patterns, subtle hand gestures) that the model can absorb when the attention mechanism is properly designed to concentrate on informative regions.
(3) The training efficiency gap between Full Attn and sparse methods widens dramatically with resolution. At 480P, Prism achieves comparable training speed to Full Attn (both are fast). 
At 2K, Prism achieves 2.5$\times$ training speedup, as the quadratic attention cost that dominates Full Attn at 2K is directly addressed by sparse attention. 
PyramidFlow is faster than Full Attn at 1080P+ due to LoRA-based refinement, but its GPU memory remains high from multi-stage latent storage.
(4) Competitors improve on some metrics (MANIQA, MotionQ) from 1080P to 2K, as higher resolution inherently provides more spatial detail, but degrade on others (AQ, TF, DeSync), revealing that they can absorb fine-grained spatial information but cannot handle the overwhelming redundancy. 
A particularly important observation is that all competitors degrade on cross-modal synchronization metrics cpCER and DeSync from 1080P to 2K. 
At higher resolution, sound-producing regions occupy a proportionally smaller fraction of the total token sequence, making audio-visual coupling increasingly difficult to capture without explicit cross-modal guidance. 
VMoBA, SpargeAttn2, and PyramidFlow lack any audio-aware mechanism in their attention design, so their block shapes or window partitions are entirely agnostic to where audio-visual coupling occurs, and this deficiency worsens as the proportion of audio-relevant tokens shrinks at 2K.
Prism is the only method that improves on all metrics from 480P to 2K, as its audio-to-video cross-attention norm guidance steers finer block partitioning toward sound-producing regions regardless of their spatial proportion, while video channel-wise variance adapts shapes to visual content, jointly ensuring that higher resolution translates into quality gains. At 480P, Prism trails Full Attn slightly, as the short sequence has little redundancy and the dynamic machinery provides minimal benefit. The advantage emerges at 720P, grows at 1080P, and becomes dominant at 2K where Prism simultaneously absorbs 2K spatial richness, resists massive redundancy, and captures precise audio-visual coupling.

Qualitative results (Fig. \ref{fig:ablation_different_resolution}) further reveal that resolution scaling constitutes a fundamentally distinct capability scaling axis from model parameters and training data.
Scaling parameters enriches the model's representational capacity, and scaling data broadens the distribution coverage, yet both operate on a fixed spatial grid that inherently bounds the granularity of learnable visual details and motion dynamics.
Resolution scaling lifts this spatial ceiling, exposing genuinely new information such as skin textures, subtle finger articulations, and fine lip movements that simply do not exist at lower resolution grids regardless of how large the model or dataset becomes.
However, this new scaling axis is uniquely fragile. 
Unlike parameter and data scaling, where increasing scale generally leads to better performance, resolution scaling introduces an increasing proportion of redundant tokens that can actively degrade training quality when attention fails to distinguish informative content from repetitive backgrounds.
Competitors trained at 2K exhibit sharper textures than their 480p counterparts yet simultaneously suffer from worsened flickering and structural collapse, revealing that they absorb the new spatial information but cannot resist the accompanying redundancy.
Prism resolves this tension through dynamic sparse attention that selectively concentrates on informative regions, making resolution scaling reliably translate into monotonic quality improvement across all visual and cross-modal metrics and establishing native resolution scaling as a practical and effective capability scaling paradigm complementary to parameter and data scaling.

\subsection{Native 2K Training Loss}
\label{sec: native_2k_training_loss}

\begin{figure*}[t!]
\begin{center}
\includegraphics[width=1.0\linewidth]{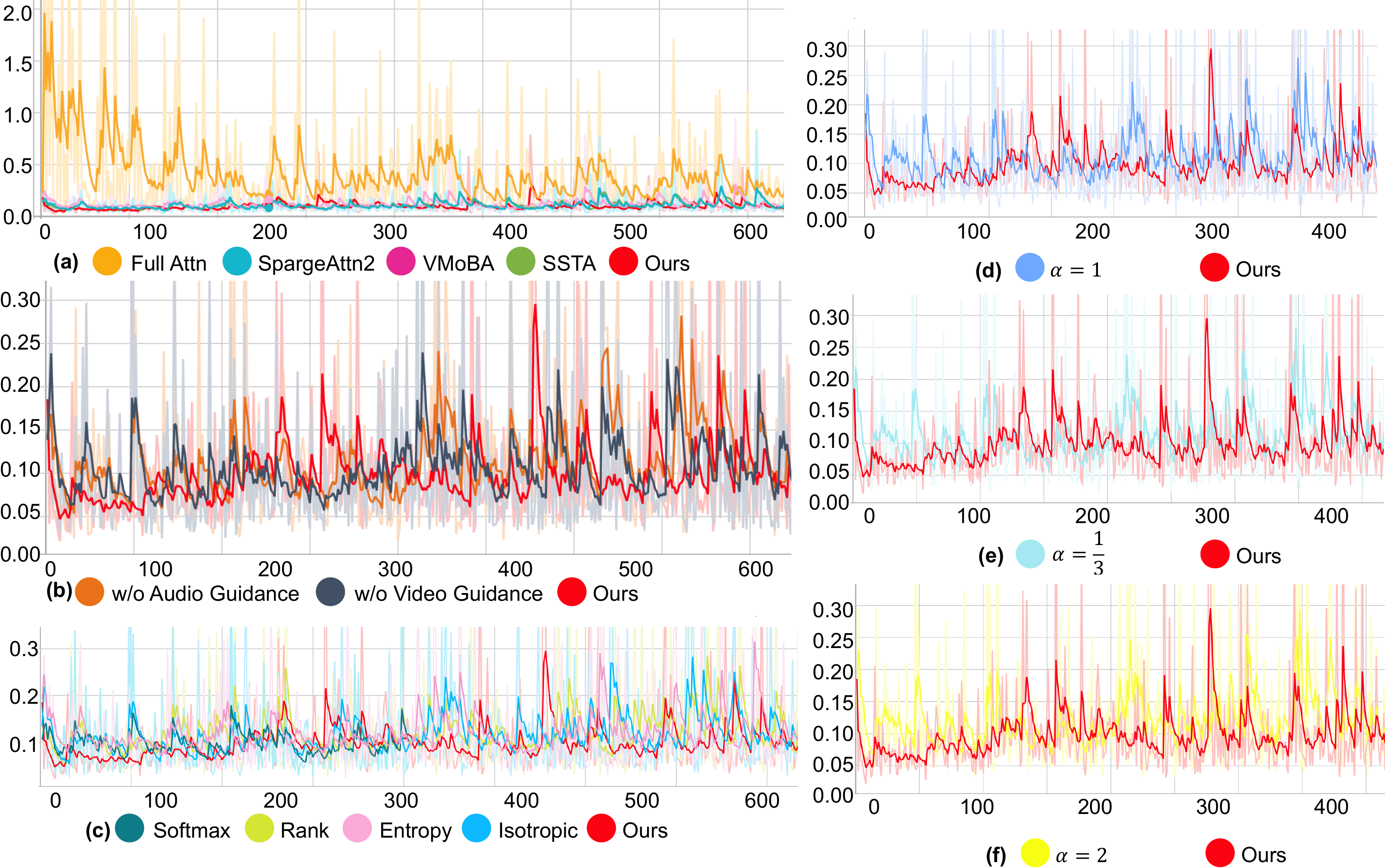}
\end{center}
\vspace{-0.3cm}
   \caption{Visualization of video training loss.
   }
\label{fig:ablation_video_loss}
\vspace{-0.4cm}
\end{figure*}

\begin{figure*}[t!]
\begin{center}
\includegraphics[width=1.0\linewidth]{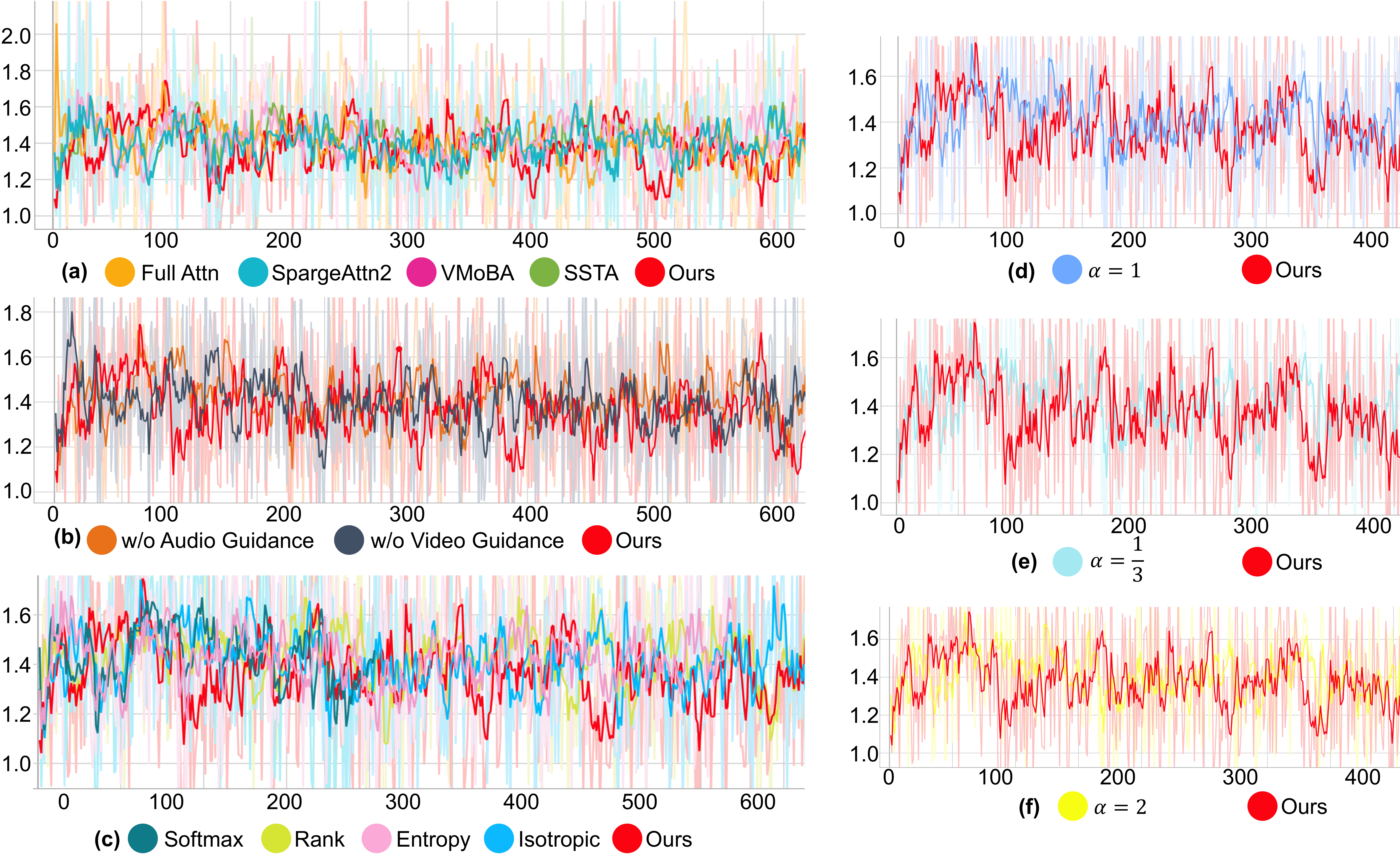}
\end{center}
\vspace{-0.3cm}
   \caption{Visualization of audio training loss.
   }
\label{fig:ablation_audio_loss}
\vspace{-0.4cm}
\end{figure*}

Fig .\ref{fig:ablation_video_loss} and Fig. \ref{fig:ablation_audio_loss} show the video and audio training loss during native high-resolution training.
A cross-cutting observation is that video loss and audio loss reflect two fundamentally different challenges at 2K. Video loss primarily captures how well the attention mechanism resists attention weight dilution from visual redundancy, while audio loss captures whether cross-modal coupling is preserved throughout training. 
In Fig.~\ref{fig:ablation_video_loss}(a) and Fig.~\ref{fig:ablation_audio_loss}(a), Full Attn exhibits the highest video loss with extreme instability and persistent large spikes throughout training, revealing that the massive redundant tokens at 2K not only dilute attention weights on informative interactions but also inject severe noise into the optimization trajectory, preventing stable convergence. Its audio loss shows a clear upward drift over training, confirming that attention weight dilution progressively erodes the pretrained cross-modal coupling and the model increasingly loses its ability to coordinate video and audio generation.
SpargeAttn2, VMoBA, and SSTA converge well below Full Attn on the video loss, confirming that sparse attention effectively filters visual redundancy at 2K. Notably, these three methods cluster at a similar video loss level despite their architectural differences, suggesting that fixed-shape sparse attention has an inherent performance ceiling for visual modeling at 2K, and further improvement requires content-adaptive block partitioning rather than merely different fixed partitioning strategies. 
However, their audio losses still exhibit an upward trend similar to Full Attn, revealing that their content-agnostic block shapes can partially address visual redundancy but fundamentally fail to preserve audio-visual coupling where sound-producing regions occupy a diminishing spatial proportion at 2K.
Prism achieves both the lowest video loss and the lowest audio loss with the most stable convergence, breaking through the performance ceiling of fixed-shape methods on video loss while simultaneously preventing the audio loss drift that all competitors exhibit.
In Fig.~\ref{fig:ablation_video_loss}(b) and Fig.~\ref{fig:ablation_audio_loss}(b), the asymmetric degradation pattern reveals the complementary roles of the two guidance signals. \textit{w/o} Video Guidance causes the most severe video loss degradation with frequent large spikes, while its audio loss degrades moderately, confirming that video channel-wise variance is the primary driver for concentrating gradients on visually informative regions. \textit{w/o} Audio Guidance yields a lower video loss than \textit{w/o} Video Guidance yet produces a noticeably higher and rising audio loss, revealing that audio-to-video cross-attention norm guidance primarily governs cross-modal stability rather than visual quality. 
It shows that block shape quality for visual modeling and block shape quality for audio-visual coupling are governed by different guidance signals operating on orthogonal aspects of the information structure, and their joint operation in Prism is necessary to simultaneously optimize both modalities.
In Fig.~\ref{fig:ablation_video_loss}(c) and Fig.~\ref{fig:ablation_audio_loss}(c), Isotropic produces the highest and most unstable loss on both modalities, confirming that ignoring directional information structure and treating all spatiotemporal axes uniformly fundamentally corrupts block representatives at 2K. Rank shows substantial instability with erratic spikes, as discarding variance magnitude information causes the mapping to assign identical block shapes to zones with very different variance magnitudes. Softmax and Entropy achieve intermediate performance. Softmax saturates for large variance values and loses the ability to distinguish axes with similarly strong variance, while Entropy cannot capture the concave $g_d^{-1/2}$ relationship.
In Fig.~\ref{fig:ablation_video_loss}(d)(e)(f) and Fig.~\ref{fig:ablation_audio_loss}(d)(e)(f), the three $\alpha$ values reveal a clear progression. $\alpha{=}1/3$ converges slowly with elevated loss on both modalities, as the block shapes respond too weakly to directional variance, leaving blocks nearly isotropic and mixing semantically dissimilar tokens. $\alpha{=}1$ improves substantially over $\alpha{=}1/3$ and approaches Prism, yet a persistent gap remains on both video and audio losses. $\alpha{=}2$ is the most revealing setting, as it initially tracks Prism closely in early training but diverges dramatically in later stages with rising loss and severe instability on both modalities. This delayed divergence exposes a compounding error mechanism. 
At each training step, overly sensitive shape assignments amplify small variation indicator estimation noise into large changes in block shapes. These unstable partitions then corrupt the mean-pooled representations, distorting block selection scores and further degrading the variance signal used for shape estimation in the next step. This feedback loop is manageable when features are still noisy in early training, but can become catastrophic as features stabilize, with persistent block-shape oscillations preventing the model from refining fine-grained details.
The Lagrange-derived $\alpha{=}1/2$ uniquely avoids this instability while maintaining sufficient responsiveness, yielding the only stable converging trajectory across both video and audio losses throughout training.

\subsection{Attention Visualization Results}
\label{sec: attention_visualization_results}

\begin{figure*}[t!]
\begin{center}
\includegraphics[width=1.0\linewidth]{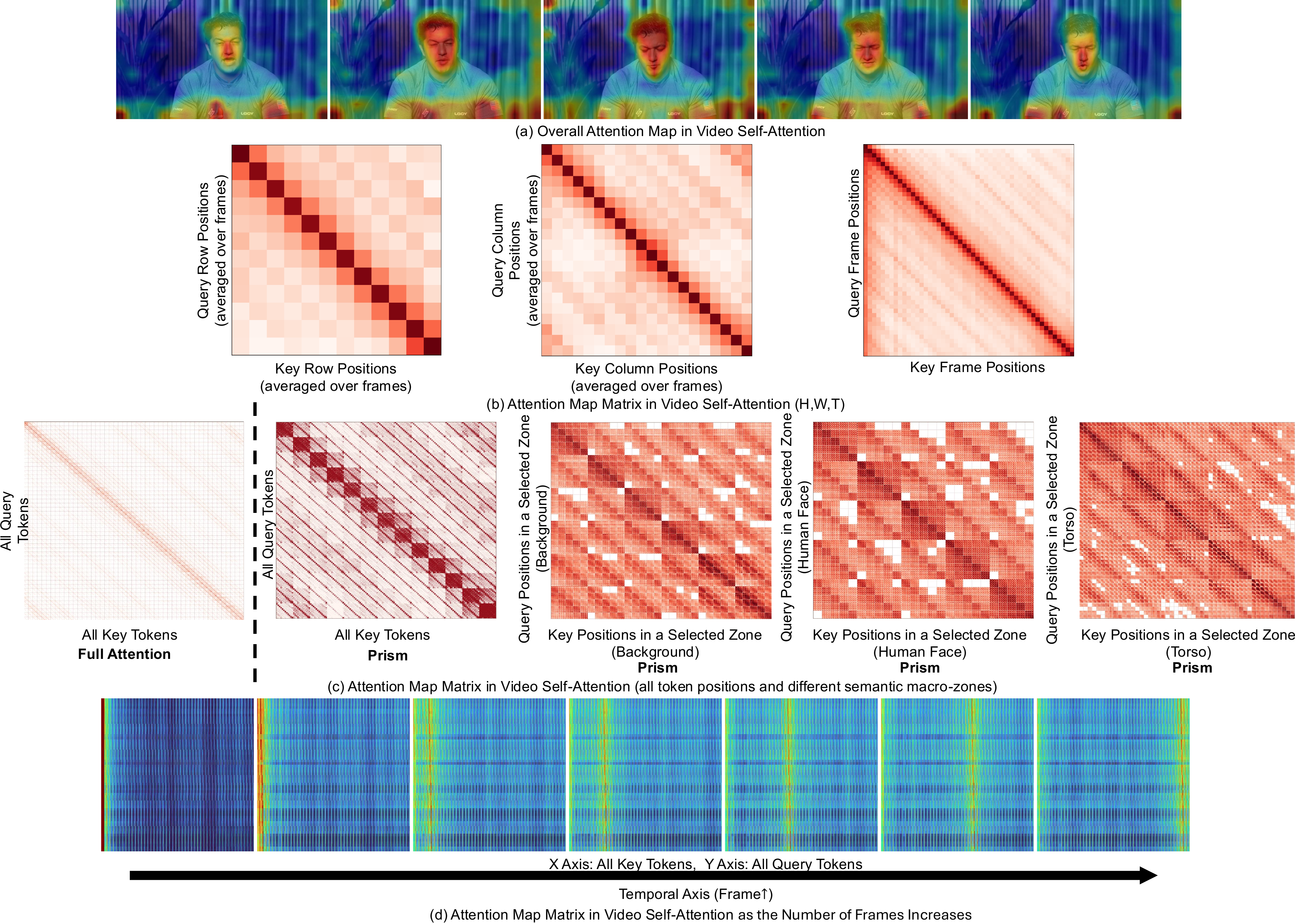}
\end{center}
\vspace{-0.4cm}
   \caption{Visualization of our attention maps.
   }
\label{fig:attention_map_vis}
\vspace{-0.4cm}
\end{figure*}

\begin{figure*}[t!]
\begin{center}
\includegraphics[width=1.0\linewidth]{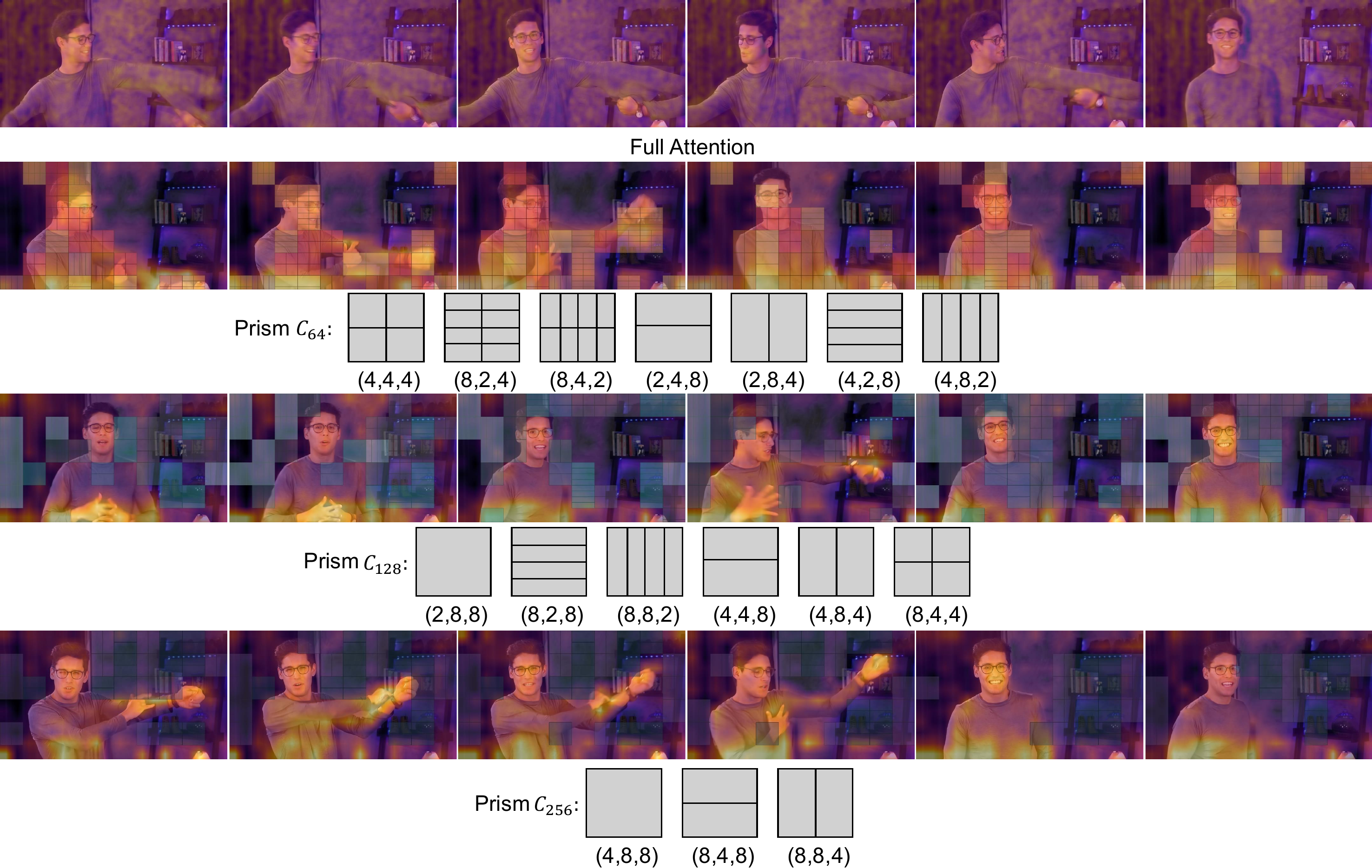}
\end{center}
\vspace{-0.4cm}
   \caption{Visualization of dynamic block shapes.
   }
\label{fig:block_shape_attention_map_vis}
\vspace{-0.4cm}
\end{figure*}

Fig. \ref{fig:attention_map_vis} and Fig. \ref{fig:block_shape_attention_map_vis} show our attention visualization results.
We illustrate both the attention matrix and the attention feature map of full attention after native high-resolution joint video-audio training in Fig. \ref{fig:attention_map_vis} and Fig. \ref{fig:block_shape_attention_map_vis}, indicating the attention weight dilution issue.
In Fig.~\ref{fig:attention_map_vis}(a), the overall attention maps overlaid on the video frames reveal that attention concentrates heavily on the face and mouth regions while assigning minimal weight to static backgrounds, confirming that Prism's dynamic sparse attention successfully identifies and prioritizes semantically informative regions even within the ultra-long 2K token sequence.
In Fig.~\ref{fig:attention_map_vis}(b), the three attention matrices along the H, W, and T axes exhibit strong diagonal dominance in the spatial dimensions (H and W) with a broader yet still diagonal pattern along the temporal axis (T). The tight spatial diagonal indicates that each query primarily attends to its spatially neighboring tokens, reflecting the spatial locality of visual content at 2K, where fine details such as skin textures and fabric patterns are inherently localized. The broader temporal diagonal reveals that queries attend to a wider range of temporally neighboring tokens, capturing the motion dynamics that span multiple frames. This anisotropic attention structure directly validates our core motivation that block shapes should be spatially compact yet temporally extended for regions with smooth spatial content and rapid temporal variation, and vice versa for regions with complex spatial detail. Fig.~\ref{fig:attention_map_vis}(c) further demonstrates our dynamic block shape patterns during native high-resolution joint video-audio training.

Fig.~\ref{fig:block_shape_attention_map_vis} validates that Prism's shape assignment aligns precisely with the information structure. 
$\mathcal{C}_{64}$ shapes (fine-grained blocks) concentrate on the face and hand regions, which are the primary sound-producing and visually dynamic areas. 
$\mathcal{C}_{256}$ shapes (coarse blocks) are assigned to the static bookshelf and wall backgrounds, where tokens are highly repetitive and a single large block suffices to represent them without information loss. $\mathcal{C}_{128}$ shapes cover the intermediate regions such as the torso and clothing boundaries.
Within each block size tier, the directional shape assignment is equally principled. 
For instance, in $\mathcal{C}_{64}$, the face region when the person is actively speaking receives temporally fine shapes with $b_T{=}2$, such as $(2,4,8)$ or $(2,8,4)$, as the mouth opens and closes rapidly across frames, demanding fine temporal partitioning to separate visually distinct mouth states into different blocks. When the lips remain static, the same face region shifts to temporally coarser shapes with $b_T{=}8$, such as $(8,2,4)$ and $(8,4,2)$, which instead allocate finer partitioning to the spatial axes where facial detail remains rich. This temporal adaptivity within the same spatial region is precisely what fixed-shape methods cannot achieve and is essential for representing sound-producing regions whose temporal dynamics are inherently non-stationary. 
Background regions are assigned to $\mathcal{C}_{256}$ predominantly with large spatial edges, reflecting that these regions are spatially homogeneous and benefit from coarse spatial grouping.
This visualization confirms that the dynamic block shape is not an arbitrary assignment but a principled adaptation to the local directional information structure. The block size tier captures the information density (fine blocks for informative regions and coarse blocks for redundant regions), while the block shape within each tier captures the directional anisotropy (shorter edges along axes of rapid variation). Together, they ensure that tokens within each block remain semantically coherent, which is the foundational requirement for reliable mean-pooled block representatives in the downstream sparse selection.

\subsection{Speed and GPU Resource Comparison}
\label{sec: speed_gpu_resource_comparison}

We compare the inference speed and GPU memory consumption between our Prism and previous joint video-audio generation models, as shown in Table \ref{table:speed_gpu}.
The reported GPU Mem refers to the inference GPU memory consumption in all tables.
The inference Speed is measured for native 10-second (FPS=24) 2K joint video-audio generation on 4 NVIDIA H800 GPUs. Notably, we utilize the full version of LTX-2.3~\citep{hacohen2026ltx} instead of the distilled version for fair performance comparison. 
Compared to our Full Attn backbone, Prism is over 3$\times$ faster and uses 47\% less GPU memory while improving AQ by 45\%, MotionQ by 68\%, and DeSync by 40\%, demonstrating that dynamic sparse attention simultaneously unlocks both efficiency and quality gains at 2K by filtering redundant computation that actively harms full attention.
Compared to SpargeAttn2~\citep{zhang2026spargeattention2}, Prism uses comparable inference time and GPU memory yet delivers better AQ and MotionQ.
Compared to LTX-2.3, Prism achieves comparable speed with half the GPU memory while outperforming it across all quality metrics, indicating that native 2K sparse attention is a more effective path than the generate-then-upsample pipeline.

Furthermore, we decompose Prism's total 2K inference latency (331s) into four stages (Algorithm. \ref{alg:prism_alignmth}): Stage 1 (dynamic shape computation) takes 18s (5.4\%), Stage 2 (token rearrangement) takes 5s (1.5\%), Stage 3 (block scoring and hybrid Top-$k$/Top-$p$ selection) takes 22s (6.6\%), and Stage 4 (sparse attention kernel and output scatter) takes 286s (86.4\%).
Stages 1--3 together account for only 45s (13.6\%). This shows that the dynamic shape assignment machinery introduces marginal overhead relative to the attention kernel.

\begin{table}[t!]\small
\caption{Inference speed and GPU memory comparison. GPU Mem is per-GPU memory during 4-GPU parallel inference, as the token count at 2K (10s, FPS=24) reaches approximately 870K, making single-GPU inference prohibitively expensive.
}
\vspace{-0.05in}
\begin{center}
\renewcommand\arraystretch{1.1}
\scalebox{0.76}{
\begin{tabular}{lcccccccccc}
\toprule
Model     & AQ$\uparrow$ & TF$\uparrow$ & PQ$\uparrow$ & CU$\uparrow$ & DeSync$\downarrow$ & cpCER$\downarrow$ & MANIQA$\uparrow$ & MotionQ$\uparrow$ & Inference Speed$\downarrow$ & Infer GPU Mem$\downarrow$ \\ \midrule
LTX-2.3~\citep{hacohen2026ltx}   & 0.48 & 0.943 & 7.05 & 6.83 & 0.95 & 0.382 & 0.403 & 0.66 & 347s & 78.4G \\
MagiHuman~\citep{davinci-magihuman-2026} & 0.40 & 0.912 & 6.72 & 6.14 & 1.08 & 0.420 & 0.337 & 0.58 & 386s & 74.2G \\
Ovi~\citep{low2025ovi}         & 0.38 & 0.891 & 6.68 & 5.92 & 1.12 & 0.468 & 0.326 & 0.42 & 724s & 68.7G \\
MOVA (Full Attn)~\citep{team2026mova} & 0.42 & 0.903 & 6.80 & 6.26 & 1.05 & 0.374 & 0.363 & 0.53 & 1091s & 72.8G \\ \midrule
VMoBA~\citep{wu2025vmoba}      & 0.46 & 0.924 & 7.04 & 6.52 & 0.91 & 0.328 & 0.381 & 0.59 & 483s & 43.6G \\
SpargeAttn2~\citep{zhang2026spargeattention2} & 0.47 & 0.928 & 7.10 & 6.55 & 0.90 & 0.319 & 0.384 & 0.61 & 248s & 36.7G \\
PyramidFlow~\citep{jin2025pyramidal} & 0.51 & 0.944 & 7.08 & 6.59 & 0.84 & 0.295 & 0.391 & 0.70 & 1164s & 65.8G \\ \midrule
Ours      & \textbf{0.61} & \textbf{0.982} & \textbf{7.69} & \textbf{7.34} & \textbf{0.63} & \textbf{0.187} & \textbf{0.438} & \textbf{0.89} & 331s & 38.0G \\ \bottomrule
\end{tabular}
}
\end{center}
\label{table:speed_gpu}
\vspace{-0.20in}
\end{table}

\subsection{Multi-Speaker Results}
\label{sec: multi_speaker_results}

\begin{figure*}[t!]
\begin{center}
\includegraphics[width=0.95\linewidth]{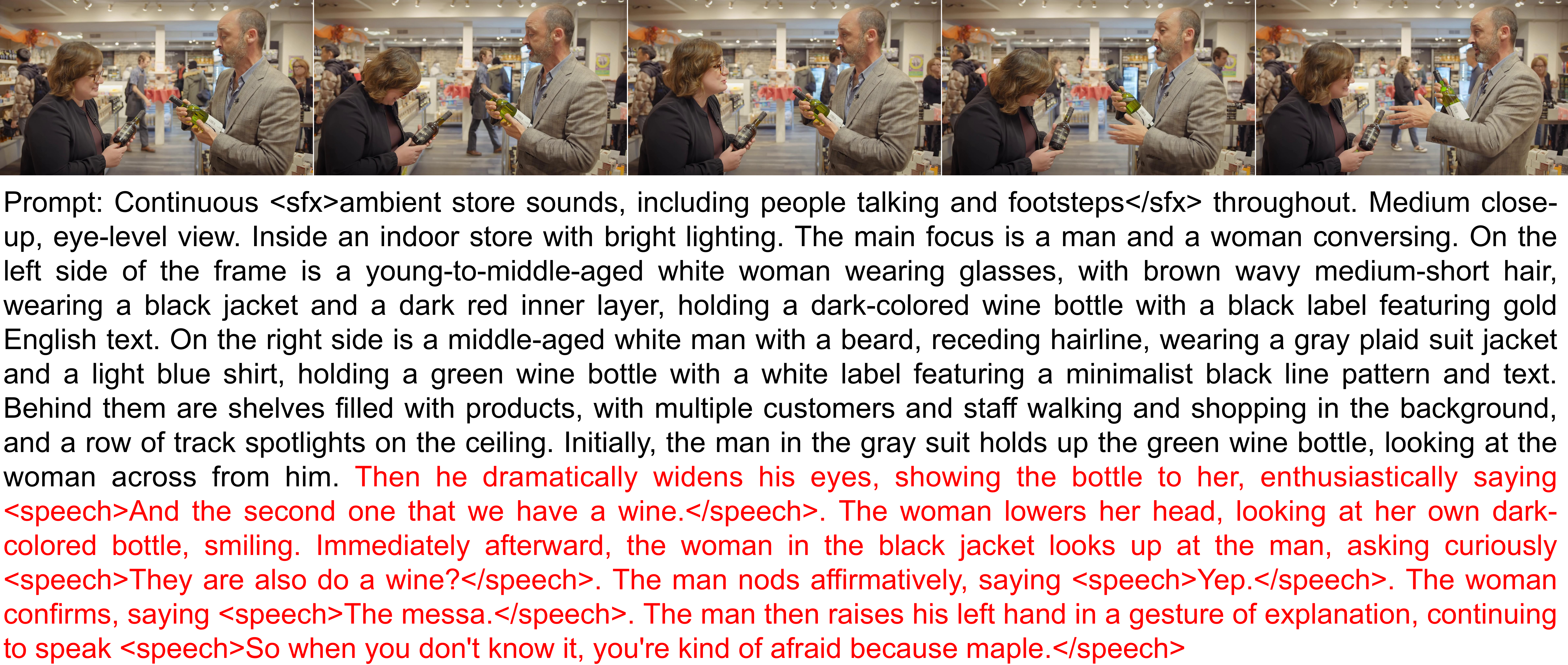}
\end{center}
\vspace{-0.4cm}
   \caption{ Synthesized video-audio content involving multiple speakers. Please refer to the demo video for audio.
   }
\label{fig:multi_speaker}
\vspace{-0.4cm}
\end{figure*}

Fig \ref{fig:multi_speaker} shows a multi-speaker case. The result demonstrates that our model is capable of handling scenarios involving multi-person communication.

\subsection{Complex Scene Results}
\label{sec: complex_scene_results}

\begin{figure*}[t!]
\begin{center}
\includegraphics[width=0.95\linewidth]{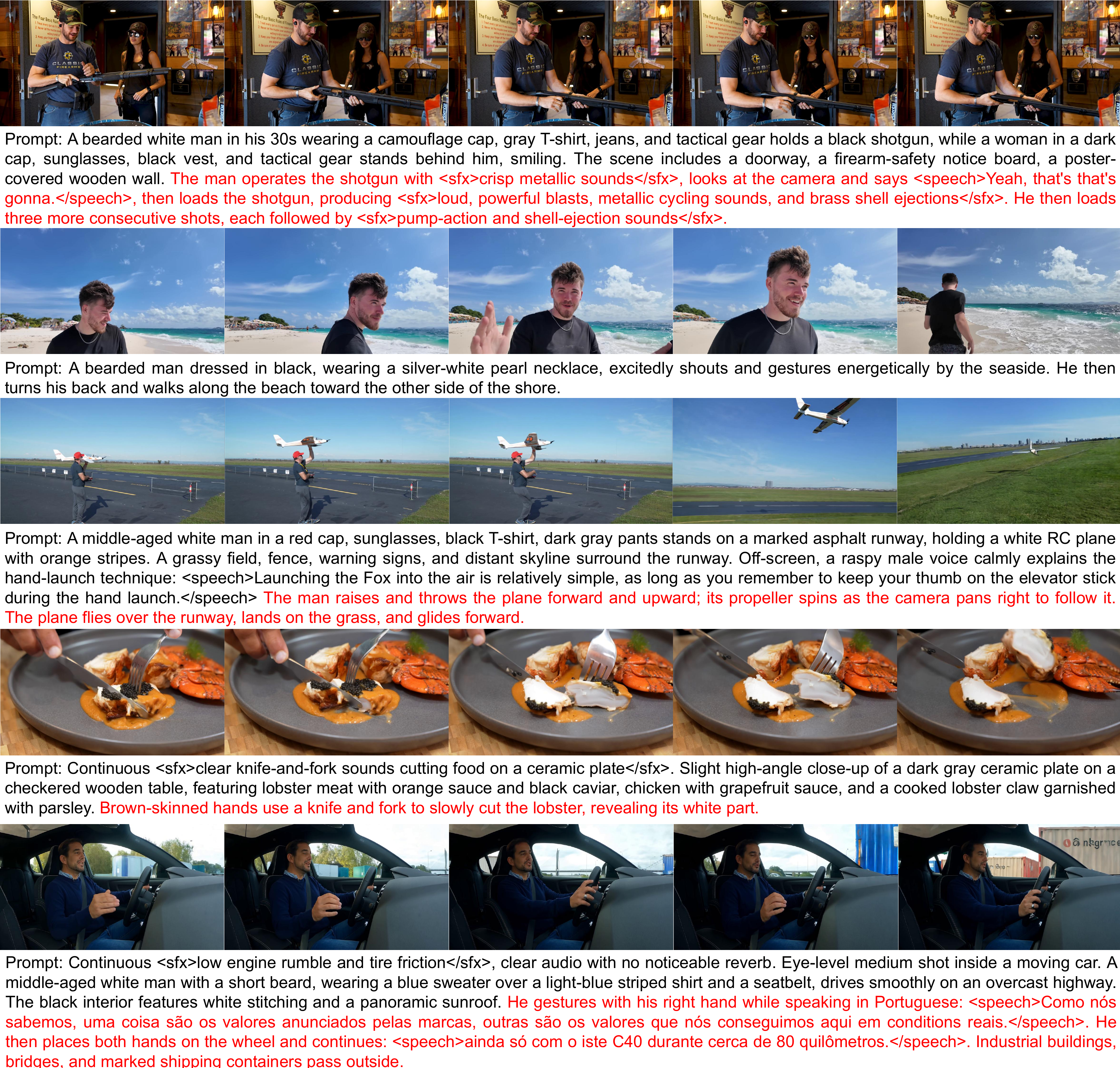}
\end{center}
\vspace{-0.3cm}
   \caption{Complex scene results (1/5). Please refer to the demo video for audio.
   }
\label{fig:complex_scene_1}
\vspace{-0.3cm}
\end{figure*}

\begin{figure*}[t!]
\begin{center}
\includegraphics[width=1.0\linewidth]{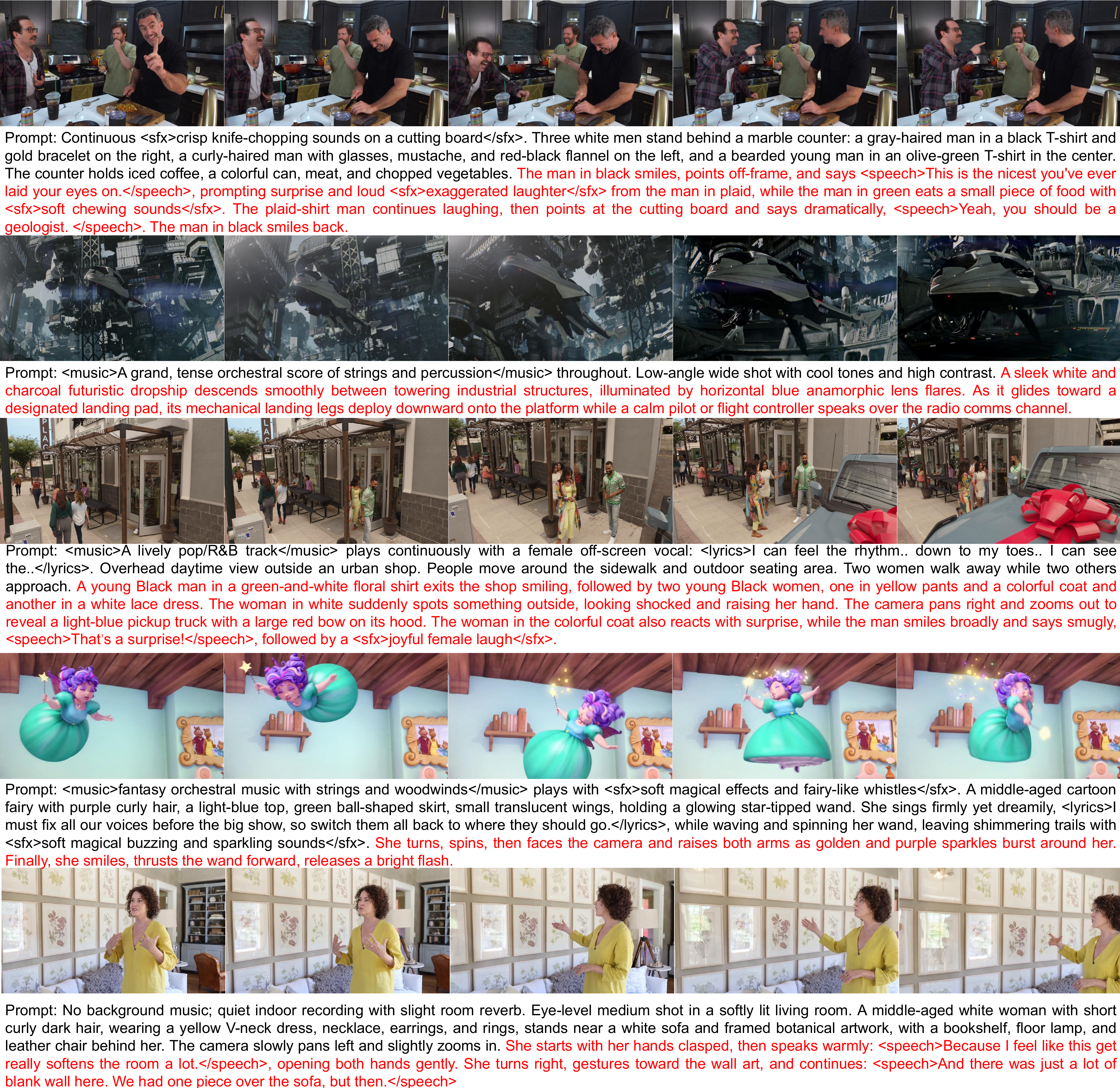}
\end{center}
\vspace{-0.3cm}
   \caption{Complex scene results (2/5). Please refer to the demo video for audio.
   }
\label{fig:complex_scene_2}
\vspace{-0.6cm}
\end{figure*}

\begin{figure*}[t!]
\begin{center}
\includegraphics[width=1.0\linewidth]{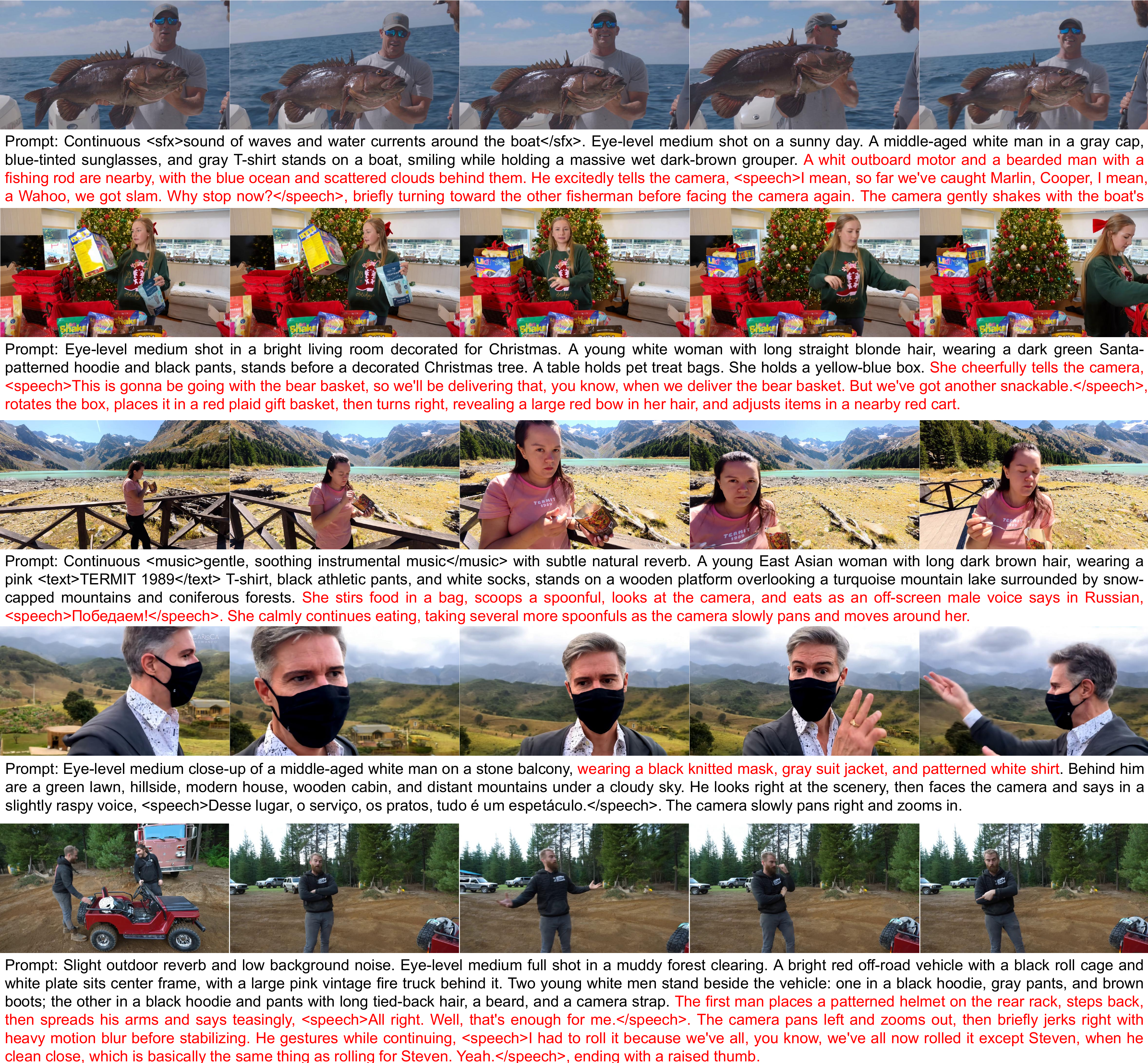}
\end{center}
\vspace{-0.3cm}
   \caption{Complex scene results (3/5). Please refer to the demo video for audio.
   }
\label{fig:complex_scene_3}
\vspace{-0.3cm}
\end{figure*}

\begin{figure*}[t!]
\begin{center}
\includegraphics[width=1.0\linewidth]{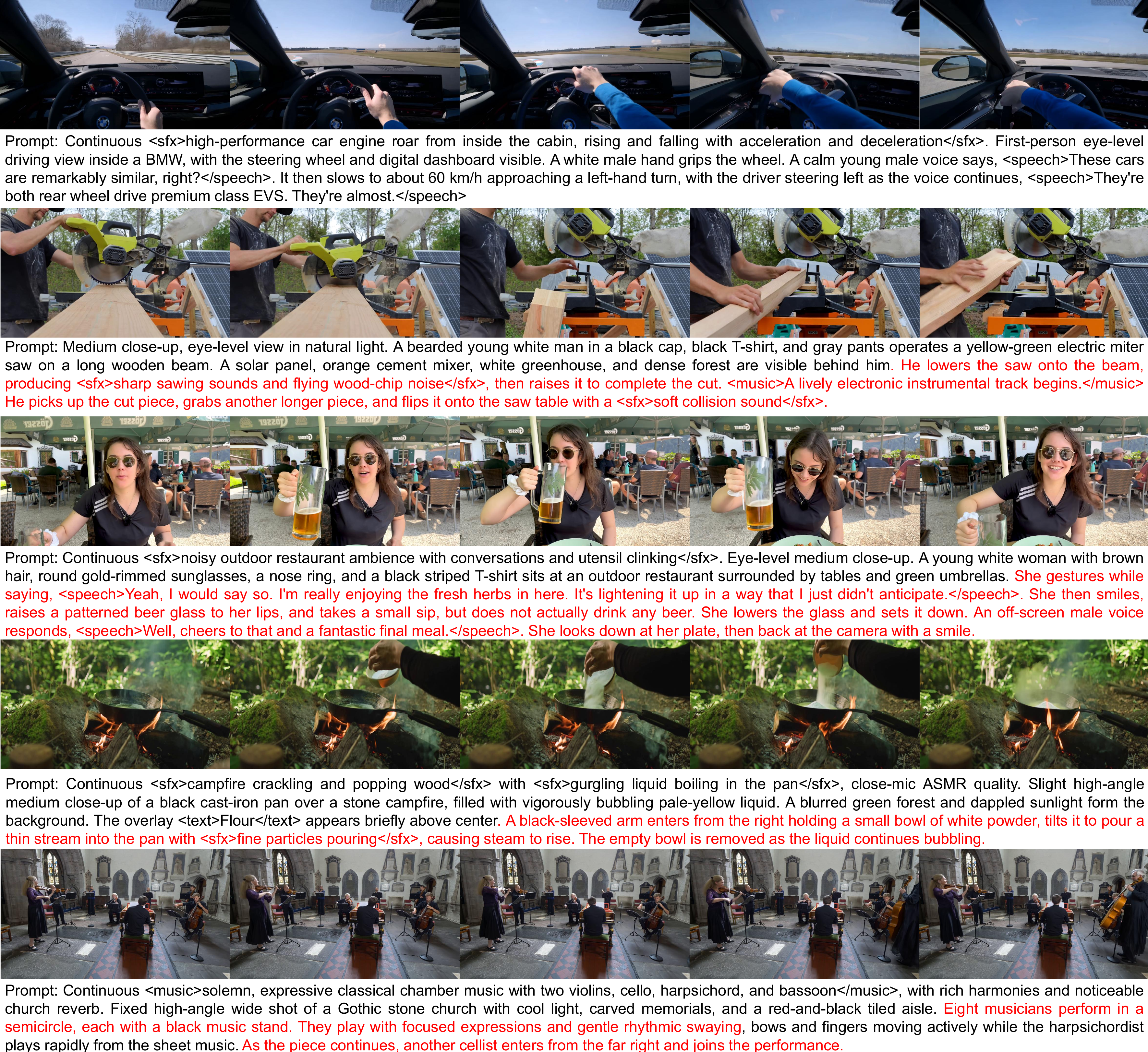}
\end{center}
\vspace{-0.3cm}
   \caption{Complex scene results (4/5). Please refer to the demo video for audio.
   }
\label{fig:complex_scene_4}
\vspace{-0.4cm}
\end{figure*}

\begin{figure*}[t!]
\begin{center}
\includegraphics[width=1.0\linewidth]{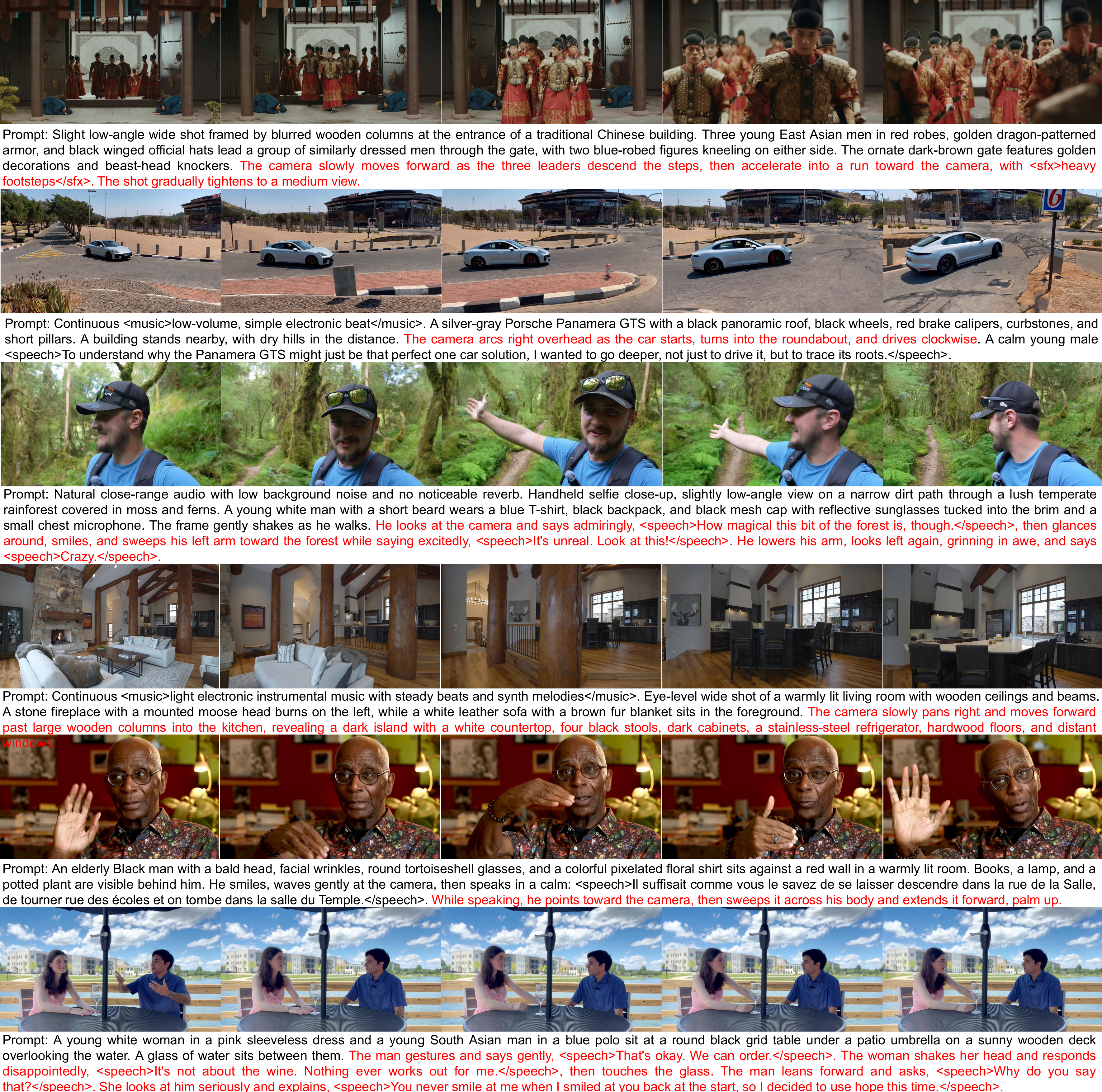}
\end{center}
\vspace{-0.3cm}
   \caption{Complex scene results (5/5). Please refer to the demo video for audio.
   }
\label{fig:complex_scene_5}
\vspace{-0.4cm}
\end{figure*}

Fig. \ref{fig:complex_scene_1}, Fig. \ref{fig:complex_scene_2}, Fig. \ref{fig:complex_scene_3}, Fig. \ref{fig:complex_scene_4}, and Fig. \ref{fig:complex_scene_5} show  the complex scene results.
The test cases encompass challenging scenarios that stress-test native 2K generation capabilities, including continuous human-object interactions with precise spatial contact, large-scale body movements with rapid spatial displacement, fine-grained hand and finger articulations, rich visual details such as intricate clothing textures, jewelry patterns, hairstyle variations, and facial features, multi-character interactions with overlapping motion dynamics, and environment-interaction sounds that adhere to physical plausibility.
For example, Fig.~\ref{fig:complex_scene_1} demonstrates narrative-driven character-environment interactions involving complex body movements and physically plausible ambient sounds that respect the acoustic properties of the scene.
Fig.~\ref{fig:complex_scene_2} showcases ordered multi-event sequences where each action must be completed before the next begins in the correct temporal order, requiring the model to maintain both temporal coherence and fine spatial detail across long horizons.
Fig.~\ref{fig:complex_scene_3} presents intricate human-object interactions that demand precise hand-object contact and coordinated limb motion, which are only resolvable at 2K where individual finger positions and object edges are clearly defined.
Fig.~\ref{fig:complex_scene_4} features scenes with dynamic perspective changes, testing whether the model preserves structural coherence and audio-visual synchronization under viewpoint variation.
Fig.~\ref{fig:complex_scene_5} contains cases with intense multi-character interaction dynamics, requiring the model to maintain identity consistency and body structure integrity for each character.
We can observe that Prism handles all these complex scenarios while preserving sharp visual details, stable body structures under large motion, and precise audio-visual synchronization throughout the generated sequences. These results directly validate that native 2K training with dynamic sparse attention enables the model to learn motion patterns at a spatial granularity that lower-resolution methods fundamentally cannot achieve.

\subsection{User Study Details}
\label{sec: user_study_details}

\begin{figure*}[t!]
\begin{center}
\includegraphics[width=1.0\linewidth]{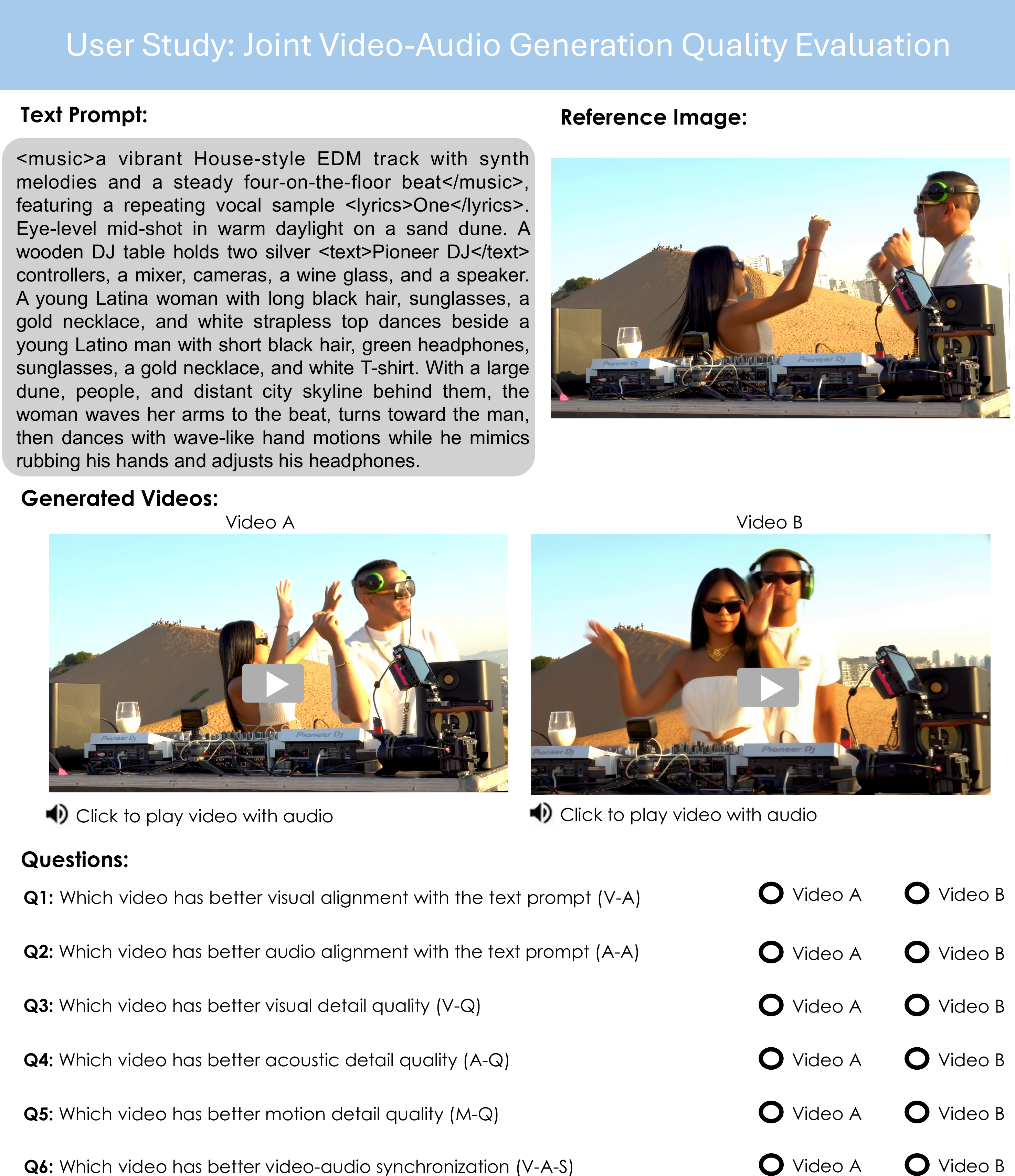}
\end{center}
\vspace{-0.3cm}
   \caption{The user study screenshot.
   }
\label{fig:user_study}
\vspace{-0.4cm}
\end{figure*}

Table \ref{table:user_study} shows the user study results. Fig. \ref{fig:user_study} shows a screenshot of our user study.
The 30 selected videos are randomly sampled from our 2K-Bench dataset, which contains 300 unseen 2K videos (10 seconds long) collected from social media platforms (YouTube and BiliBili). These videos include various ethnicities, genders, and indoor/outdoor settings.

A total of 200 individuals participated in this evaluation. Eligibility was restricted to adults aged 18 and older with normal or corrected-to-normal visual acuity and no reported auditory deficits. Participants were recruited through internal university mailing lists and campus announcements. The participant pool comprised approximately 75\% students and 25\% faculty members, with disciplinary backgrounds spanning computer science, engineering, and the arts. The sample maintains an approximately even gender balance.
The evaluation followed a two-alternative forced choice (2AFC) paradigm. On every trial, text prompts are first displayed, after which two videos appear in randomized order, one produced by Prism and the other by a competitor. Participants indicated which video they judged superior along six criteria as shown in Table \ref{table:user_study}. Every participant assessed the full set of 30 cases, with the left-right ordering randomized independently per trial to mitigate positional bias. All responses were gathered through an online survey interface, and preference rates were computed by aggregating judgments across the entire participant pool.

In terms of IRB approval, we consulted our university's institutional ethics board before conducting the study. The study was granted an exemption from full IRB review under the institution's minimal-risk research provisions, based on three conditions. First, the study involved only perceptual comparison of pre-generated video outputs without any intervention, deception, or interaction beyond viewing and selecting preferences. Second, no personally identifiable information was collected or stored beyond aggregate statistics. Third, participation was entirely voluntary, and participants were informed of the study's purpose and their right to withdraw at any time.
Each participant generally took 25 minutes to complete the study and received \$5 as compensation.

\begin{table}[t!]\small
\caption{User preference of Prism compared to other competitors. A higher score indicates users prefer more to our model.
}
\vspace{-0.05in}
\begin{center}
\renewcommand\arraystretch{1.1}
\scalebox{0.92}{
\begin{tabular}{l|cccccc}
\toprule
Prism vs     & V-A$\uparrow$ & A-A$\uparrow$ & V-Q$\uparrow$ & A-Q$\uparrow$ & M-Q$\uparrow$ & V-A-S$\uparrow$ \\ \midrule
Ovi~\citep{low2025ovi}        & 93.4\% & 92.6\% & 96.9\% & 96.2\% & 98.1\% & 91.7\% \\
LTX-2.3~\citep{hacohen2026ltx}   & 91.2\% & 88.7\% & 94.3\% & 93.5\% & 95.7\% & 87.4\% \\
MagiHuman~\citep{davinci-magihuman-2026} & 92.8\% & 90.3\% & 95.7\% & 94.1\% & 97.6\% & 89.5\% \\
MOVA~\citep{team2026mova}      & 92.0\% & 91.3\% & 96.5\% & 95.8\% & 97.6\% & 91.3\% \\ \bottomrule
\end{tabular}
}
\end{center}
\label{table:user_study}
\vspace{-0.10in}
\end{table}

\subsection{Human Subjects Data Concern}
\label{sec: human_subject_data_concern}

Both the training and testing datasets contain videos depicting human subjects with potentially identifiable features. All source videos were collected from publicly available platforms (BiliBili and YouTube), and we reached out to each original uploader through platform messaging to obtain explicit consent for non-commercial academic use. When the people shown in a video differ from the uploader, we additionally required the uploader to verify that they possessed legitimate authorization over the depicted individuals' likeness. Any video lacking such verification was removed from the dataset.

\subsection{Ethics Concerns}
\label{sec: ethics_concerns}

Prism enables native 2K joint video-audio synthesis that can benefit applications such as virtual character creation, cinematic production, and immersive storytelling. At the same time, the high visual fidelity of 2K generation heightens the potential for harmful misuse, as the richer spatial detail and sharper motion dynamics make synthesized content harder to distinguish from authentic recordings. Possible misuse scenarios include fabricating realistic impersonation videos, manipulating an individual's appearance or speech without authorization, and generating misleading audiovisual material.
We advocate several safeguards to address these risks. Generated outputs should carry both perceptible and imperceptible watermarks for provenance tracking. Deployment pipelines should incorporate automated deepfake detection and content moderation before any public release. Access should be governed by authenticated APIs with comprehensive usage logging. Finally, usage policies should explicitly prohibit the generation of content depicting real individuals without their informed consent.

\subsection{Limitations and Future Work}

\begin{figure*}[t!]
\begin{center}
\includegraphics[width=1.0\linewidth]{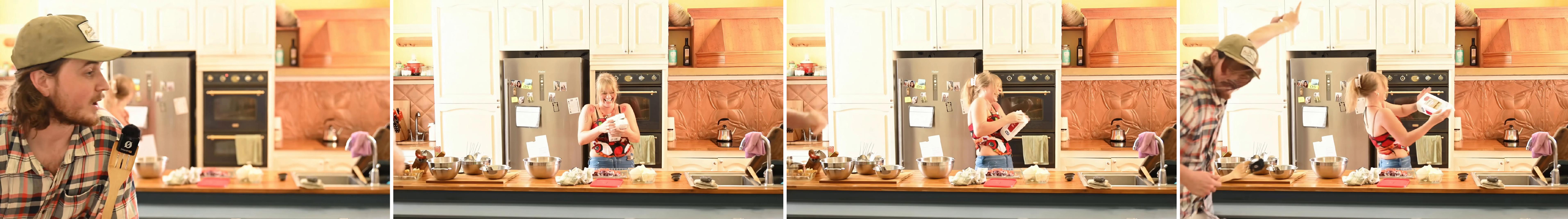}
\end{center}
\vspace{-0.3cm}
   \caption{One failure case of our Prism.
   }
\label{fig:limitation}
\vspace{-0.4cm}
\end{figure*}

Fig. \ref{fig:limitation} shows one failure case of our Prism.
When the camera focus target switches frequently (e.g., rapid zoom transitions between distant and close-up subjects), the generated content tends to exhibit blurriness or body distortion during transition frames. This is due to our macro-zone having a fixed temporal span of $Z_T{=}8$ frames, and a rapid camera change within that span causes the zone to contain highly heterogeneous content (e.g., both distant background and close-up face), for which a single block shape cannot faithfully represent both. Introducing adaptive temporal zone boundaries that split at detected camera transitions could address this and is left for future work.
Furthermore, due to limited GPU resources, we have not yet scaled our training beyond 2K resolution or explored larger DiT backbones. Scaling to 4K would further quadruple the token count and introduce new challenges for sparse attention design, while larger backbones with more parameters could potentially absorb richer visual details from the 2K spatial grid. We left it for future work.

\newpage

\begin{figure*}[t!]
\begin{center}
\includegraphics[width=1.0\linewidth]{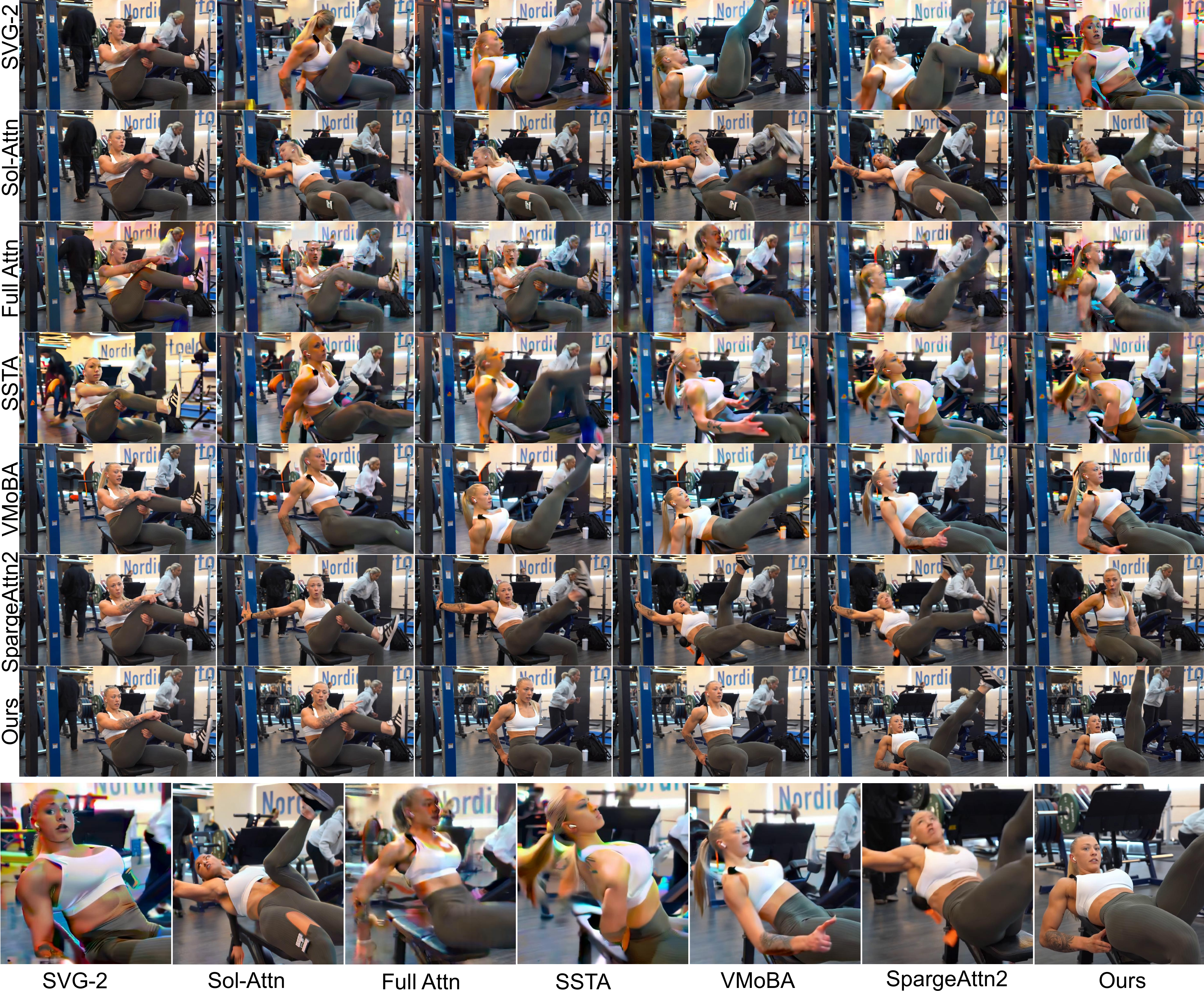}
\end{center}
\vspace{-0.4cm}
   \caption{Ablation study on different sparse attention methods.
   }
\label{fig:ablation_sparse_attn}
\vspace{-0.4cm}
\end{figure*}

\begin{figure*}[t!]
\begin{center}
\includegraphics[width=1.0\linewidth]{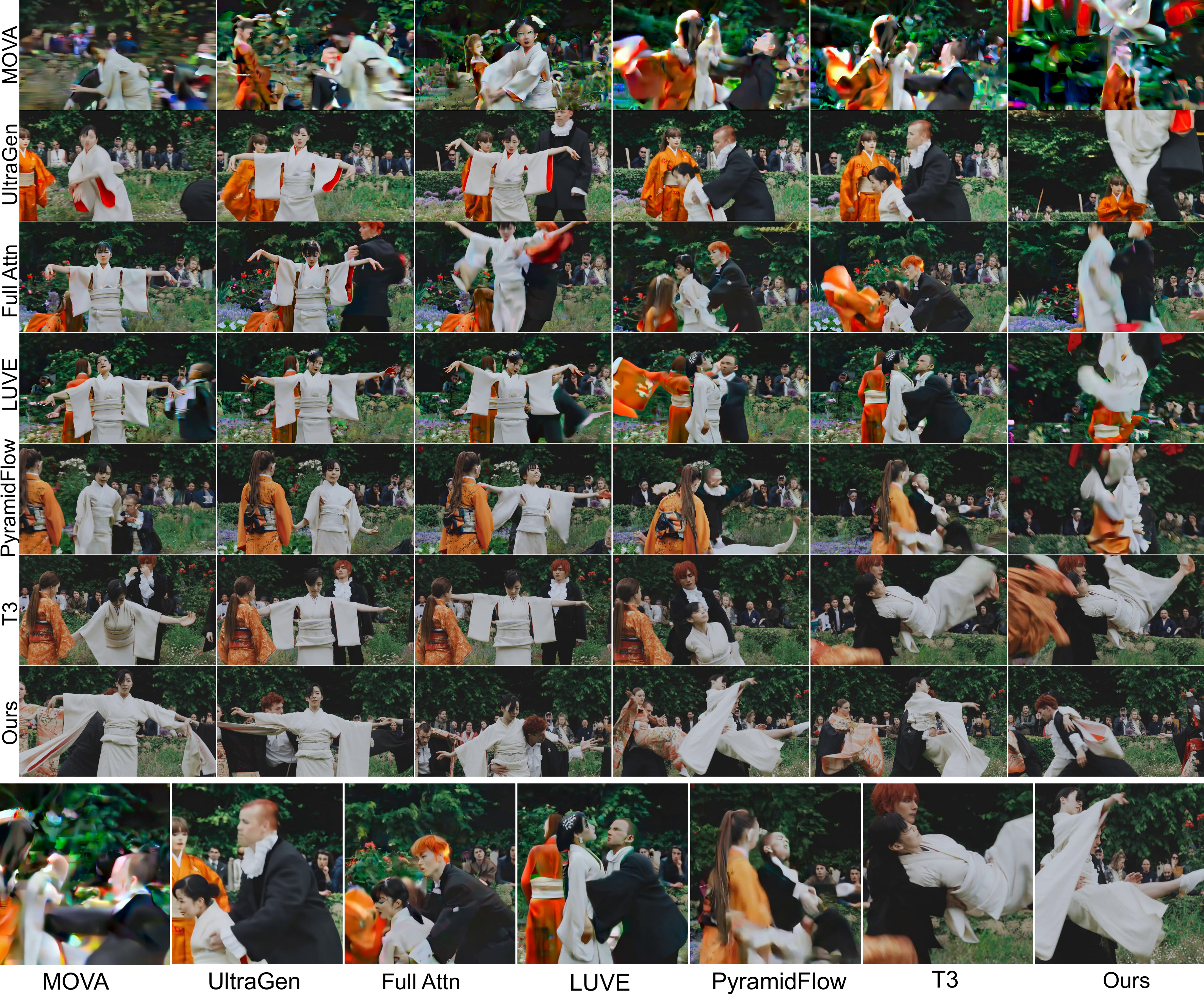}
\end{center}
\vspace{-0.4cm}
   \caption{Ablation study on different native high-resolution training methods.
   }
\label{fig:ablation_native_training}
\vspace{-0.4cm}
\end{figure*}

\end{document}